\pdfmapfile{=rpxfonts.map} % vendored IBM Plex fonts (top level)
\documentclass[letterpaper, 10 pt, conference]{ieeeconf}
\IEEEoverridecommandlockouts
\usepackage[utf8]{inputenc}
\usepackage[T1]{fontenc}
\usepackage[scaled=0.92]{helvet}
\usepackage{hyperref}
\hypersetup{
  colorlinks=true,
  linkcolor=rpxR!80!black,
  citecolor=rpxR!80!black,
  urlcolor=rpxP!80!black
}
\usepackage{url}
\usepackage{booktabs}
\usepackage{amsfonts}
\usepackage{amsmath}
\usepackage{amssymb}
\usepackage{graphicx}
\let\labelindent\relax  % ieeeconf already defines it
\usepackage{enumitem}
\usepackage{multirow}
\usepackage{subcaption}
\usepackage[table]{xcolor}
\usepackage{pifont}
\usepackage{stfloats}
\usepackage{cuted}
\usepackage{capt-of}
\usepackage[normalem]{ulem}
\usepackage{microtype}
\usepackage{tabularx}
\usepackage{placeins}
\usepackage{environ}
\usepackage{flushend}  % balance the two columns on the last page
\usepackage{tikz}
\usepackage{tcolorbox}
\usetikzlibrary{arrows.meta,positioning,calc,fit,backgrounds,decorations.pathreplacing}
\usepackage{pgfplots}
\pgfplotsset{compat=1.18}
\usepgfplotslibrary{groupplots}  % better character spacing & line breaks
\providecommand{\degree}{\ensuremath{^\circ}}

\definecolor{rpxInk}{HTML}{1F2937}
\definecolor{rpxSlate}{HTML}{475569}
\definecolor{rpxAccent}{HTML}{334155}
\definecolor{rpxRule}{HTML}{94A3B8}
\definecolor{rpxHair}{HTML}{CBD5E1}
\definecolor{rpxPale}{HTML}{EEF2F6}
\newcommand{\rpxImageRadius}{2.5pt}
\newsavebox{\rpxImageBox}
\newcommand{\rpxRoundedImage}[2][]{%
  \begingroup
  \sbox{\rpxImageBox}{\includegraphics[#1]{#2}}%
  \begin{tikzpicture}[baseline=0pt]
    \path[use as bounding box] (0,0) rectangle
      (\wd\rpxImageBox,\dimexpr\ht\rpxImageBox+\dp\rpxImageBox\relax);
    \begin{scope}
      \clip[rounded corners=\rpxImageRadius] (0,0) rectangle
        (\wd\rpxImageBox,\dimexpr\ht\rpxImageBox+\dp\rpxImageBox\relax);
      \node[anchor=south west,inner sep=0pt,outer sep=0pt] at (0,0)
        {\usebox{\rpxImageBox}};
    \end{scope}
  \end{tikzpicture}%
  \endgroup
}
\definecolor{tierE}{HTML}{2E7D46}
\definecolor{tierM}{HTML}{E0A21A}
\definecolor{tierH}{HTML}{B91C1C}
\definecolor{rpxWorst}{HTML}{FEE2E2}
\definecolor{rpxR}{HTML}{4338CA}% RPX wordmark, sunrise: indigo -> rose -> amber (X deepened from F59E0B for text on white)
\definecolor{rpxP}{HTML}{E11D48}
\definecolor{rpxX}{HTML}{D97706}
\definecolor{rpxPhClu}{HTML}{4A1D5E}% phase marks (plum ramp), as in Figs. 1-3
\definecolor{rpxPhInt}{HTML}{8A3FA8}
\definecolor{rpxPhCln}{HTML}{C79BDB}
\pgfdeclarehorizontalshading{rpxsunrise}{100bp}{color(0bp)=(rpxR); color(25bp)=(rpxR); color(50bp)=(rpxP); color(75bp)=(rpxX); color(100bp)=(rpxX)}
\newsavebox{\rpxbox}\newlength{\rpxw}\newlength{\rpxh}\newlength{\rpxd}
\DeclareRobustCommand{\coolname}{\leavevmode\begingroup
  \sbox{\rpxbox}{\textsc{RPX}}\setlength{\rpxw}{\wd\rpxbox}\setlength{\rpxh}{\ht\rpxbox}\setlength{\rpxd}{\dp\rpxbox}%
  \hbox to\rpxw{\pdfliteral{q 7 Tr}\rlap{\textsc{RPX}}\pdfliteral{0 Tr}%
  \rlap{\tikz[baseline=0pt]{\useasboundingbox (0,-\rpxd) rectangle (\rpxw,\rpxh);
    \shade[shading=rpxsunrise] (0,-\rpxd) rectangle (\rpxw,\rpxh);}}%
  \pdfliteral{Q}\hss}\endgroup}
\newcommand{\webpage}{https://irvlutd.github.io/RPX}
\newcommand{\todo}[1]{}  % disabled for camera-ready
\definecolor{rpxYes}{HTML}{1E1E1E}\definecolor{rpxPart}{HTML}{1E1E1E}\definecolor{rpxNo}{HTML}{BDB8AF}% Table I marks: ink, as in the heat-table theme
\DeclareRobustCommand{\cmark}{\tikz[baseline=-0.6ex]\fill[rpxYes] (0,0) circle (2.3pt);}
\newcommand{\logoH}{6.5pt}
\newcommand{\lnvidia}{\raisebox{-0.8pt}{\includegraphics[height=\logoH]{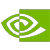}}\,}
\newcommand{\lalibaba}{\raisebox{-0.8pt}{\includegraphics[height=\logoH]{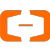}}\,}
\newcommand{\lopengvlab}{\raisebox{-0.8pt}{\includegraphics[height=\logoH]{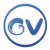}}\,}
\newcommand{\lgoogle}{\raisebox{-0.8pt}{\includegraphics[height=\logoH]{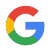}}\,}
\newcommand{\ldeepseek}{\raisebox{-0.8pt}{\includegraphics[height=\logoH]{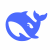}}\,}
\newcommand{\lmicrosoft}{\raisebox{-0.8pt}{\includegraphics[height=\logoH]{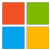}}\,}
\DeclareRobustCommand{\pmark}{\tikz[baseline=-0.6ex]{\draw[rpxPart,line width=0.6pt] (0,0) circle (2.0pt);\fill[rpxPart] (0,0) -- (90:2.0pt) arc (90:270:2.0pt) -- cycle;}}
\DeclareRobustCommand{\xmark}{\tikz[baseline=-0.6ex]{\path (0,0) circle (2.0pt);\fill[rpxNo] (0,0) circle (0.85pt);}}
\definecolor{rpxCmpRPX}{HTML}{F6E9E6}\definecolor{rpxCmpBand}{HTML}{F3F2EF}
\definecolor{rpxTableInk}{HTML}{2E2D29}
\definecolor{rpxTableAccent}{HTML}{8C1515}
\definecolor{rpxTableRobust}{HTML}{3B6854}
\definecolor{rpxTableRule}{HTML}{D8D7D3}
\definecolor{rpxTableSummary}{HTML}{FBF5F3}

\definecolor{rpxHQ}{HTML}{F2B830}\definecolor{rpxHQd}{HTML}{E0A21A}\definecolor{rpxHR}{HTML}{3F9A5A}\definecolor{rpxHRd}{HTML}{2E7D46}%
\definecolor{rpxHTrack}{HTML}{ECE9E2}\definecolor{rpxHInk}{HTML}{1E1E1E}\definecolor{rpxHMute}{HTML}{77736B}
\newcommand{\rpxHCell}[3]{\pgfmathtruncatemacro{\rpxHp}{round(#2*92)}\edef\rpxHx{\noexpand\cellcolor{#1!\rpxHp!white}}\rpxHx#3}
\newcommand{\hq}[1]{\rpxHCell{rpxHQ}{#1}{#1}}\newcommand{\hqB}[1]{\rpxHCell{rpxHQ}{#1}{\textbf{#1}}}
\newcommand{\hr}[1]{\rpxHCell{rpxHR}{#1}{#1}}\newcommand{\hrB}[1]{\rpxHCell{rpxHR}{#1}{\textbf{#1}}}
\newlength{\rpxHBarW}
\newcommand{\rpxHBar}[2]{\begin{tikzpicture}[baseline=-0.6ex]\fill[rpxHTrack,rounded corners=1.4pt](0,-1.4pt)rectangle(\rpxHBarW,1.4pt);\fill[#2,rounded corners=1.4pt](0,-1.4pt)rectangle(#1\rpxHBarW,1.4pt);\end{tikzpicture}}
\newcommand{\hvq}[2][0]{\ifnum#1=1\textbf{#2}\else#2\fi\hspace{2.5pt}\rpxHBar{#2}{rpxHQd}}
\newcommand{\hvr}[2][0]{\ifnum#1=1\textbf{#2}\else#2\fi\hspace{2.5pt}\rpxHBar{#2}{rpxHRd}}
\newcommand{\hk}[1]{{\color{rpxHMute}#1}}
\newcommand{\rpxHGroup}[1]{{\fontsize{6.4}{6.4}\selectfont\color{rpxHInk}#1}}
\newcommand{\rpxHHead}[1]{{\fontsize{6.4}{6.4}\selectfont\color{rpxHMute}#1}}
\newcommand{\rpxHSection}[1]{{\fontsize{6.4}{6.4}\selectfont\color{rpxHInk}\textbf{#1}}}
\newcommand{\hg}{\arrayrulecolor{white}\specialrule{0.4pt}{0pt}{0pt}}
\newcommand{\rpxHSwatchQ}{\textcolor{rpxHQ!92!white}{\rule{5pt}{5pt}}}
\newcommand{\rpxHSwatchR}{\textcolor{rpxHR!92!white}{\rule{5pt}{5pt}}}
\definecolor{rpxHL}{HTML}{F2B830}\newcommand{\rpxHSwatchL}{\textcolor{rpxHL!92!white}{\rule{5pt}{5pt}}}
\newcommand{\rpxHeatStyle}[2]{%
  \fontfamily{IBMPlexMono-TLF}\fontsize{#1}{\dimexpr#1pt+1.6pt\relax}\selectfont\color{rpxHInk}%
  \setlength{\tabcolsep}{#2}\renewcommand{\arraystretch}{1.0}%
  \setlength{\aboverulesep}{0pt}\setlength{\belowrulesep}{0pt}\setlength{\cmidrulewidth}{0.4pt}\setlength{\cmidrulekern}{3pt}}
\newcommand{\rpxHTop}{\arrayrulecolor{rpxHMute}\specialrule{0.5pt}{0pt}{0pt}}
\newcommand{\rpxHMid}{\arrayrulecolor{rpxHMute}\specialrule{0.4pt}{0pt}{1pt}}
\newcommand{\rpxHBarrier}{\arrayrulecolor{white}\specialrule{3pt}{0pt}{0pt}\arrayrulecolor{rpxHInk}\specialrule{0.9pt}{0pt}{0pt}}
\newcommand{\rpxHBottom}{\arrayrulecolor{rpxHMute}\specialrule{0.5pt}{1pt}{0pt}}
\newcommand{\hrange}[2]{#1--#2\hspace{3pt}\begin{tikzpicture}[baseline=-0.6ex]\fill[rpxHTrack,rounded corners=1.4pt](0,-1.4pt)rectangle(40pt,1.4pt);\fill[rpxHRd,rounded corners=1.4pt](#1*40pt,-1.4pt)rectangle(#2*40pt,1.4pt);\draw[rpxHInk,line width=0.5pt](0.86*40pt,-2.6pt)--(0.86*40pt,2.6pt);\end{tikzpicture}}
\newcommand{\rpxHGroupRules}[1]{\arrayrulecolor{rpxHMute!60}#1}

\makeatletter
\renewcommand{\@IEEEauthorblockconfadjspace}{-1.0em}
\makeatother

\title{\LARGE \bf Same Scene, Different Story: \\Evaluating Robot Perception Across Scene Phases in the Wild}

\author{\authorblockN{Jishnu Jaykumar P\textsuperscript{1,\textcolor{red}{*}},
Itay Kadosh\textsuperscript{2,\textcolor{red}{*}},
Narendhiran Vijayakumar\textsuperscript{3,\textcolor{red}{*}},
Srinanditha Kamath\textsuperscript{4},
Sai Haneesh Allu\textsuperscript{1},}
\authorblockN{Govind Tyagi Rangappa\textsuperscript{1},
Animesh Maheshwari\textsuperscript{1},
Jikai Wang\textsuperscript{1},
Yu Xiang\textsuperscript{1}\\[0.25em]\textsuperscript{\textcolor{red}{*}}\textit{Equal contribution}\hspace{0.8em}\textsuperscript{1}UT Dallas\hspace{0.8em}\textsuperscript{2}UF\hspace{0.8em}\textsuperscript{3}IIIT Hyderabad\hspace{0.8em}\textsuperscript{4}UT Austin}}

\renewcommand{\dbltopfraction}{0.9}

\begin{document}
\bstctlcite{IEEEexample:BSTcontrol}

\maketitle
\begin{strip}
    \vspace{-1.0cm}
    \centering
    \begin{tikzpicture}
      \node[inner sep=0pt,anchor=north west] (hero) at (0,0) {\rpxRoundedImage[width=\textwidth]{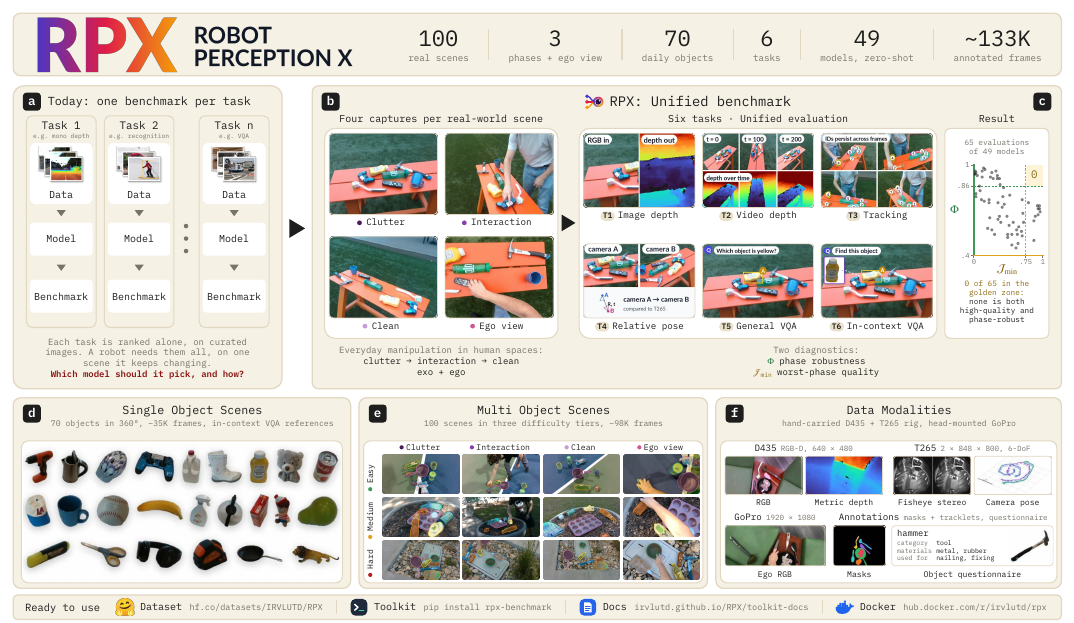}};
      % clickable release strip (figure coordinates in pt of the 515.52-pt-wide artwork)
      \newcommand{\rpxHeroLink}[5]{%
        \pgfmathsetlengthmacro\rpxLW{#3/515.52*\textwidth}\pgfmathsetlengthmacro\rpxLH{#4/515.52*\textwidth}%
        \pgfmathsetlengthmacro\rpxLX{#1/515.52*\textwidth}\pgfmathsetlengthmacro\rpxLY{-#2/515.52*\textwidth}%
        \node[inner sep=0pt,anchor=north west] at (\rpxLX,\rpxLY) {\href{#5}{\rule{0pt}{\rpxLH}\hspace*{\rpxLW}}};}%
      % generated by build_hero_v2.py: clickable regions of the Fig. 1 release strip
\rpxHeroLink{55.00}{286.40}{100.92}{10.20}{https://huggingface.co/datasets/IRVLUTD/RPX}
\rpxHeroLink{167.32}{286.40}{98.46}{10.20}{https://pypi.org/project/rpx-benchmark/}
\rpxHeroLink{277.18}{286.40}{111.96}{10.20}{https://irvlutd.github.io/RPX/toolkit-docs/}
\rpxHeroLink{400.54}{286.40}{102.96}{10.20}{https://hub.docker.com/r/irvlutd/rpx}

    \end{tikzpicture}
    \vspace{-4pt}
    \captionof{figure}{\coolname{} overview. \textbf{(a)}~Today each task has its own data and metric (thumbnails: KITTI~\cite{geiger2012kitti}, COCO~\cite{lin2014coco}). \textbf{(b,\,c)}~\coolname{}: \textbf{(b)}~four captures per scene, before (Clutter), during (Interaction), and after (Clean) manipulation plus an egocentric view; \textbf{(c)}~six tasks, one evaluation, two diagnostics ($\Phi$~phase robustness, $\mathcal{J}_{\min}$~worst-phase quality); none of the 65 evaluations reaches both $\mathcal{J}_{\min}{\geq}0.75$ and $\Phi{\geq}0.86$. \textbf{(d)}~Each object~\cite{padalunkal2023fewsol} captured alone in $360^{\circ}$. \textbf{(e)}~Multi-object scenes by difficulty and phase in real-world indoor and outdoor conditions. \textbf{(f)}~Sensors and annotation modalities. Bottom: dataset, toolkit, and docs.}
    \label{fig:overview}
    \vspace{2pt}
\end{strip}

\vspace{0pt}
% ── ICRA 2027 Abstract ──
% Fill the first column (and only the first column) at the standard font size.
\begin{abstract}\linespread{1.18}\selectfont
Robots increasingly pick perception models off the shelf, and their errors propagate to the actions that follow.
Yet these models are ranked task by task on separate leaderboards, not on the same scene as a robot changes it.
We present \coolname{} (Robot Perception X): 100 real-world scenes captured with commodity RGB-D sensors before, during, and after manipulation, plus an egocentric view, with ground truth shared across six perception tasks.
Two diagnostics, phase robustness~$\Phi$ and worst-phase quality~$\mathcal{J}_{\min}$, place every model on a common scale.
Across 49 models and 65 zero-shot evaluations, the right model depends on the task and scene condition: depth and tracking show significant phase effects ($p{<}0.001$), rankings shift with prompting mode and input modality, and no evaluation reaches both $\mathcal{J}_{\min}{\geq}0.75$ and $\Phi{\geq}0.86$.
\coolname{} pinpoints the phases and scene difficulty tiers where each model struggles; the dataset and toolkit are public.
Project page is at \url{https://irvlutd.github.io/RPX}.
\end{abstract}

\section{Introduction}
\label{sec:intro}

Robots increasingly rely on off-the-shelf pretrained perception models.
Modular pipelines run frozen depth estimators, trackers, and segmenters before grasping and navigation~\cite{liu2024okrobot}; the choice of vision-language backbone changes VLA policy performance, as shown in simulation~\cite{li2025robovlms}; agentic systems call perception models during task execution~\cite{huang2023voxposer, ahn2022saycan, lu2026aspire}; and off-the-shelf trackers, detectors, and segmenters label robot training data~\cite{wen2024atm, zawalski2024ecot, xu2024im2flow2act}.
Perception errors propagate to the actions that follow: retrieving the wrong object is the most frequent specific failure in OK-Robot's home trials~\cite{liu2024okrobot}, perception and planning failures dominate pick-and-place errors in SceneReplica~\cite{khargonkar2024scenereplica}, and a missed object or merged mask cascades into a wrong grasp~\cite{padalunkal2024iteach}.

In everyday manipulation, a robot needs several perception capabilities on one view, and the scene changes as it works.
A tabletop starts cluttered, a hand rearranges objects, and the scene ends clean, observed externally or from the manipulator's viewpoint.
Robot cameras move on arms, heads, or mobile bases, and are commodity RGB-D sensors whose images differ from the curated web imagery most models train on.
Yet each model is chosen on its own benchmark: depth on NYU~Depth~V2~\cite{silberman2012indoor}, tracking on DAVIS~\cite{perazzi2016benchmark}, VLMs on internet QA, with no shared manipulation protocol (Table~\ref{tab:comparison}).

We present \coolname{} (\textbf{R}obot \textbf{P}erception \textbf{X}; Fig.~\ref{fig:overview}), a unified real-world benchmark that lets roboticists select pretrained perception models from evidence gathered under robot conditions rather than from leaderboard rank on unrelated data.
Each of 100 scenes is captured with an Intel D435 (RGB-D, $640{\times}480$) and T265 (6-DoF pose), carried through the scene by a person as a mobile-robot proxy, in three phases: \textbf{Clutter} (before manipulation), \textbf{Interaction} (during), and \textbf{Clean} (after), plus an egocentric view during Interaction.
Six perception tasks share the same scenes, which are split into Easy, Medium, and Hard tiers by RPX-DS, a weighted composite of 27 descriptors, including human annotation effort.
Two diagnostics, phase robustness~$\Phi$ and worst-phase quality~$\mathcal{J}_{\min}$ (Sec.~\ref{sec:eval-primitives}), place every model in a common quality--robustness space across tasks, phases, and difficulty tiers.

\vspace{0.3em}
\noindent\textbf{Contributions.}
\begin{enumerate}[nosep,leftmargin=*,label=(\arabic*)]
    \item We present \coolname{}, a real-world benchmark for selecting pretrained perception models on the scene transitions robots produce: 100 scenes, six tasks, three exocentric phases, and an egocentric view.
    \item We evaluate 49 models (65 task evaluations) zero-shot and show that the right model depends on the task and scene condition: depth and tracking show significant phase effects ($p{<}0.001$), relative pose shows no significant within-tier Clutter--Clean effect, rankings change with prompting mode and input modality, and none of the 65 evaluations reaches both $\mathcal{J}_{\min}{\geq}0.75$ and $\Phi{\geq}0.86$.
    \item We release the dataset and an open-source toolkit with documentation, Docker images, and model adapters for reproducible evaluation and extension to new models, metrics, and tasks the data modalities support.
\end{enumerate}

% ── ICRA 2027 Related Work ──
\section{Related Work}
\label{sec:related}

\begin{table}[!th]
\centering
\caption{Comparison with benchmarks relevant to robot perception. {\cmark}~supported; {\pmark}~partial; {\xmark}~absent. Depth GT: measured, full-resolution depth. $^\|$Policy-learning dataset. $^\star$Static extrinsics only. $^\ddagger$Trajectories/trials. $^\S$Seven BOP-Classic-Core datasets. $^\dagger$Tasks, not scenes. $^\circ$Binocular IR reported. $^\P$336 real objects ${\times}$ 9 views; plus 40K synthetic scenes.}
\label{tab:comparison}
\rpxHeatStyle{6.6}{1.2pt}%
\newcommand{\cmpHead}[1]{\rotatebox{72}{\fontsize{6.2}{6.2}\selectfont\color{rpxHInk}#1}}%
\newcommand{\cmpGroup}[1]{\rpxHGroup{#1}}%
\newcommand{\cmpSection}[1]{\noalign{\vskip 1.6pt}\multicolumn{9}{@{}l}{\cellcolor{rpxHTrack}\rule[-2pt]{0pt}{9pt}\hspace{4pt}\rpxHSection{#1}}\\\noalign{\vskip 0.8pt}}%
\newcommand{\cmpStat}[1]{\cellcolor{rpxHQ!22!white}\textbf{#1}}%
\vspace{2pt}
\resizebox{\columnwidth}{!}{%
\begin{tabular}{@{\hspace{4pt}}>{\fontsize{5.8}{7.4}\selectfont\raggedright\arraybackslash}p{110pt} >{\columncolor{rpxHQ!22!white}}c cccc ccc@{\hspace{4pt}}}
\rpxHTop
\multicolumn{1}{@{\hspace{4pt}}l}{\rule{0pt}{9pt}} & \multicolumn{1}{c}{} & \multicolumn{4}{c}{\cmpGroup{Perception}} & \multicolumn{3}{c}{\cmpGroup{Robot data}} \\
\rpxHGroupRules{\cmidrule(lr){3-6} \cmidrule(lr){7-9}}
 & \raisebox{1pt}{\rmfamily\fontsize{8.5}{8.5}\selectfont\textbf{\coolname{}}}
 & \cmpHead{FewSOL~\cite{padalunkal2023fewsol}}
 & \cmpHead{ScanNet~\cite{dai2017scannet}}
 & \cmpHead{YCB-V~\cite{xiang2018posecnn}}
 & \cmpHead{BOP-Classic$^\S$~\cite{nguyen2025bop}}
 & \cmpHead{DROID$^\|$~\cite{khazatsky2024droid}}
 & \cmpHead{RH20T~\cite{fang2023rh20t}}
 & \cmpHead{OK-Robot~\cite{liu2024okrobot}} \\
\rpxHMid
\cmpSection{data properties}
Real-world RGB-D             & \cmark & \cmark & \cmark & \cmark & \cmark & \pmark & \cmark & \pmark \\
Metric depth GT              & \cmark & \cmark & \cmark & \cmark & \cmark & \pmark & \pmark & \xmark \\
6-DoF camera pose            & \cmark & \cmark & \cmark & \cmark & \pmark & \pmark$^\star$ & \pmark$^\star$ & \xmark \\
Calibrated stereo pairs      & \cmark & \xmark & \xmark & \xmark & \xmark & \cmark & \pmark$^\circ$ & \xmark \\
Instance masks               & \cmark & \cmark & \cmark & \cmark & \cmark & \xmark & \xmark & \xmark \\
Object questionnaire         & \cmark & \cmark & \xmark & \xmark & \xmark & \xmark & \xmark & \xmark \\
Multi-task GT, same scenes   & \cmark & \cmark & \cmark & \cmark & \cmark & \xmark & \xmark & \xmark \\
\cmpSection{protocol}
Three-phase capture          & \cmark & \xmark & \xmark & \xmark & \xmark & \xmark & \xmark & \xmark \\
Manipulation recorded        & \cmark & \xmark & \xmark & \xmark & \xmark & \cmark & \cmark & \xmark \\
Egocentric stream            & \cmark & \xmark & \xmark & \xmark & \xmark & \pmark & \pmark & \xmark \\
Shared objects across phases & \cmark & \xmark & \xmark & \xmark & \xmark & \xmark & \xmark & \xmark \\
Single + multi-object scenes & \cmark & \pmark & \xmark & \xmark & \cmark & \xmark & \xmark & \xmark \\
Scene difficulty strata      & \cmark & \xmark & \xmark & \xmark & \xmark & \xmark & \xmark & \xmark \\
\cmpSection{scale}
\# Scenes     & \cmpStat{100} & 336$^\P$ & 1.5K & 92 & var. & 564 & 147$^\dagger$ & 10 \\
\# Frames     & \cmpStat{\boldmath${\sim}$133K} & 3K$^\P$ & 2.5M & 134K & var. & 76K$^\ddagger$ & 110K$^\ddagger$ & 171$^\ddagger$ \\
Focus & \cmpStat{\textit{deploy.}} & few-shot & recon. & pose & pose & imit. & imit. & manip. \\
\rpxHBottom
\end{tabular}%
}
\end{table}

\noindent\textbf{Pretrained models as robot perception.}\enspace
Pretrained vision models serve as encoders in VLA policies~\cite{kim2024openvla, black2024pi0, bjorck2025gr00t}, as preprocessing stages in modular pipelines~\cite{liu2024okrobot}, as few-shot object recognizers for grasping~\cite{padalunkal2024protoclip}, and as pseudo-label generators for world models~\cite{zhen2025tesseract}.
Adapting pretrained features to novel instances improves detection in manipulation scenes~\cite{lu2024nidsnet}.
Li~et~al.~\cite{li2025robovlms} demonstrated substantial differences in VLA performance across backbones; on real hardware, adapting perception to the deployment scene~\cite{padalunkal2024iteach} directly improves manipulation.
\coolname{} evaluates these models together on real scenes captured with consumer RGB-D sensors common in robot perception research, under scene change.

\smallskip\noindent\textbf{Per-task benchmarks.}\enspace
Depth models are developed on NYU~Depth~V2~\cite{silberman2012indoor}, trackers on DAVIS~\cite{perazzi2016benchmark}, detectors on COCO~\cite{lin2014coco}, and odometry on KITTI~\cite{geiger2012kitti}, each with its own camera, scenes, and domain.
RoboDepth~\cite{kong2023robodepth} tests depth robustness by applying synthetic corruptions to existing benchmark images; the perturbations are added after capture, not caused by scene change.
SenseShift6D~\cite{han2025senseshift6d} sweeps sensor parameters for 6-DoF pose but evaluates a single task on static scenes.
\coolname{} differs in two ways: the perturbation is the manipulation itself, recorded with commodity robot sensors, and all six tasks share the same scenes.

\smallskip\noindent\textbf{Robot datasets.}\enspace
Table~\ref{tab:comparison} compares \coolname{} against datasets a practitioner might use instead.
YCB-Video~\cite{xiang2018posecnn} and BOP-Classic~\cite{nguyen2025bop} provide object-pose and mask ground truth (GT), but do not use a shared before--during--after manipulation protocol.
DROID~\cite{khazatsky2024droid} and RH20T~\cite{fang2023rh20t} record manipulation at scale but lack per-frame perception GT.
HOI4D~\cite{liu2022hoi4d} and HOT3D~\cite{banerjee2025hot3d} capture hand--object interaction with dense labels.
GraspClutter6D~\cite{shin2025graspclutter6d} provides multi-sensor depth in heavy clutter, but the scenes are static.
SceneReplica~\cite{khargonkar2024scenereplica} benchmarks manipulation by replicating tabletop arrangements, focusing on grasping success rather than perception evaluation.
\coolname{} provides paired captures of the same scene through manipulation, with ground truth for multiple perception tasks on shared data.

% ── ICRA 2027 Benchmark Design ──
\section{Benchmark Design}
\label{sec:method}

\subsection{Three-Phase Protocol}
\label{sec:three-phase}

Each scene is recorded in three successive phases that bracket a manipulation episode:
\begin{itemize}[nosep,leftmargin=1.2em]
    \item[\textcolor{rpxPhClu}{\scriptsize$\blacksquare$}] \textbf{Clutter}: the scene as found, objects in their initial arrangement.
    \item[\textcolor{rpxPhInt}{\scriptsize$\blacksquare$}] \textbf{Interaction}: a person rearranges the objects by hand, as when collecting robot demonstrations~\cite{padalunkal2024iteach, allu2025hrt1, khazatsky2024droid}.
    \item[\textcolor{rpxPhCln}{\scriptsize$\blacksquare$}] \textbf{Clean}: the scene after rearrangement.
\end{itemize}
\noindent An exocentric RGB-D stream covers all three phases; an egocentric stream records Interaction from the manipulator's viewpoint (Sec.~\ref{sec:dataset}).
The paired design keeps the objects, sensors, and environment shared across phases, reducing between-scene confounds while measuring the changes that accompany manipulation.

\subsection{Evaluation Primitives}
\label{sec:eval-primitives}

Averaging over metrics can mask the collapse of one behind the stability of another, and reveals nothing about whether model output changes when the scene does.
We instead summarize each model with two unit-free diagnostics: a phase-robustness effect size~$\Phi$ and a worst-phase quality index~$\mathcal{J}_{\min}$.

\begin{figure}[!tb]
    \centering
    \rpxRoundedImage[width=\columnwidth]{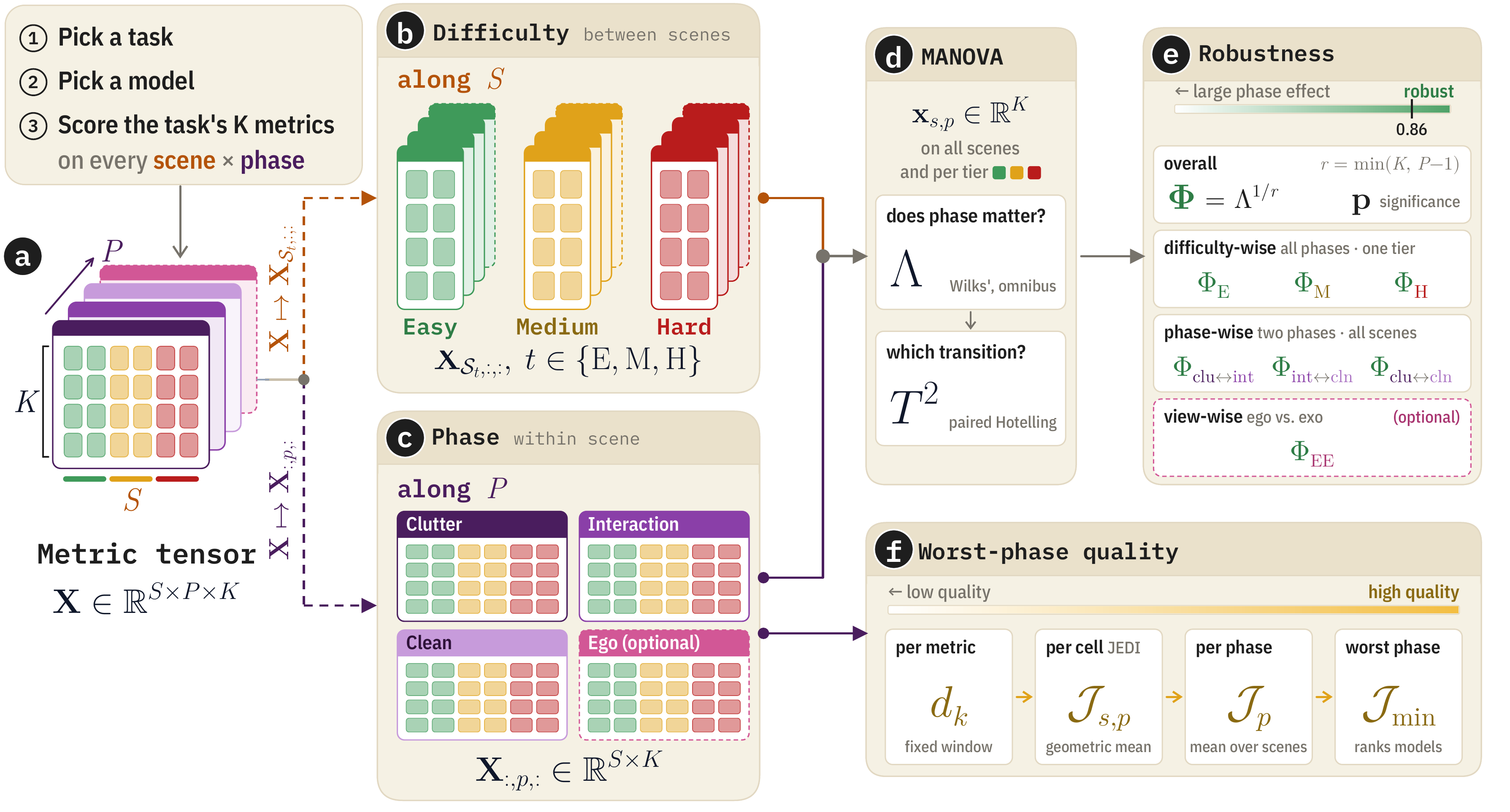}
    \caption{Evaluation pipeline. \textbf{(a)}~Per task and model (top left), the task's $K$ metric scores per scene and phase form a tensor (ego dashed). \textbf{(b)}~Slices along $S$ form difficulty tiers; \textbf{(c)}~slices along $P$ are compared within each scene and feed the quality score. \textbf{(d)}~Repeated-measures MANOVA yields Wilks' $\Lambda$ and per-transition tests, also run per tier. \textbf{(e)}~Phase robustness $\Phi$: overall, per transition, per tier. \textbf{(f)}~Per-metric desirabilities combine into $\mathcal{J}_p$; the worst phase, $\mathcal{J}_{\min}$, ranks models.}
    \label{fig:metric-pipeline}
\end{figure}

Each task produces $K$ metrics per (scene, phase) cell, forming a tensor $\mathbf{X}\in\mathbb{R}^{S\times P\times K}$ with $S{=}100$ scenes and $P{=}3$ phases ($P{=}2$ for T4), plus the ego capture for T3 and T5--T6 (Fig.~\ref{fig:metric-pipeline}).
Slicing along $P$ yields the per-phase $S{\times}K$ matrices compared within each scene; slicing along $S$ yields the difficulty tiers $\mathcal{S}_t$, $t\in\{\mathrm{E},\mathrm{M},\mathrm{H}\}$, on which every test below is repeated.
For image depth, $K{=}4$ (AbsRel, RMSE, SILog, $\delta_1$~\cite{eigen2014depth}); for tracking, $K{=}3$ (HOTA~\cite{luiten2021hota}, MOTA~\cite{bernardin2008evaluating}, IDSW); Table~\ref{tab:summary} lists $K$ for every task.
Metrics within a task are correlated but not redundant, motivating a joint test.

\smallskip
\noindent\textbf{Phase robustness\/ $\Phi$.}\enspace
To test whether the $K$-metric vector shifts across phases, we use a one-way repeated-measures MANOVA~\cite{wilks1932certain}.
Let $\mathbf{x}_{s,p} \in \mathbb{R}^K$ be the z-scored, direction-aligned metric vector for scene $s$ in phase $p$, and $\tilde{\mathbf{x}}_{s,p} = \mathbf{x}_{s,p} - \bar{\mathbf{x}}_{s,\cdot}$ the same vector centered on the scene mean.
The between-phase and within-phase scatter are captured by two $K{\times}K$ SSCP matrices:
\begin{equation}
\mathbf{H} = S \!\sum_{p=1}^{P} \bar{\mathbf{x}}_{\cdot,p}\,\bar{\mathbf{x}}_{\cdot,p}^\top, \quad
\mathbf{E} = \!\sum_{s,p} (\tilde{\mathbf{x}}_{s,p} {-} \bar{\mathbf{x}}_{\cdot,p})(\tilde{\mathbf{x}}_{s,p} {-} \bar{\mathbf{x}}_{\cdot,p})^\top
\label{eq:sscp}
\end{equation}
where $S$ is the number of scenes and $\bar{\mathbf{x}}_{\cdot,p}$ is the phase mean of the centered data.
$\mathbf{H}$ captures the signal (between-phase variance); $\mathbf{E}$ captures the noise (within-phase variance).
The effect size follows from Wilks' $\Lambda$:
\begin{equation}
\Lambda = \frac{|\mathbf{E}|}{|\mathbf{H}+\mathbf{E}|}, \qquad
\eta_p^2 = 1 - \Lambda^{1/r}, \qquad \Phi = 1 - \eta_p^2,
\label{eq:phi}
\end{equation}
with $r = \min(K,\, P{-}1)$.
$\Phi$ near~1 indicates a small estimated phase effect; $\Phi \leq 0.86$ corresponds to a large effect under the $\eta^2 \geq 0.14$ convention~\cite{cohen1988power}.
Significance is assessed via Rao's $F$-approximation~\cite{rao1951asymptotic}, with $P{-}1$ hypothesis and $(S{-}1)(P{-}1)$ error degrees of freedom, assuming multivariate sphericity.
This omnibus $\Phi$ tests all $P$ phases jointly.
To pinpoint which transition drives the effect (e.g., Clutter${\leftrightarrow}$Interaction vs.\ Interaction${\leftrightarrow}$Clean), we run the same procedure on each pair of phases via paired Hotelling $T^2$ tests~\cite{hotelling1931generalization} on the difference vectors $\mathbf{x}_{s,p_a} - \mathbf{x}_{s,p_b}$, Holm-corrected~\cite{holm1979simple} over the $\binom{P}{2}$ pairs.
We also repeat the omnibus test within each difficulty tier $\mathcal{S}_t$, yielding per-tier $\Phi$ values (E, M, H in the tables).
T4's reported summary $\Phi$ and quality scores average the three difficulty-tier results; its $p$ entries refer to the individual tier tests.

\smallskip
\noindent\textbf{Worst-phase quality\/ $\mathcal{J}_{\min}$.}\enspace
Each metric $k$ is mapped to a desirability $d_k = \mathrm{clip}\bigl((v_k - v_k^{\mathrm{worst}})/(v_k^{\mathrm{best}} - v_k^{\mathrm{worst}}),\,0,\,1\bigr)$, where $(v_k^{\mathrm{worst}}, v_k^{\mathrm{best}})$ is a fixed window per metric mapping worst to 0 and best to 1 (e.g., AbsRel: $0.5{\to}0$; $\delta_1$: $0{\to}1$; HOTA: $0{\to}1$).
We compute the geometric mean of the $K$ desirabilities~\cite{harrington1965desirability} in each scene--phase cell, the Joint Empirical Desirability Index (JEDI)~$\mathcal{J}_{s,p}$, then average over scenes:
\begin{equation}
\begin{aligned}
\mathcal{J}_{s,p} &= \Bigl(\prod_{k=1}^{K} d_{k,s,p}\Bigr)^{\!1/K}, \\
\mathcal{J}_p &= \frac{1}{S}\sum_s \mathcal{J}_{s,p}, \qquad
\mathcal{J}_{\min} = \min_{p}\; \mathcal{J}_p.
\end{aligned}
\label{eq:jedi}
\end{equation}
The geometric mean ensures that a single collapsed metric drives the corresponding cell's quality toward zero.
$\mathcal{J}_{\min}$, the quality of the worst phase, is our primary ranking criterion.
Overall $\mathcal{J}$ averages the exocentric phase scores; Ego is reported separately and excluded from $\mathcal{J}_{\min}$.
The $K$ metrics and their ranges are task-specific, so equal quality scores across tasks do not imply equal downstream utility.

\smallskip
\noindent\textbf{Why two numbers?}\enspace
$\Phi$ and $\mathcal{J}_{\min}$ are complementary by construction.
A model can score high $\Phi$ (phases look alike) despite low $\mathcal{J}_{\min}$ (poor quality across phases).
The diagnostics favor both high $\mathcal{J}_{\min}$ (high quality in the worst phase) and high $\Phi$ (small estimated phase effects).

\subsection{Capture Protocol and Dataset}
\label{sec:dataset}\label{sec:esd}

\begin{figure}[!tb]
    \centering
    \rpxRoundedImage[width=\columnwidth]{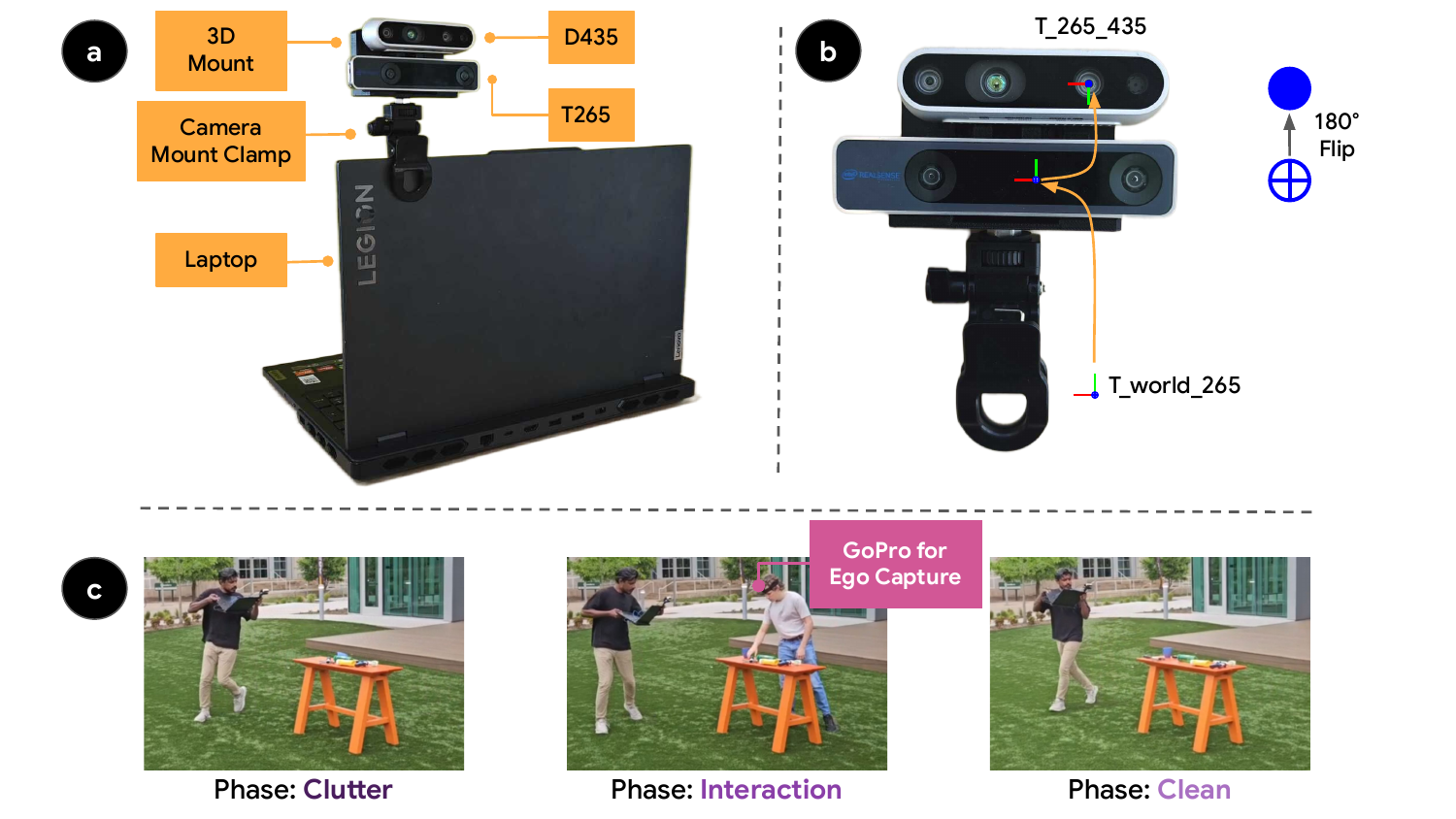}
    \caption{Capture setup. \textbf{(a)}~D435\,+\,T265 rig on a laptop mount, carried by hand. \textbf{(b)}~Sensor detail with the T265$\to$D435 extrinsic and world$\to$T265 pose frames. \textbf{(c)}~Three phases of one scene; the GoPro is visible during Interaction.}
    \label{fig:rig}
\end{figure}

The protocol compares successive scene states around the same manipulation episode.
Two people record each scene using a single rig (Fig.~\ref{fig:rig}).
One carries the sensors: a RealSense D435 (RGB-D, $640{\times}480$) and a T265 (6-DoF visual-inertial odometry) on a shared laptop mount (Fig.~\ref{fig:rig}a); Intel's reference T265$\to$D435 extrinsic (Fig.~\ref{fig:rig}b) gives every D435 frame a metric camera pose.
The second person rearranges objects during Interaction and wears a head-mounted GoPro ($1920{\times}1080$, RGB only, $6.75\times$ the exocentric resolution) that records the egocentric stream.
The carrier acts as a mobile-robot proxy, reproducing a working robot's moving viewpoint rather than a static tripod capture; since RPX evaluates perception only and records no robot joint or state data, a human carrier suffices and makes capture faster and safer.
All three phases are recorded with the same sensors and object set; viewpoint, framing, and lighting vary naturally, so the phase factor tested in Sec.~\ref{sec:eval-primitives} measures their combined change across the manipulation episode (Fig.~\ref{fig:rig}c).

\textbf{Difficulty tiers.}\enspace Scenes are stratified into Easy~(33), Medium~(33), and Hard~(34) by a weighted composite of 27 descriptors---covering annotation effort, scene complexity, occlusion, depth quality, photometric conflict, temporal stability, camera motion, and fisheye/stereo characteristics---all computed by the released \href{https://pypi.org/project/rpx-benchmark/}{toolkit}.

\textbf{Contents.}\enspace The dataset comprises ${\sim}133$K frames across 100 indoor and outdoor scenes (outdoor examples in Fig.~\ref{fig:overview}e) with 70 everyday objects, organized into two streams.
The \textbf{multi-object stream} (${\sim}98$K frames) contains $100 \times 3$ phase captures at 250 D435 RGB-D frames per phase, plus the egocentric GoPro stream during Interaction.
Each exocentric frame has metric depth, a calibrated T265 fisheye stereo pair and 6-DoF camera pose, instance-level masks (SAM\,2~\cite{ravi2024sam2} initialized with GroundingDINO, then manually corrected), and per-object attributes from a questionnaire adapted from FewSOL~\cite{padalunkal2023fewsol}, used by VQA; the RGB-only ego stream has masks, tracklets, and VQA question--answer pairs.
The \textbf{single-object stream} (${\sim}35$K frames) captures each of the 70 objects in isolation on a fiducial-marker board, as in FewSOL~\cite{padalunkal2023fewsol}, with the same per-frame modalities and masks, providing the reference images for in-context grounding (T6); the board gives an independent audit of the T265 poses (Sec.~\ref{sec:experiments}).

\begin{table*}[t]
\centering
\caption{\textbf{Summary: significant phase effects in depth and tracking; none detected in relative pose.} Top/bottom model by $\mathcal{J}_{\min}$; $\Phi$ range on $[0,1]$; tick: $\Phi{=}0.86$ (at or below: large effect). T4: 2 phases (Clutter vs.\ Clean), tier-averaged scores; other tasks: 3 phases. Hardest transition: lowest mean transition $\Phi$ across models. (most): four of 24 VQA model--task pairs do not reach $p{<}.001$. $^\ddag$Text-prompted (no GT masks). $^\S$Tied at displayed precision with DeepSeek-VL2 1.0B.}
\label{tab:summary}
\rpxHeatStyle{6.3}{1.6pt}%
\vspace{2pt}
\begin{tabular*}{\textwidth}{@{\extracolsep{\fill}}l c c l c l c l l l@{}}
\rpxHTop
\rule{0pt}{9pt}\rpxHHead{task} & \rpxHHead{$K$} & \rpxHHead{$M$} & \rpxHHead{top model} & \rpxHHead{$\mathcal{J}_{\min}$} & \rpxHHead{bottom model} & \rpxHHead{$\mathcal{J}_{\min}$} & \rpxHHead{$\Phi$ range} & \rpxHHead{hardest transition} & \rpxHHead{$p$} \\[1pt]
\rpxHMid
T1 image depth     & 4 & 10 & Metric3D V2~\cite{hu2024metric3dv2} & \hqB{0.71} & Lotus-2~\cite{he2024lotus} & \hq{0.35} & \hrange{0.53}{0.79} & clu$\to$int & ${<}.001$ \\\hg
T2 video depth     & 6 & 10 & VGGT~\cite{wang2024vggt} & \hqB{0.70} & RollingDepth~\cite{rollingdepth2024} & \hq{0.20} & \hrange{0.45}{0.77} & clu$\to$int & ${<}.001$ \\\hg
T3 tracking        & 3 & 11 & SAM~2~Long~\cite{zhang2024sam2long} & \hqB{0.96} & SAM~3.1$^\ddag$~\cite{carion2025sam3} & \hq{0.72} & \hrange{0.50}{0.75} & int$\to$cln & ${<}.001$ \\\hg
T4 rel.\ pose      & 3 & 10 & DA3~\cite{lin2026da3} & \hqB{0.38} & MonST3R~\cite{zhang2024monst3r} & \hq{0.08} & \hrange{0.86}{0.97} & clu$\to$cln & ${>}.05$ (tier) \\\hg
T5 regular VQA     & 3 & 12 & Qwen3-VL 8B~\cite{yang2025qwen3vl} & \hqB{0.57} & Florence 2 0.23B$^\S$~\cite{xiao2024florence2} & \hq{0.12} & \hrange{0.62}{0.98} & clu$\to$cln & ${<}.001$ (most) \\\hg
T6 in-context VQA  & 3 & 12 & Qwen3-VL 8B~\cite{yang2025qwen3vl} & \hqB{0.23} & Florence 2 0.77B~\cite{xiao2024florence2} & \hq{0.01} & \hrange{0.65}{0.96} & clu$\to$cln & ${<}.001$ (most) \\
\rpxHBottom
\end{tabular*}
\end{table*}

% ── ICRA 2027 Tasks — 6 tasks, up to 12 models each ──
\section{Tasks \& Models}
\label{sec:tasks}

All six tasks are evaluated on the same 100 scenes (Tables~\ref{tab:t1}--\ref{tab:t3},~\ref{tab:vqa}; T4 results in Sec.~\ref{sec:experiments}).
A model in several tasks counts once toward the 49 models, even with per-task checkpoints (DA3); SAM~3.1 box and text prompts count as two. Task-specific configurations yield 65 evaluations across six tasks.
$\Phi$ and $\mathcal{J}$ are computed from each task's standard metrics, listed below.

\smallskip
\noindent\textbf{T1\,--\,T2: Image \& Video Depth.}\enspace
Both tasks yield metric depth from the D435 RGB stream (relative outputs: GT-fitted scale and shift), scored on D435 depth.
T1 evaluates individual frames with $K{=}4$ metrics (AbsRel, RMSE, SILog, $\delta_1$).
T2 evaluates the full phase clip (250 frames) and adds two temporal consistency metrics (TGM, TGSE), giving $K{=}6$.

\smallskip
\noindent\textbf{T3: Multi-Object Tracking.}\enspace
Eight semi-supervised trackers are initialized with ground-truth first-frame masks; SAM~3.1 receives bounding boxes; text-prompted SAM~3.1 and GSAM2 ($^\ddag$) receive only an open-vocabulary label.

\smallskip
\noindent\textbf{T4: Relative Camera Pose Estimation.}\enspace
Ten models estimate the relative pose between ${\sim}20$K fixed image pairs.
Each pair is scored via AUC@$\tau$, the area under the pose-error curve over the larger of rotation error and translation-direction error.
Because objects move during Interaction and pairwise pose assumes a static scene, T4 is scored on Clutter and Clean only ($P{=}2$).

\smallskip
\noindent\textbf{T5\,--\,T6: Regular \& In-Context VQA.}\enspace
Six VLM families at two sizes each (12 models) respond with a bounding box through their native grounding interface.
A missing or unparseable box counts as a complete miss (Acc\,$0$, Center\,$0$, GIoU at its floor of $-1$).
The $K{=}3$ metric vector is Acc@$0.5$ (detection), GIoU~\cite{rezatofighi2019giou} (spatial quality), and Center-in-GT (coarse localization).
T5 asks a model to localize an object from a text description built from its colour, material, or function (${\sim}12{,}000$ scored questions per model across 5 attribute types; spatial-reference questions are released but not scored).
T6 asks the same 5 attribute question types conditioned on a single reference image of the target object from the single-object stream rather than its text description (${\sim}6{,}000$ scored questions per model).

\smallskip
\noindent\textbf{Ego--exo evaluation.}\enspace
T3 and T5--T6 are additionally evaluated on the egocentric GoPro stream during Interaction, with ground-truth masks and tracklets for both viewpoints, enabling a direct comparison between the ego and exo cameras at $6.75\times$ the spatial resolution.

\smallskip
\noindent\textbf{Extensibility.}\enspace
The PyPI toolkit \href{https://pypi.org/project/rpx-benchmark/}{\texttt{rpx-benchmark}} provides task runners, metric evaluation, and adapter interfaces for additional models.
The \href{https://irvlutd.github.io/RPX/toolkit-docs/}{toolkit documentation} describes installation, reference evaluation workflows, and custom-model integration.

\section{Experiments}
\label{sec:experiments}

\begin{figure}[!b]
    \centering
    \rpxRoundedImage[width=\columnwidth]{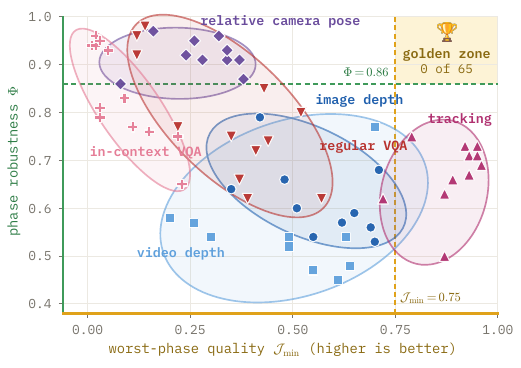}
    \caption{\textbf{No evaluation is both high-quality and phase-robust.} $\Phi$ vs.\ $\mathcal{J}_{\min}$ for all 65 evaluations; each ellipse is the smallest one enclosing a task's points. Relative camera pose (\textcolor[HTML]{6A4C9C}{$\blacklozenge$}) spans $\Phi = 0.86$--$0.97$; tracking (\textcolor[HTML]{B0306F}{$\blacktriangle$}) achieves high quality but degrades; depth and VQA span a broad range. Dashed: $\Phi = 0.86$ and an illustrative $\mathcal{J}_{\min} = 0.75$ bar; no evaluation reaches the golden zone.}
    \label{fig:phi-jmin}
\end{figure}

All 49 models are evaluated zero-shot on 100 scenes across three phases (two for T4), with the protocols of Sec.~\ref{sec:tasks}.
Table~\ref{tab:summary} summarizes the top and bottom model per task; Fig.~\ref{fig:phi-jmin} plots all 65 evaluations in the $\Phi$--$\mathcal{J}_{\min}$ plane.
Scene-level analyses cover 100 scenes; T4 significance is reported within the 33/33/34 tiers.

% ══════════════════════════════════════════════════
% Main summary table
% ══════════════════════════════════════════════════
% ══════════════════════════════════════════════════
% T1: Image Depth
% ══════════════════════════════════════════════════

\begin{table*}[t]
\centering
\caption{\textbf{Image depth (T1): no model is both high-quality and phase-robust.} Ranked by $\mathcal{J}_{\min}$; shade encodes value on a fixed $[0,1]$ scale (darker is higher): \rpxHSwatchQ\,quality $\mathcal{J}$, \rpxHSwatchR\,robustness $\Phi$. $K{=}4$: AbsRel, RMSE, SILog, $\delta_1$. Out: D depth, P pose, N normals, GS Gaussian splats. Columns follow Fig.~\ref{fig:metric-pipeline}; bold $\Phi$: best non-$\ddagger$ model. $^\ddagger$Relative depth, affine-aligned per phase, which can inflate $\Phi$.}
\label{tab:t1}
\rpxHeatStyle{6.3}{1.2pt}%
\vspace{2pt}
\begin{tabular*}{\textwidth}{@{\extracolsep{\fill}}r l l l r l l c c c c c c c c c c@{}}
\rpxHTop
\multicolumn{5}{@{}l}{\rule{0pt}{9pt}\rpxHGroup{Model}} & \multicolumn{2}{l}{\rpxHGroup{Verdict}} & \multicolumn{4}{c}{\rpxHGroup{Quality $\mathcal{J}$ by phase}} & \multicolumn{3}{c}{\rpxHGroup{Transition $\Phi$}} & \multicolumn{3}{c}{\rpxHGroup{Scene-tier $\Phi$}} \\
\rpxHGroupRules{\cmidrule(r){1-5}\cmidrule(lr){6-7}\cmidrule(lr){8-11}\cmidrule(lr){12-14}\cmidrule(l){15-17}}
\rpxHHead{\#} & \rpxHHead{name} & \rpxHHead{venue} & \rpxHHead{out} & \rpxHHead{params} & \rpxHHead{$\mathcal{J}_{\min}$} & \rpxHHead{$\Phi$} & \rpxHHead{all} & \rpxHHead{clu} & \rpxHHead{int} & \rpxHHead{cln} & \rpxHHead{clu$\leftrightarrow$int} & \rpxHHead{int$\leftrightarrow$cln} & \rpxHHead{clu$\leftrightarrow$cln} & \rpxHHead{E} & \rpxHHead{M} & \rpxHHead{H} \\[1pt]
\rpxHMid
\hk{1} & Metric3D V2~\cite{hu2024metric3dv2} & TPAMI'24 & D,N & 1.38~B & \hvq[1]{0.71} & \hvr{0.68} & \hqB{0.72} & \hq{0.72} & \hqB{0.72} & \hq{0.71} & \hr{0.44} & \hr{0.50} & \hr{0.56} & \hr{0.63} & \hr{0.58} & \hr{0.57} \\\hg
\hk{2} & DA3 Metric-L~\cite{lin2026da3} & ICLR'26 & D,P,N,GS & 0.33~B & \hvq{0.70} & \hvr{0.53} & \hqB{0.72} & \hqB{0.73} & \hq{0.70} & \hqB{0.73} & \hr{0.23} & \hr{0.28} & \hr{0.43} & \hr{0.49} & \hr{0.54} & \hr{0.46} \\\hg
\hk{3} & UniDepth V2~\cite{piccinelli2024unidepthv2} & TPAMI'26 & D & 0.35~B & \hvq{0.69} & \hvr{0.56} & \hq{0.71} & \hq{0.70} & \hq{0.69} & \hq{0.72} & \hr{0.31} & \hr{0.30} & \hr{0.53} & \hr{0.56} & \hr{0.55} & \hr{0.48} \\\hg
\hk{4} & FE2E$^\ddagger$~\cite{wang2025fe2e} & CVPR'26 & D & 12.51~B & \hvq{0.65} & \hvr{0.59} & \hq{0.68} & \hq{0.70} & \hq{0.65} & \hq{0.68} & \hr{0.39} & \hr{0.41} & \hr{0.49} & \hr{0.62} & \hr{0.55} & \hr{0.49} \\\hg
\hk{5} & Depth Pro~\cite{bochkovskii2024depthpro} & ICLR'25 & D & 0.95~B & \hvq{0.62} & \hvr{0.57} & \hq{0.63} & \hq{0.62} & \hq{0.64} & \hq{0.63} & \hr{0.30} & \hr{0.34} & \hr{0.56} & \hr{0.58} & \hr{0.54} & \hr{0.54} \\\hg
\hk{6} & DA-V2 Large~\cite{yang2024depthv2} & NeurIPS'24 & D & 0.34~B & \hvq{0.55} & \hvr{0.54} & \hq{0.60} & \hq{0.60} & \hq{0.55} & \hq{0.66} & \hr{0.27} & \hr{0.33} & \hr{0.43} & \hr{0.59} & \hr{0.49} & \hr{0.51} \\\hg
\hk{7} & ZipDepth$^\ddagger$~\cite{li2026zipdepth} & ECCV'26 & D & 6.14~M & \hvq{0.51} & \hvr{0.60} & \hq{0.56} & \hq{0.58} & \hq{0.51} & \hq{0.60} & \hr{0.42} & \hr{0.37} & \hr{0.54} & \hr{0.63} & \hr{0.59} & \hr{0.46} \\\hg
\hk{8} & MoGe-2 ViT-L~\cite{wang2025moge} & NeurIPS'25 & D,P & 0.33~B & \hvq{0.48} & \hvr{0.66} & \hq{0.52} & \hq{0.48} & \hq{0.54} & \hq{0.53} & \hr{0.43} & \hrB{0.51} & \hr{0.51} & \hr{0.66} & \hr{0.66} & \hr{0.61} \\\hg
\hk{9} & HyDen~\cite{zhang2026hyden} & ICLR'26 & D & 0.37~B & \hvq{0.42} & \hvr[1]{0.79} & \hq{0.44} & \hq{0.42} & \hq{0.45} & \hq{0.45} & \hrB{0.73} & \hr{0.50} & \hrB{0.64} & \hrB{0.76} & \hrB{0.78} & \hrB{0.70} \\\hg
\hk{10} & Lotus-2$^\ddagger$~\cite{he2024lotus} & arXiv'25 & D & 17.59~B & \hvq{0.35} & \hvr{0.64} & \hq{0.36} & \hq{0.35} & \hq{0.38} & \hq{0.35} & \hr{0.34} & \hr{0.66} & \hr{0.44} & \hr{0.64} & \hr{0.61} & \hr{0.55} \\
\rpxHBottom
\end{tabular*}
\end{table*}
\subsection{T1: Image Depth}
All ten models show a significant phase effect ($\Phi \leq 0.79$, $p < 0.001$; Table~\ref{tab:t1}).
Metric3D~V2 has the best worst-phase quality ($\mathcal{J}_{\min} = 0.71$, $\Phi = 0.68$).
DA3 Metric-L is close in quality ($\mathcal{J}_{\min} = 0.70$) but the least robust ($\Phi = 0.53$); both of its transitions into and out of Interaction are weak ($\Phi = 0.23$, $0.28$).
HyDen is the most robust ($\Phi = 0.79$) but scores much lower ($\mathcal{J}_{\min} = 0.42$).
For seven of the ten models, Clutter--Interaction is the weakest transition.
Among relative-depth models, only FE2E ($\mathcal{J}_{\min} = 0.65$) nears the best metric models, and the phase effects stay significant despite per-phase affine alignment.

\begin{table*}[t]
\centering
\caption{\textbf{Video depth (T2): VGGT leads on both quality and robustness.} $K{=}6$: AbsRel, RMSE, SILog, $\delta_1$, TGM, TGSE. Same diagnostic format as Table~\ref{tab:t1}. Full $250$-frame video per capture; relative output, GT scale-shift per clip.}
\label{tab:t2}
\rpxHeatStyle{6.3}{1.2pt}%
\vspace{2pt}
\begin{tabular*}{\textwidth}{@{\extracolsep{\fill}}r l l l r l l c c c c c c c c c c@{}}
\rpxHTop
\multicolumn{5}{@{}l}{\rule{0pt}{9pt}\rpxHGroup{Model}} & \multicolumn{2}{l}{\rpxHGroup{Verdict}} & \multicolumn{4}{c}{\rpxHGroup{Quality $\mathcal{J}$ by phase}} & \multicolumn{3}{c}{\rpxHGroup{Transition $\Phi$}} & \multicolumn{3}{c}{\rpxHGroup{Scene-tier $\Phi$}} \\
\rpxHGroupRules{\cmidrule(r){1-5}\cmidrule(lr){6-7}\cmidrule(lr){8-11}\cmidrule(lr){12-14}\cmidrule(l){15-17}}
\rpxHHead{\#} & \rpxHHead{name} & \rpxHHead{venue} & \rpxHHead{out} & \rpxHHead{params} & \rpxHHead{$\mathcal{J}_{\min}$} & \rpxHHead{$\Phi$} & \rpxHHead{all} & \rpxHHead{clu} & \rpxHHead{int} & \rpxHHead{cln} & \rpxHHead{clu$\leftrightarrow$int} & \rpxHHead{int$\leftrightarrow$cln} & \rpxHHead{clu$\leftrightarrow$cln} & \rpxHHead{E} & \rpxHHead{M} & \rpxHHead{H} \\[1pt]
\rpxHMid
\hk{1} & VGGT~\cite{wang2024vggt} & CVPR'25 & D,P,N & 1.26~B & \hvq[1]{0.70} & \hvr[1]{0.77} & \hqB{0.72} & \hq{0.70} & \hqB{0.73} & \hq{0.72} & \hrB{0.63} & \hr{0.58} & \hrB{0.64} & \hrB{0.70} & \hrB{0.65} & \hrB{0.74} \\\hg
\hk{2} & DA3 (video)~\cite{lin2026da3} & ICLR'26 & D,P,N,GS & 0.41~B & \hvq{0.64} & \hvr{0.48} & \hqB{0.72} & \hqB{0.78} & \hq{0.64} & \hqB{0.74} & \hr{0.23} & \hr{0.26} & \hr{0.46} & \hr{0.46} & \hr{0.42} & \hr{0.37} \\\hg
\hk{3} & ViGeo~\cite{vigeo2025} & arXiv'26 & D & 1.26~B & \hvq{0.63} & \hvr{0.54} & \hq{0.64} & \hq{0.63} & \hq{0.63} & \hq{0.67} & \hr{0.23} & \hr{0.42} & \hr{0.35} & \hr{0.53} & \hr{0.54} & \hr{0.48} \\\hg
\hk{4} & GemDepth~\cite{gemdepth2025} & ICML'26 & D & 0.71~B & \hvq{0.61} & \hvr{0.45} & \hq{0.69} & \hq{0.76} & \hq{0.61} & \hq{0.71} & \hr{0.17} & \hr{0.31} & \hr{0.34} & \hr{0.46} & \hr{0.39} & \hr{0.37} \\\hg
\hk{5} & DVD v1.1~\cite{dvd2024} & arXiv'26 & D & 2.25~B & \hvq{0.55} & \hvr{0.47} & \hq{0.62} & \hq{0.69} & \hq{0.55} & \hq{0.62} & \hr{0.19} & \hr{0.31} & \hr{0.39} & \hr{0.42} & \hr{0.45} & \hr{0.35} \\\hg
\hk{6} & ChronoDepth~\cite{shao2024chronodepth} & CVPR'25 & D & 2.25~B & \hvq{0.49} & \hvr{0.54} & \hq{0.54} & \hq{0.59} & \hq{0.49} & \hq{0.55} & \hr{0.28} & \hr{0.49} & \hr{0.33} & \hr{0.54} & \hr{0.54} & \hr{0.42} \\\hg
\hk{6} & MonST3R~\cite{zhang2024monst3r} & ICLR'25 & D,P & 0.57~B & \hvq{0.49} & \hvr{0.52} & \hq{0.55} & \hq{0.58} & \hq{0.49} & \hq{0.57} & \hr{0.27} & \hr{0.37} & \hr{0.39} & \hr{0.49} & \hr{0.55} & \hr{0.38} \\\hg
\hk{8} & Video-DA~\cite{yang2025videodepthanything} & CVPR'25 & D & 0.38~B & \hvq{0.30} & \hvr{0.54} & \hq{0.33} & \hq{0.35} & \hq{0.35} & \hq{0.30} & \hr{0.21} & \hr{0.55} & \hr{0.32} & \hr{0.57} & \hr{0.49} & \hr{0.44} \\\hg
\hk{9} & DepthCrafter~\cite{hu2024depthcrafter} & CVPR'25 & D & 2.25~B & \hvq{0.26} & \hvr{0.57} & \hq{0.30} & \hq{0.31} & \hq{0.31} & \hq{0.26} & \hr{0.33} & \hr{0.54} & \hr{0.30} & \hr{0.60} & \hr{0.49} & \hr{0.47} \\\hg
\hk{10} & RollingDepth~\cite{rollingdepth2024} & CVPR'25 & D & 1.29~B & \hvq{0.20} & \hvr{0.58} & \hq{0.22} & \hq{0.23} & \hq{0.24} & \hq{0.20} & \hr{0.27} & \hrB{0.59} & \hr{0.34} & \hr{0.60} & \hr{0.54} & \hr{0.47} \\
\rpxHBottom
\end{tabular*}
\end{table*}
\subsection{T2: Video Depth}\label{sec:exp-t2}
Video depth is the one task where quality and robustness align: VGGT leads on both axes ($\mathcal{J}_{\min} = 0.70$, $\Phi = 0.77$; Table~\ref{tab:t2}).
DA3 matches VGGT's overall quality but not its phase stability; most of the gap is attributable to Interaction ($\Phi_{\mathrm{clu}\leftrightarrow\mathrm{int}} = 0.23$).
At the bottom of the ranking, Video-DA, DepthCrafter, and RollingDepth combine moderate phase stability with $\mathcal{J}_{\min} \leq 0.30$, offering neither quality nor robustness.
A result common to all ten models: near-range AbsRel at $0.3$--$1.0$\,m, the range in which a robot grasps, increases $2.0$--$4.4{\times}$ during Interaction vs.\ Clutter, where D435 depth may be less reliable.

\begin{table*}[t]
\centering
\caption{\textbf{Tracking (T3): quality is near ceiling; robustness separates the trackers.} $K{=}3$: HOTA, MOTA, IDSW. GT mask/box initialization; ego = GoPro, Interaction only. $^\ddag$Text-prompted (no GT masks). $\Phi_{\text{EE}}$: ego vs.\ exo-Interaction (camera, resolution and view change together). }
\label{tab:t3}
\rpxHeatStyle{5.9}{0.8pt}%
\vspace{2pt}
\begin{tabular*}{\textwidth}{@{\extracolsep{\fill}}r l l l r l l c c c c c c c c c c c c@{}}
\rpxHTop
\multicolumn{5}{@{}l}{\rule{0pt}{9pt}\rpxHGroup{Model}} & \multicolumn{2}{l}{\rpxHGroup{Verdict}} & \multicolumn{5}{c}{\rpxHGroup{Quality $\mathcal{J}$ by phase}} & \multicolumn{4}{c}{\rpxHGroup{Transition $\Phi$}} & \multicolumn{3}{c}{\rpxHGroup{Scene-tier $\Phi$}} \\
\rpxHGroupRules{\cmidrule(r){1-5}\cmidrule(lr){6-7}\cmidrule(lr){8-12}\cmidrule(lr){13-16}\cmidrule(l){17-19}}
\rpxHHead{\#} & \rpxHHead{name} & \rpxHHead{venue} & \rpxHHead{prompt} & \rpxHHead{params} & \rpxHHead{$\mathcal{J}_{\min}$} & \rpxHHead{$\Phi$} & \rpxHHead{all} & \rpxHHead{clu} & \rpxHHead{int} & \rpxHHead{cln} & \rpxHHead{ego} & \rpxHHead{clu$\leftrightarrow$int} & \rpxHHead{int$\leftrightarrow$cln} & \rpxHHead{clu$\leftrightarrow$cln} & \rpxHHead{$\Phi_{\text{EE}}$} & \rpxHHead{E} & \rpxHHead{M} & \rpxHHead{H} \\[1pt]
\rpxHMid
\hk{1} & SAM~2 Long~\cite{zhang2024sam2long} & ICCV'25 & mask & 224~M & \hvq[1]{0.96} & \hvr{0.69} & \hqB{0.98} & \hqB{0.98} & \hqB{0.96} & \hqB{0.99} & \hqB{0.96} & \hr{0.42} & \hr{0.40} & \hr{0.94} & \hr{0.78} & \hr{0.71} & \hr{0.66} & \hr{0.60} \\\hg
\hk{2} & SAM~2~\cite{ravi2024sam2} & ICLR'25 & mask & 224~M & \hvq{0.95} & \hvr{0.73} & \hq{0.97} & \hqB{0.98} & \hq{0.95} & \hqB{0.99} & \hq{0.93} & \hr{0.46} & \hr{0.48} & \hr{0.96} & \hr{0.91} & \hr{0.74} & \hrB{0.79} & \hr{0.61} \\\hg
\hk{2} & DAM4SAM~\cite{videnovic2024dam4sam} & CVPR'25 & mask & 224~M & \hvq{0.95} & \hvr{0.71} & \hq{0.97} & \hqB{0.98} & \hq{0.95} & \hqB{0.99} & \hq{0.95} & \hr{0.46} & \hr{0.45} & \hr{0.93} & \hr{0.86} & \hr{0.73} & \hr{0.68} & \hr{0.62} \\\hg
\hk{4} & EdgeTAM~\cite{zhou2024edgetam} & CVPR'25 & mask & 14~M & \hvq{0.93} & \hvr{0.71} & \hq{0.96} & \hq{0.97} & \hq{0.93} & \hq{0.98} & \hq{0.92} & \hr{0.43} & \hr{0.45} & \hr{0.94} & \hrB{0.96} & \hr{0.67} & \hr{0.74} & \hr{0.65} \\\hg
\hk{4} & Cutie~\cite{cheng2024cutie} & CVPR'24 & mask & 35~M & \hvq{0.93} & \hvr{0.67} & \hq{0.96} & \hq{0.97} & \hq{0.93} & \hq{0.98} & \hq{0.94} & \hr{0.39} & \hr{0.36} & \hr{0.91} & \hr{0.88} & \hr{0.68} & \hr{0.60} & \hr{0.66} \\\hg
\hk{6} & SAM 2++ \cite{zhang2025sam2plusplus} & arXiv'25 & mask & 101~M & \hvq{0.92} & \hvr{0.73} & \hq{0.96} & \hqB{0.98} & \hq{0.92} & \hq{0.98} & \hq{0.86} & \hr{0.43} & \hr{0.48} & \hrB{0.98} & \hr{0.66} & \hr{0.70} & \hr{0.76} & \hr{0.66} \\\hg
\hk{7} & XMem~\cite{cheng2022xmem} & ECCV'22 & mask & 62~M & \hvq{0.89} & \hvr{0.66} & \hq{0.94} & \hq{0.96} & \hq{0.89} & \hq{0.98} & \hq{0.91} & \hr{0.42} & \hr{0.34} & \hr{0.88} & \hr{0.92} & \hr{0.65} & \hr{0.62} & \hr{0.65} \\\hg
\hk{8} & MiTS~\cite{xu2023mits} & ICCV'23 & mask & 33~M & \hvq{0.87} & \hvr{0.63} & \hq{0.93} & \hq{0.95} & \hq{0.87} & \hq{0.97} & \hq{0.86} & \hr{0.44} & \hr{0.31} & \hr{0.73} & \hr{0.80} & \hr{0.62} & \hr{0.67} & \hr{0.54} \\\hg
\hk{8} & SAM~3.1~\cite{carion2025sam3} & 2026 & bbox & 873~M & \hvq{0.87} & \hvr{0.50} & \hq{0.94} & \hq{0.97} & \hq{0.87} & \hq{0.97} & \hq{0.90} & \hr{0.21} & \hr{0.21} & \hr{0.93} & \hr{0.71} & \hr{0.53} & \hr{0.49} & \hr{0.41} \\
\rpxHMid
\hk{1} & GSAM2$^\ddag$~\cite{liu2023groundingdino,ren2024groundedsam} & 2024 & text & 397~M & \hvq{0.79} & \hvr[1]{0.75} & \hq{0.85} & \hq{0.87} & \hq{0.79} & \hq{0.90} & \hq{0.86} & \hrB{0.55} & \hrB{0.50} & \hr{0.93} & \hr{0.75} & \hrB{0.84} & \hr{0.72} & \hrB{0.67} \\\hg
\hk{2} & SAM~3.1$^\ddag$~\cite{carion2025sam3} & 2026 & text & 873~M & \hvq{0.72} & \hvr{0.62} & \hq{0.79} & \hq{0.83} & \hq{0.72} & \hq{0.82} & \hq{0.74} & \hr{0.29} & \hr{0.35} & \hr{0.96} & \hr{0.55} & \hr{0.66} & \hr{0.58} & \hr{0.52} \\
\rpxHBottom
\end{tabular*}
\end{table*}
\subsection{T3: Tracking}
Every tracker exhibits a significant phase effect ($\Phi \leq 0.75$, $p < 0.001$; Table~\ref{tab:t3}), and the degradation is sharply localized: transitions involving Interaction yield $\Phi = 0.21$--$0.55$, while Clutter$\leftrightarrow$Clean has weaker effects ($\Phi = 0.73$--$0.98$).
Because quality is uniformly high (even the weakest mask-initialized model reaches $\mathcal{J}_{\min} = 0.87$), it is $\Phi$ rather than $\mathcal{J}_{\min}$ that differentiates these models.
SAM~2~Long leads in quality ($\mathcal{J}_{\min} = 0.96$); GSAM2, a text-prompted tracker, reaches the highest $\Phi$ ($0.75$) but trades quality for phase stability ($\mathcal{J}_{\min} = 0.79$), and SAM~3.1 with text prompts has the lowest quality ($\mathcal{J}_{\min} = 0.72$).
SAM~3.1 with box prompts has the lowest $\Phi$ of any tracker ($0.50$).

\textbf{Ego--exo gap.}
Switching to the head-mounted camera barely changes tracking quality (SAM~2~Long and DAM4SAM keep $\mathcal{J} \geq 0.95$ on ego), yet 7 of 11 trackers show a large ego--exo effect ($\Phi_{\text{EE}} \leq 0.86$).
The switch thus shifts the joint HOTA, MOTA, and IDSW profile while quality stays high, reflecting the combined change in viewpoint, resolution, and sensor.

\begin{table}[t]
\centering
\caption{\textbf{Relative pose (T4): phase-robust but low quality.} $K{=}3$: AUC@$5^\circ$, @$10^\circ$, @$20^\circ$. Exocentric only (no GoPro pose ground truth); Clutter vs.\ Clean.}
\label{tab:t4}
\rpxHeatStyle{6.3}{1.2pt}%\setlength{\rpxHBarW}{16pt}%
\vspace{2pt}
\resizebox{\columnwidth}{!}{%
\begin{tabular}{@{}r l r l l c c c c c c c@{}}
\rpxHTop
\multicolumn{3}{@{}l}{\rule{0pt}{9pt}\rpxHGroup{Model}} & \multicolumn{2}{l}{\rpxHGroup{Verdict}} & \multicolumn{3}{c}{\rpxHGroup{Quality $\mathcal{J}$}} & \multicolumn{1}{c}{\rpxHGroup{$\Phi$}} & \multicolumn{3}{c}{\rpxHGroup{Tier $\Phi$}} \\
\rpxHGroupRules{\cmidrule(r){1-3}\cmidrule(lr){4-5}\cmidrule(lr){6-8}\cmidrule(lr){9-9}\cmidrule(l){10-12}}
\rpxHHead{\#} & \rpxHHead{name} & \rpxHHead{params} & \rpxHHead{$\mathcal{J}_{\min}$} & \rpxHHead{$\Phi$} & \rpxHHead{all} & \rpxHHead{clu} & \rpxHHead{cln} & \rpxHHead{clu$\leftrightarrow$cln} & \rpxHHead{E} & \rpxHHead{M} & \rpxHHead{H} \\[1pt]
\rpxHMid
\hk{1} & DA3~\cite{lin2026da3} & 1.36~B & \hvq[1]{0.38} & \hvr{0.87} & \hqB{0.42} & \hqB{0.38} & \hqB{0.45} & \hr{0.87} & \hr{0.87} & \hr{0.89} & \hr{0.86} \\\hg
\hk{2} & Reloc3R~\cite{reloc3r2025} & 0.43~B & \hvq{0.37} & \hvr{0.91} & \hq{0.40} & \hq{0.37} & \hq{0.42} & \hr{0.91} & \hr{0.92} & \hr{0.89} & \hr{0.91} \\\hg
\hk{3} & Pi3X~\cite{fan2025pi3x} & 0.99~B & \hvq{0.34} & \hvr{0.93} & \hq{0.38} & \hq{0.34} & \hq{0.41} & \hr{0.93} & \hr{0.93} & \hr{0.93} & \hr{0.92} \\\hg
\hk{3} & VGGT~\cite{wang2024vggt} & 1.26~B & \hvq{0.34} & \hvr{0.91} & \hq{0.37} & \hq{0.34} & \hq{0.41} & \hr{0.91} & \hr{0.96} & \hr{0.94} & \hr{0.85} \\\hg
\hk{5} & MASt3R~\cite{leroy2024mast3r} & 0.69~B & \hvq{0.32} & \hvr{0.96} & \hq{0.34} & \hq{0.32} & \hq{0.37} & \hr{0.96} & \hrB{0.97} & \hr{0.97} & \hr{0.93} \\\hg
\hk{6} & CUT3R~\cite{wang2025cut3r} & 0.79~B & \hvq{0.28} & \hvr{0.91} & \hq{0.31} & \hq{0.28} & \hq{0.34} & \hr{0.91} & \hr{0.95} & \hr{0.86} & \hr{0.92} \\\hg
\hk{7} & Fast3R~\cite{smart2025fast3r} & 0.65~B & \hvq{0.26} & \hvr{0.95} & \hq{0.28} & \hq{0.26} & \hq{0.30} & \hr{0.95} & \hr{0.95} & \hr{0.95} & \hr{0.95} \\\hg
\hk{8} & MUSt3R~\cite{duisterhof2025must3r} & 0.42~B & \hvq{0.24} & \hvr{0.92} & \hq{0.27} & \hq{0.24} & \hq{0.29} & \hr{0.92} & \hr{0.91} & \hr{0.93} & \hr{0.92} \\\hg
\hk{9} & DUSt3R~\cite{wang2024dust3r} & 0.57~B & \hvq{0.16} & \hvr[1]{0.97} & \hq{0.18} & \hq{0.16} & \hq{0.19} & \hrB{0.97} & \hr{0.95} & \hrB{0.99} & \hrB{0.96} \\\hg
\hk{10} & MonST3R~\cite{zhang2024monst3r} & 0.57~B & \hvq{0.08} & \hvr{0.86} & \hq{0.09} & \hq{0.08} & \hq{0.11} & \hr{0.86} & \hr{0.90} & \hr{0.81} & \hr{0.86} \\
\rpxHBottom
\end{tabular}}
\end{table}
\subsection{T4: Relative Camera Pose}
T4 compares relative pose between Clutter and Clean; Interaction is excluded (Sec.~\ref{sec:tasks}), so the results do not establish invariance during manipulation.
Unlike the other tasks, no model reaches statistical significance in any difficulty tier ($p > 0.05$ with 33--34 scenes per tier; $\Phi = 0.86$--$0.97$; Table~\ref{tab:t4}).
Models differ mainly in quality, which is low ($\mathcal{J}_{\min} = 0.08$--$0.38$, led by DA3); DUSt3R has the smallest estimated phase effect ($\Phi = 0.97$) but $\mathcal{J}_{\min} = 0.16$.
As a ground-truth check, the static AR-tag board of the 70 single-object scenes, placed in the world frame via the T265 poses, jitters by a median of $1.53\degree$\,/\,$4.68$\,cm, an upper bound that includes detection and sync error.

\begin{table*}[t]
\centering
\caption{\textbf{VQA (T5--T6): the most phase-stable models have the lowest quality.} Sizes paired; \#: $\mathcal{J}_{\min}$ rank in block. $K{=}3$: Acc@$0.5$, GIoU, Center-in-GT. Regular: 5 attribute question types, ${\sim}12{,}000$ scored questions per model; in-context: the same 5 types conditioned on a reference image instead of text, ${\sim}6{,}000$ questions. Counts include ego questions. Each block has its own per-scene MANOVA. $^\dag$MoE active parameters. $\Phi_{\text{EE}}$ as in Table~\ref{tab:t3}.}
\label{tab:vqa}
\rpxHeatStyle{6.3}{1.2pt}%
\vspace{2pt}
\begin{tabular*}{\textwidth}{@{\extracolsep{\fill}}r l r l l c c c c c c c c c c c c@{}}
\rpxHTop
\multicolumn{3}{@{}l}{\rule{0pt}{9pt}\rpxHGroup{Model}} & \multicolumn{2}{l}{\rpxHGroup{Verdict}} & \multicolumn{5}{c}{\rpxHGroup{Quality $\mathcal{J}$ by phase}} & \multicolumn{4}{c}{\rpxHGroup{Transition $\Phi$}} & \multicolumn{3}{c}{\rpxHGroup{Scene-tier $\Phi$}} \\
\rpxHGroupRules{\cmidrule(r){1-3}\cmidrule(lr){4-5}\cmidrule(lr){6-10}\cmidrule(lr){11-14}\cmidrule(l){15-17}}
\rpxHHead{\#} & \rpxHHead{name} & \rpxHHead{params} & \rpxHHead{$\mathcal{J}_{\min}$} & \rpxHHead{$\Phi$} & \rpxHHead{all} & \rpxHHead{clu} & \rpxHHead{int} & \rpxHHead{cln} & \rpxHHead{ego} & \rpxHHead{clu$\leftrightarrow$int} & \rpxHHead{int$\leftrightarrow$cln} & \rpxHHead{clu$\leftrightarrow$cln} & \rpxHHead{$\Phi_{\text{EE}}$} & \rpxHHead{E} & \rpxHHead{M} & \rpxHHead{H} \\[1pt]
\rpxHMid
\multicolumn{17}{@{}l}{\cellcolor{rpxHTrack}\rule[-2pt]{0pt}{9.5pt}\,\rpxHSection{T5: Regular grounding}} \\
\hk{1} & \lalibaba Qwen3-VL~\cite{yang2025qwen3vl} & 8~B & \hvq[1]{0.57} & \hvr{0.62} & \hqB{0.61} & \hqB{0.63} & \hqB{0.57} & \hqB{0.61} & \hqB{0.69} & \hr{0.31} & \hr{0.80} & \hr{0.35} & \hr{0.48} & \hr{0.55} & \hr{0.56} & \hr{0.67} \\
\hk{4} & \lalibaba Qwen3-VL~\cite{yang2025qwen3vl} & 2~B & \hvq{0.43} & \hvr{0.85} & \hq{0.46} & \hq{0.43} & \hq{0.49} & \hq{0.45} & \hq{0.62} & \hr{0.85} & \hr{0.71} & \hr{0.62} & \hr{0.50} & \hr{0.74} & \hr{0.85} & \hr{0.84} \\\hg
\hk{2} & \lnvidia Cosmos Reason2~\cite{nvidia2026cosmosreason2} & 8~B & \hvq{0.52} & \hvr{0.80} & \hq{0.54} & \hq{0.55} & \hq{0.52} & \hq{0.54} & \hq{0.67} & \hr{0.75} & \hr{0.63} & \hr{0.58} & \hr{0.23} & \hr{0.70} & \hr{0.80} & \hr{0.80} \\
\hk{9} & \lnvidia Cosmos Reason2~\cite{nvidia2026cosmosreason2} & 2~B & \hvq{0.22} & \hvr{0.77} & \hq{0.25} & \hq{0.28} & \hq{0.26} & \hq{0.22} & \hq{0.38} & \hr{0.64} & \hr{0.86} & \hr{0.45} & \hr{0.48} & \hr{0.76} & \hr{0.72} & \hr{0.77} \\\hg
\hk{3} & \lopengvlab InternVL~3.5~\cite{zhu2025internvl3} & 14~B & \hvq{0.44} & \hvr{0.74} & \hq{0.48} & \hq{0.52} & \hq{0.44} & \hq{0.49} & \hq{0.55} & \hr{0.59} & \hr{0.74} & \hr{0.45} & \hr{0.53} & \hr{0.70} & \hr{0.70} & \hr{0.76} \\
\hk{6} & \lopengvlab InternVL~3.5~\cite{zhu2025internvl3} & 1~B & \hvq{0.39} & \hvr{0.62} & \hq{0.42} & \hq{0.45} & \hq{0.39} & \hq{0.43} & \hq{0.41} & \hr{0.43} & \hr{0.63} & \hr{0.32} & \hrB{0.71} & \hr{0.60} & \hr{0.60} & \hr{0.64} \\\hg
\hk{5} & \ldeepseek DeepSeek-VL2~\cite{wu2024deepseekvl2} & 4.5~B\textsuperscript{\dag} & \hvq{0.41} & \hvr{0.72} & \hq{0.48} & \hq{0.54} & \hq{0.41} & \hq{0.48} & \hq{0.53} & \hr{0.47} & \hr{0.64} & \hr{0.56} & \hr{0.26} & \hr{0.72} & \hr{0.66} & \hr{0.70} \\
\hk{11} & \ldeepseek DeepSeek-VL2~\cite{wu2024deepseekvl2} & 1.0~B\textsuperscript{\dag} & \hvq{0.12} & \hvr{0.92} & \hq{0.15} & \hq{0.19} & \hq{0.12} & \hq{0.14} & \hq{0.19} & \hr{0.77} & \hr{0.98} & \hr{0.82} & \hr{0.44} & \hr{0.90} & \hrB{0.94} & \hr{0.87} \\\hg
\hk{8} & \lgoogle PaliGemma~2~\cite{beyer2024paligemma2} & 10~B & \hvq{0.35} & \hvr{0.75} & \hq{0.38} & \hq{0.43} & \hq{0.35} & \hq{0.37} & \hq{0.42} & \hr{0.49} & \hr{0.87} & \hr{0.50} & \hr{0.60} & \hr{0.69} & \hr{0.77} & \hr{0.76} \\
\hk{7} & \lgoogle PaliGemma~2~\cite{beyer2024paligemma2} & 3~B & \hvq{0.37} & \hvr{0.66} & \hq{0.39} & \hq{0.42} & \hq{0.37} & \hq{0.39} & \hq{0.44} & \hr{0.50} & \hr{0.66} & \hr{0.33} & \hr{0.67} & \hr{0.66} & \hr{0.64} & \hr{0.66} \\\hg
\hk{10} & \lmicrosoft Florence~2~\cite{xiao2024florence2} & 0.77~B & \hvq{0.14} & \hvr[1]{0.98} & \hq{0.15} & \hq{0.16} & \hq{0.15} & \hq{0.14} & \hq{0.16} & \hrB{0.99} & \hr{0.95} & \hrB{0.95} & \hr{0.65} & \hr{0.96} & \hrB{0.94} & \hrB{0.98} \\
\hk{11} & \lmicrosoft Florence~2~\cite{xiao2024florence2} & 0.23~B & \hvq{0.12} & \hvr{0.96} & \hq{0.13} & \hq{0.14} & \hq{0.12} & \hq{0.12} & \hq{0.16} & \hr{0.91} & \hrB{0.99} & \hr{0.90} & \hr{0.69} & \hrB{0.98} & \hr{0.92} & \hr{0.92} \\
\rpxHBarrier
\multicolumn{17}{@{}l}{\cellcolor{rpxHTrack}\rule[-2pt]{0pt}{9.5pt}\,\rpxHSection{T6: In-context grounding}} \\
\hk{1} & \lalibaba Qwen3-VL~\cite{yang2025qwen3vl} & 8~B & \hvq[1]{0.23} & \hvr{0.65} & \hqB{0.27} & \hqB{0.31} & \hqB{0.23} & \hqB{0.28} & \hq{0.34} & \hr{0.40} & \hr{0.74} & \hr{0.35} & \hr{0.63} & \hr{0.57} & \hr{0.70} & \hr{0.64} \\
\hk{2} & \lalibaba Qwen3-VL~\cite{yang2025qwen3vl} & 2~B & \hvq{0.22} & \hvr{0.75} & \hq{0.23} & \hq{0.25} & \hq{0.22} & \hq{0.22} & \hqB{0.35} & \hr{0.67} & \hr{0.76} & \hr{0.37} & \hr{0.67} & \hr{0.73} & \hr{0.74} & \hr{0.73} \\\hg
\hk{3} & \lopengvlab InternVL~3.5~\cite{zhu2025internvl3} & 14~B & \hvq{0.15} & \hvr{0.76} & \hq{0.19} & \hq{0.24} & \hq{0.15} & \hq{0.17} & \hq{0.12} & \hr{0.58} & \hr{0.86} & \hr{0.46} & \hr{0.67} & \hr{0.68} & \hr{0.75} & \hr{0.79} \\
\hk{7} & \lopengvlab InternVL~3.5~\cite{zhu2025internvl3} & 1~B & \hvq{0.03} & \hvr{0.79} & \hq{0.05} & \hq{0.07} & \hq{0.03} & \hq{0.04} & \hq{0.02} & \hr{0.64} & \hr{0.82} & \hr{0.52} & \hr{0.61} & \hr{0.76} & \hr{0.83} & \hr{0.72} \\\hg
\hk{4} & \lnvidia Cosmos Reason2~\cite{nvidia2026cosmosreason2} & 8~B & \hvq{0.11} & \hvr{0.77} & \hq{0.13} & \hq{0.15} & \hq{0.11} & \hq{0.14} & \hq{0.33} & \hr{0.50} & \hr{0.96} & \hr{0.47} & \hr{0.29} & \hr{0.79} & \hr{0.73} & \hr{0.73} \\
\hk{5} & \lnvidia Cosmos Reason2~\cite{nvidia2026cosmosreason2} & 2~B & \hvq{0.09} & \hvr{0.83} & \hq{0.10} & \hq{0.12} & \hq{0.09} & \hq{0.09} & \hq{0.21} & \hr{0.66} & \hr{0.97} & \hr{0.51} & \hr{0.58} & \hr{0.72} & \hr{0.86} & \hr{0.83} \\\hg
\hk{6} & \ldeepseek DeepSeek-VL2~\cite{wu2024deepseekvl2} & 4.5~B\textsuperscript{\dag} & \hvq{0.05} & \hvr{0.93} & \hq{0.07} & \hq{0.10} & \hq{0.05} & \hq{0.05} & \hq{0.10} & \hr{0.83} & \hrB{1.00} & \hr{0.83} & \hr{0.75} & \hr{0.89} & \hrB{0.96} & \hr{0.88} \\
\hk{10} & \ldeepseek DeepSeek-VL2~\cite{wu2024deepseekvl2} & 1.0~B\textsuperscript{\dag} & \hvq{0.02} & \hvr{0.94} & \hq{0.02} & \hq{0.03} & \hq{0.02} & \hq{0.02} & \hq{0.01} & \hr{0.84} & \hr{0.99} & \hr{0.84} & \hrB{0.94} & \hr{0.92} & \hr{0.95} & \hr{0.87} \\\hg
\hk{7} & \lgoogle PaliGemma~2~\cite{beyer2024paligemma2} & 10~B & \hvq{0.03} & \hvr{0.95} & \hq{0.03} & \hq{0.04} & \hq{0.03} & \hq{0.03} & \hq{0.09} & \hrB{0.95} & \hr{0.93} & \hr{0.85} & \hr{0.49} & \hrB{0.95} & \hr{0.86} & \hrB{0.96} \\
\hk{7} & \lgoogle PaliGemma~2~\cite{beyer2024paligemma2} & 3~B & \hvq{0.03} & \hvr{0.81} & \hq{0.04} & \hq{0.04} & \hq{0.03} & \hq{0.05} & \hq{0.07} & \hr{0.53} & \hr{0.83} & \hr{0.73} & \hr{0.73} & \hr{0.76} & \hr{0.79} & \hr{0.84} \\\hg
\hk{12} & \lmicrosoft Florence~2~\cite{xiao2024florence2} & 0.77~B & \hvq{0.01} & \hvr{0.94} & \hq{0.01} & \hq{0.02} & \hq{0.01} & \hq{0.01} & \hq{0.02} & \hr{0.81} & \hr{0.95} & \hr{0.91} & \hr{0.88} & \hr{0.90} & \hr{0.91} & \hr{0.90} \\
\hk{10} & \lmicrosoft Florence~2~\cite{xiao2024florence2} & 0.23~B & \hvq{0.02} & \hvr[1]{0.96} & \hq{0.02} & \hq{0.02} & \hq{0.02} & \hq{0.02} & \hq{0.02} & \hr{0.87} & \hr{0.98} & \hrB{0.92} & \hr{0.72} & \hr{0.93} & \hr{0.89} & \hr{0.93} \\
\rpxHBottom
\end{tabular*}
\end{table*}
\subsection{T5--T6: Visual Question Answering}
Qwen3-VL~8B achieves the best worst-phase quality in both VQA modes: $\mathcal{J}_{\min} = 0.57$ for regular grounding and $0.23$ for in-context grounding (Table~\ref{tab:vqa}).
The switch from text to a reference image (a ${\sim}60\%$ drop for Qwen3-VL~8B) shifts the ranking: Qwen3-VL~2B rises from fourth to second, and Cosmos~Reason2~2B from ninth to fifth.
A cautionary pattern emerges in the $\Phi$ column: the most phase-stable models (including Florence~2, $\Phi = 0.94$--$0.98$) sit near the quality floor in every phase.
Eight of the 24 model--mode pairs do not reach $\alpha = 0.001$ (all with $\Phi \geq 0.92$), four of them Florence~2 variants; their high $\Phi$ must be interpreted alongside uniformly low quality.
Among models with meaningful quality, the best is also among the most fragile (Qwen3-VL~8B, $\Phi = 0.62$--$0.65$), reinforcing that $\mathcal{J}_{\min}$ should be inspected before $\Phi$.

\textbf{Ego--exo gap.}
On regular questions, all 12 VLMs score higher on the GoPro stream than on exo-Interaction (gains $+0.01$ to $+0.15$, largest for Cosmos~Reason2~8B: $\mathcal{J}=0.52{\to}0.67$, $\Phi_{\text{EE}} = 0.23$), reflecting the combined change in viewpoint, resolution, and sensor.
In-context grounding has mixed effects: at displayed precision, eight models improve, one is unchanged, and three decline. Cosmos~Reason2~8B triples its quality ($0.11{\to}0.33$).

\section{Discussion}
\label{sec:discussion}

\textbf{Phase effects concentrate at Interaction.}
Image depth, video depth, and tracking all exhibit significant phase effects ($\Phi \leq 0.79$, $\leq 0.77$, and $\leq 0.75$), concentrated at Interaction: Clutter$\to$Interaction is the weakest transition in 15 of 20 depth models, and for every tracker both Interaction transitions ($\Phi = 0.21$--$0.55$) are weaker than Clutter$\leftrightarrow$Clean ($0.73$--$0.98$).
At near range ($0.3$--$1.0$\,m), AbsRel rises $2.0$--$4.4{\times}$ during Interaction for all ten video-depth models (Sec.~\ref{sec:exp-t2}), though D435 depth may be less reliable there.
Relative pose stands apart: $\Phi \geq 0.86$, and no difficulty-tier test reaches significance (Clutter--Clean only).

\textbf{The highest-quality model is usually not the most phase-robust.}
In five of six tasks, the model with the best $\mathcal{J}_{\min}$ is not the most phase-robust (Metric3D~V2 vs.\ HyDen, SAM~2~Long vs.\ GSAM2, DA3 vs.\ DUSt3R, and Qwen3-VL~8B vs.\ Florence~2 in both VQA modes); only in video depth does VGGT lead both.
A high $\Phi$ can accompany uniformly poor predictions (Florence~2: $\Phi \geq 0.94$, $\mathcal{J}_{\min} \leq 0.14$), so neither diagnostic suffices alone.

\textbf{The same change affects tasks differently.}
Switching from the exocentric to the GoPro view raises regular-grounding quality for all 12 VLMs, while tracking quality rises for five trackers, falls for four, and is unchanged for two (displayed precision).
Both tasks use the same GoPro and exocentric Interaction streams, so the contrast is not due to different data.

\textbf{Leaders differ across tasks.}
VGGT leads video depth but ties for third in pose at displayed precision; Qwen3-VL~8B leads both VQA modes, yet the order behind it shifts when the prompt switches from text to a reference image.

\textbf{Prompt modality affects phase robustness.}
SAM~3.1 with boxes is the most phase-sensitive tracker ($\Phi = 0.50$) but reaches $\Phi = 0.62$ with text prompts, and the text-prompted GSAM2 is the most phase-robust ($\Phi = 0.75$); changing the grounding context also changes VQA rankings.
Reporting the prompt protocol alongside model identity is necessary for interpreting these scores.

\textbf{Valid output does not imply correct grounding.}
Seven of the twelve VLMs return a syntactically valid box on ${\geq}90\%$ of in-context attribute questions, yet the best Acc@$0.5$ is $0.215$; regular grounding reaches $0.593$.
Valid box output does not ensure correct localization.

\section{Conclusion and Limitations}
\label{sec:conclusion}

We presented \coolname{}, a real-world benchmark of 100 scenes captured with consumer RGB-D sensors before, during, and after manipulation, plus an egocentric view, with ground truth shared across six perception tasks and two diagnostics, phase robustness~$\Phi$ and worst-phase quality~$\mathcal{J}_{\min}$, that compare models across tasks and scene conditions.

Evaluating 49 models zero-shot reveals three results.
First, phase sensitivity differs by task: depth and tracking show significant phase effects ($p{<}0.001$), while relative pose shows no significant within-tier Clutter--Clean effect.
Second, model rankings change across tasks, VQA modes, and prompting strategies on the same scenes, motivating task- and protocol-specific model selection.
Third, phase-resolved evaluation reveals phase-specific failures, including $2.0$--$4.4{\times}$ near-range AbsRel during Interaction; none of the 65 evaluations reaches both $\mathcal{J}_{\min}{\geq}0.75$ and $\Phi{\geq}0.86$ under the chosen thresholds.

One limitation is the T265 visual-inertial pose ground truth, whose board audit (median error ${\leq}1.53\degree$\,/\,$4.68$\,cm) covers only the single-object scenes.
Future work will test whether $\Phi$ and $\mathcal{J}_{\min}$ correlate with manipulation success.

\vspace{-0.5em}
\section*{Acknowledgments}
This work was supported in part by the National Science Foundation (NSF) under Grant No.~2520553, and the NVIDIA Academic Grant Program Award.

% \IEEEtriggeratref removed: the paper has no slack for a forced column break; Overleaf's line breaking differs slightly from local builds
\bibliographystyle{IEEEtran}
% Keep the bibliography at 8 pt; slightly tighter leading prevents a page-9 spill.
\begingroup
\renewcommand{\footnotesize}{\fontsize{7.4}{7.8}\selectfont}
\bibliography{root}
\endgroup

\clearpage
\raggedbottom
\allowdisplaybreaks[1]  % multi-line equation groups may break between lines (fills columns)
\widowpenalty=300
\clubpenalty=300
% Shared chart palette: blue emphasis, teal comparisons, distinct task colors,
% and cool neutral backgrounds. Figure 3 is unchanged.
\definecolor{rpxFigureInk}{HTML}{253449}
\definecolor{rpxFigureBlue}{HTML}{3B65A7}
\definecolor{rpxFigureTeal}{HTML}{198B82}
\definecolor{rpxFigureAmber}{HTML}{CF9232}
\definecolor{rpxFigureCoral}{HTML}{D57B58}
\definecolor{rpxFigureRed}{HTML}{BF1A2F}
\definecolor{rpxFigureSlate}{HTML}{A8B6C4}
\definecolor{rpxFigureViolet}{HTML}{8567AA}
\colorlet{rpxFigureLilac}{rpxFigureSlate}
\definecolor{rpxFigurePaleLilac}{HTML}{F3F6FA}
\colorlet{rpxFigureGold}{rpxFigureAmber}
\definecolor{rpxFigurePaleGold}{HTML}{EDF3FB}
\definecolor{rpxFigureMuted}{HTML}{64748B}
\definecolor{rpxFigureGrid}{HTML}{E3E9EF}
\definecolor{tierE}{HTML}{2E7D46}  % tiers: green / gold / red, as in Fig. 1 and the tier charts
\definecolor{tierM}{HTML}{E0A21A}
\definecolor{tierH}{HTML}{B91C1C}
% Indoor / outdoor environment palette (Stanford cardinal + warm cream)
\definecolor{rpxOutdoor}{HTML}{8C1515}
\definecolor{rpxIndoor}{HTML}{F5E8CC}
\definecolor{rpxIndoorInk}{HTML}{6B5B3E}          % legible text on cream
\newcommand{\indoorlabel}{\raisebox{0.6pt}{\fcolorbox{rpxIndoorInk}{rpxIndoor}{\rule{0pt}{4pt}\rule{4pt}{0pt}}}}
\newcommand{\outdoorlabel}{\raisebox{0.6pt}{\fcolorbox{rpxOutdoor}{rpxOutdoor}{\rule{0pt}{4pt}\rule{4pt}{0pt}}}}
\newcommand{\indoorsc}[1]{\textcolor{rpxIndoorInk}{#1}}
\newcommand{\outdoorsc}[1]{\textcolor{rpxOutdoor}{#1}}
% table header band: full-width cardinal rule behind each header row, cream text
\definecolor{rpxHead}{HTML}{8C1515}      % Stanford Cardinal Red
\definecolor{rpxHeadText}{HTML}{F5E8CC}  % Stanford cream
\definecolor{rpxSubHead}{HTML}{F5E8CC}   % second header row: cream band
\definecolor{rpxSubHeadText}{HTML}{8C1515} % ... with cardinal text
\newlength\rpxHeadH\setlength\rpxHeadH{11.2pt}
\newcommand{\rpxHeadStrut}{\rule[-3.2pt]{0pt}{\rpxHeadH}}
% direction arrows and ticks in tables: green = higher is better / provided, red = lower is better
\definecolor{rpxGood}{HTML}{15803D}\definecolor{rpxBad}{HTML}{B91C1C}
\definecolor{rpxGoodLight}{HTML}{86EFAC}\definecolor{rpxBadLight}{HTML}{FCA5A5}  % tints legible on the cardinal header band
\newcommand{\rpxUp}{\textcolor{rpxGood}{$\uparrow$}}\newcommand{\rpxDown}{\textcolor{rpxBad}{$\downarrow$}}
\newcommand{\rpxUpH}{\textcolor{rpxGoodLight}{$\uparrow$}}\newcommand{\rpxDownH}{\textcolor{rpxBadLight}{$\downarrow$}}
\newcommand{\rpxTick}{\textcolor{rpxGood}{\checkmark}}
% column rules: faint between columns, stronger between column groups; colours follow the band they cross
\definecolor{rpxColBody}{HTML}{CBD5E1}\definecolor{rpxGrpBody}{HTML}{94A3B8}
\definecolor{rpxColHead}{HTML}{B35A5A}\definecolor{rpxGrpHead}{HTML}{F5E8CC}
\definecolor{rpxColSub}{HTML}{DCC7A3}\definecolor{rpxGrpSub}{HTML}{8C1515}
\gdef\rpxColCol{rpxColBody}\gdef\rpxGrpCol{rpxGrpBody}
\newcommand{\rpxColRule}{{\color{\rpxColCol}\vrule width 0.3pt}}
\newcommand{\rpxGroupRule}{{\color{\rpxGrpCol}\vrule width 0.5pt}}
\newcommand{\rpxBodyRules}{\gdef\rpxColCol{rpxColBody}\gdef\rpxGrpCol{rpxGrpBody}}
% full-width tables: widen every column padding equally to reach the target width,
% so column rules stay centred (replaces tabular* with \extracolsep{\fill})
\newcount\rpxfitpass
\NewEnviron{rpxfit}[3]{\rpxfitpass=0\relax
  \loop\setbox0=\hbox{\begin{tabular}{#3}\BODY\end{tabular}}%
    \dimen0=#1\relax\advance\dimen0 by -\wd0\divide\dimen0 by #2\relax
    \advance\tabcolsep by \dimen0\advance\rpxfitpass by 1\relax
  \ifnum\rpxfitpass<3\repeat
  \begin{tabular}{#3}\BODY\end{tabular}}
\newcommand{\rpxHeadBand}{\gdef\rpxColCol{rpxColHead}\gdef\rpxGrpCol{rpxGrpHead}{\color{rpxHead}\hrule height \rpxHeadH depth 0pt}\vskip-\rpxHeadH}
\newcommand{\rpxSubHeadBand}{\gdef\rpxColCol{rpxColSub}\gdef\rpxGrpCol{rpxGrpSub}{\color{rpxSubHead}\hrule height \rpxHeadH depth 0pt}\vskip-\rpxHeadH}
\makeatletter\newcommand{\rpxRuleSepOff}{\global\aboverulesep=0pt\global\belowrulesep=0pt\gdef\CT@arc@{\color{white}}\global\cmidrulewidth=0.9pt}
\newcommand{\rpxRuleSepOn}{\global\aboverulesep=0.4ex\global\belowrulesep=0.65ex\gdef\CT@arc@{}\global\cmidrulewidth=0.03em}\makeatother
% Appendix float spacing: native IEEEtran two-column floats only (no strips,
% section barriers keep figure families together.
\setlength{\textfloatsep}{9pt plus 2pt minus 2pt}
\setlength{\dbltextfloatsep}{14pt plus 2pt minus 2pt}
\setlength{\floatsep}{7pt plus 2pt minus 1pt}
\setlength{\dblfloatsep}{7pt plus 2pt minus 1pt}
\renewcommand{\arraystretch}{1.06}
% full-width float pages: stack from the top with a fixed gap (no stretched bands)
\makeatletter\setlength\@dblfptop{0pt}\setlength\@dblfpsep{14pt}\setlength\@dblfpbot{0pt plus 1fil}\makeatother
\renewcommand{\topfraction}{0.85}\renewcommand{\textfraction}{0.1}\renewcommand{\floatpagefraction}{0.8}
\renewcommand{\dbltopfraction}{0.9}
% Bold appendix subheadings; the main-paper heading definitions remain unchanged.
\makeatletter
\def\subsection{\@startsection{subsection}{2}{\z@}{2ex plus .5ex minus .5ex}%
  {.7ex plus .3ex}{\normalfont\normalsize\bfseries}}
\def\subsubsection{\@startsection{subsubsection}{3}{\z@}{1.5ex plus .3ex}%
  {.5ex plus .2ex}{\normalfont\normalsize\bfseries}}
\makeatother
% Keep each table-family introduction with its first native full-width float.
% The generated table files and their numerical contents are unchanged.
\makeatletter
\newcommand{\rpxTableContext}[2]{%
  \begingroup
  \renewenvironment{table*}[1][tp]{%
    \@dblfloat{table}[##1]\noindent#1\par\smallskip
  }{\end@dblfloat}%
  \input{#2}%
  \endgroup
}
\makeatother

% Figs. 8-9 -- representative MOS scenes. Usage:
%   \repscenes{<cell width>}{tierE/Easy/scene012, tierM/Medium/scene074, ...}
% Cells are 4:3 crops of original release frames with the released instance
% masks; mask colours are keyed to the scene's object identities.
\newcommand\repscenes[2]{%
\begin{tikzpicture}[x=1cm,y=1cm]
  \pgfmathsetmacro\cw{#1}\pgfmathsetmacro\chh{\cw*0.75}
  \def\gap{0.08}\def\lab{1.36}
  \foreach \h [count=\i from 0] in {Clutter,Interaction,Clean,Ego Interaction}{
    \node[anchor=base,inner sep=0pt,font=\footnotesize] at (\lab+\i*\cw+\i*\gap+\cw/2,.14) {\h};}
  \foreach \col/\tier/\sid [count=\r from 0] in {#2}{
    \pgfmathsetmacro\yy{-\r*(\chh+\gap)}
    \node[anchor=base east,inner sep=0pt,font=\small\bfseries,text=\col] at (\lab-.14,\yy-\chh/2+.06) {\tier};
    \node[anchor=base east,inner sep=0pt,font=\fontsize{7.5}{9}\selectfont,text=rpxSlate] at (\lab-.14,\yy-\chh/2-.34) {\sid};
    \foreach \v [count=\i from 0] in {clutter,interaction,clean,ego}{
      \pgfmathsetmacro\xx{\lab+\i*(\cw+\gap)}
      \node[anchor=north west,inner sep=0pt] at (\xx,\yy)
        {\rpxRoundedImage[width=\cw cm,height=\chh cm]{assets/images/appendix/cells/\sid_\v.jpg}};
      \draw[rpxAccent,line width=.35pt,rounded corners=\rpxImageRadius] (\xx,\yy) rectangle ++(\cw,-\chh);}
  }
\end{tikzpicture}}

% = appendix layout parameters (float placement and figure scale) =
\def\RepCellW{4.04}
\def\catfs{10}\def\catlead{12}
\def\catsep{\arrayrulecolor{rpxHair}\specialrule{.25pt}{3.6pt}{3.6pt}\arrayrulecolor{black}}
\newcommand\rawfont{\footnotesize\setlength{\tabcolsep}{2pt}\renewcommand{\arraystretch}{1.05}}
% = complete scene catalogue (Appendix D): cell size fixed so that inserting
% or replacing images never reflows the catalogue pages
\newlength\catW\setlength\catW{3.62cm}\newlength\catH\setlength\catH{2.715cm}
\newlength\catG\setlength\catG{0.1cm}\newlength\catL
\setlength\catL{\dimexpr\textwidth-4\catW-3\catG-0.2cm\relax}
\newcommand\catpending{{\setlength\fboxsep{0pt}\setlength\fboxrule{.35pt}\fcolorbox{rpxHair}{white}{\parbox[c][\dimexpr\catH-.7pt][c]{\dimexpr\catW-.7pt}{\centering\fontsize{7.5}{9}\selectfont\color{rpxRule}image pending}}}}
\newcommand\catcell[2]{\parbox[c]{\catW}{\IfFileExists{assets/images/appendix/catalogue/scene#1_#2.jpg}{\rpxRoundedImage[width=\catW,height=\catH]{assets/images/appendix/catalogue/scene#1_#2.jpg}}{\catpending}}}
\newcommand\catheader{\noindent\makebox[\catL][r]{\footnotesize Scene}\hspace{0.2cm}%
  \makebox[\catW]{\footnotesize Clutter}\hspace{\catG}\makebox[\catW]{\footnotesize Interaction}\hspace{\catG}%
  \makebox[\catW]{\footnotesize Clean}\hspace{\catG}\makebox[\catW]{\footnotesize Ego Interaction}\par\vspace{3pt}}
\newcommand\catrow[4]{\noindent\parbox[c]{\catL}{\raggedleft\small\textbf{Scene #1}\\[1pt]\footnotesize #2\\\textcolor{#3}{#4}}\hspace{0.2cm}%
  \catcell{#1}{0}\hspace{\catG}\catcell{#1}{1}\hspace{\catG}\catcell{#1}{2}\hspace{\catG}\catcell{#1}{ego}\par\vspace{\catRowSep}}
% row gap: \catG on 8-row pages; wider on 7-row pages so every catalogue page spans the same height
\newlength\catRowSep\setlength\catRowSep{\catG}
% keep a table caption with at least the first rows of its body
\makeatletter\newcommand\rawneed[1]{\par\begingroup\dimen@=#1\relax\dimen@ii=\pagegoal\advance\dimen@ii-\pagetotal
  \ifdim\dimen@>\dimen@ii\ifdim\dimen@ii>\z@\vfil\fi\penalty-10000\fi\endgroup}\makeatother
\def\rawsep{\arrayrulecolor{rpxHair}\specialrule{.25pt}{1.4pt}{1.4pt}\arrayrulecolor{black}}

\section*{Appendix}
\makeatletter
\def\theHsection{Appendix.\Alph{section}}\xdef\Hy@chapapp{appendix}
\setcounter{section}{0}\setcounter{subsection}{0}\setcounter{subsubsection}{0}
\def\thesection{\Alph{section}}\def\thesectiondis{\Alph{section}.}
\def\thesubsection{\thesection.\arabic{subsection}}\def\thesubsectiondis{\thesection.\arabic{subsection}}
\makeatother
\noindent This supplement documents the data, procedures, and complete results behind the main paper.
Appendix~\ref{sec:appendix-data} describes the dataset composition, capture protocol, sensor rig, capture sites, scenes, and object catalogue (main paper Sec.~\ref{sec:dataset}).
Appendix~\ref{sec:appendix-preprocessing} covers mask, tracklet, and egocentric annotation and the \coolname{} difficulty score that defines the Easy, Medium, and Hard tiers.
Appendix~\ref{sec:appendix-eval} defines every task metric and the VQA questions, and explains the MANOVA estimator and desirability windows behind $\Phi$ and $\mathcal{J}$ (Sec.~\ref{sec:eval-primitives}).
Appendix~\ref{sec:appendix-bench} reports per-model results and hardware and latency measurements (Sec.~\ref{sec:experiments}), and Appendix~\ref{sec:appendix-toolkit} explains how to add models and tasks.

\section{Data Collection}
\label{sec:appendix-data}

\subsection{Dataset composition}

\begin{figure}[!htb]
\centering
\vspace*{2.7pt}% level with the APPENDIX heading (the top-float slot on this page sits ~20pt above the text block)
\rpxRoundedImage[width=\columnwidth]{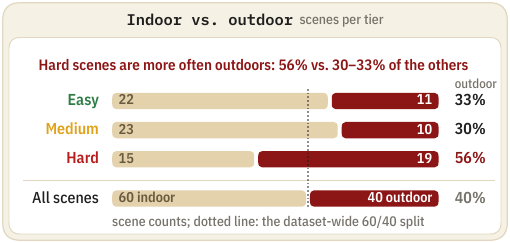}
\caption{Indoor and outdoor scenes in each RPX-DS difficulty tier: the outdoor share is 33\% of Easy, 30\% of Medium, and 56\% of Hard scenes; the full dataset is 60 indoor and 40 outdoor. Labels give scene counts.}
\label{fig:environment-tier-composition}
\end{figure}

\coolname{} contains 133,121 frames in three streams (Table~\ref{tab:appendix-dataset-summary}): multi-object exocentric captures (MOS) of 100 scenes in the Clutter, Interaction, and Clean phases; egocentric captures during Interaction; and single-object captures (SOS) of 70 objects.
Sixty scenes are indoor and 40 outdoor, recorded at 20 sites (Fig.~\ref{fig:appendix-capture-sites}); Fig.~\ref{fig:environment-tier-composition} gives their distribution over the difficulty tiers.

The question banks contain 4,529,322 single-image and 749,083 in-context question--frame records.
Evaluation uses fixed, seeded subsets of about 21,000 T5 and 10,500 T6 questions over Clutter, Interaction, Clean, and Ego Interaction; the published grounding scores use the attribute questions among them (about 12,000 and 6,000; Appendix~\ref{sec:appendix-qa}).
\begin{table}[!htb]
\centering
\caption{\coolname{} capture streams: frames, sensors, and annotations (\rpxTick{}: provided; --: not provided).}
\label{tab:appendix-dataset-summary}
\rpxHeatStyle{6.4}{2.4pt}%
\definecolor{dsMOS}{HTML}{4A1D5E}\definecolor{dsEgo}{HTML}{D25796}\definecolor{dsSOS}{HTML}{77736B}%
\newcommand{\dsSq}[1]{\tikz[baseline=-0.6ex]\fill[#1,rounded corners=0.6pt](0,-2.1pt)rectangle(4.2pt,2.1pt);\hspace{2.4pt}}%
\newcommand{\dsBand}[1]{\noalign{\vskip 1.0pt{\color{rpxHTrack}\hrule height 8.2pt depth 0pt}\vskip-8.2pt}\multicolumn{4}{@{\hspace{3pt}}l@{}}{\rule[-2.3pt]{0pt}{8.2pt}\rpxHSection{#1}}\\[-1.2pt]\noalign{\vskip 1.0pt}}%
\newcommand{\dsNo}{\textcolor{rpxHMute!55}{--}}%
\newcommand{\dsS}[1]{\hspace{2pt}{\color{rpxHMute}#1}}%
\renewcommand{\arraystretch}{1.08}%
\begin{rpxfit}{\columnwidth}{6}{@{\hspace{3pt}}l c c c@{\hspace{3pt}}}
\rpxHTop
\rule{0pt}{9pt} & \rpxHGroup{\dsSq{dsMOS}MOS} & \rpxHGroup{\dsSq{dsEgo}Ego} & \rpxHGroup{\dsSq{dsSOS}SOS} \\
 & \rpxHHead{100 scenes} & \rpxHHead{100 Interaction} & \rpxHHead{70 objects} \\[-1.4pt]
 & \rpxHHead{$\times$ 3 phases} & \rpxHHead{clips} & \rpxHHead{$\times$ 500 frames} \\
\rpxHMid
frames & 75,000\hspace{2pt}\rpxHBar{1.0}{dsMOS} & 23,121\hspace{2pt}\rpxHBar{0.30828}{dsEgo} & 35,000\hspace{2pt}\rpxHBar{0.46667}{dsSOS} \\
per capture & 250 & 195--250 & 500 \\
\dsBand{sensors}
RGB & \rpxTick & \rpxTick & \rpxTick \\
metric depth\dsS{D435} & \rpxTick & \dsNo & \rpxTick \\
fisheye stereo\dsS{T265} & \rpxTick & \dsNo & \rpxTick \\
camera pose\dsS{T265} & \rpxTick & \dsNo & \rpxTick \\
\dsBand{annotations}
instance masks & \rpxTick & \rpxTick & \rpxTick \\
tracklets & \rpxTick & \rpxTick & \dsNo \\
global object IDs & \rpxTick & \rpxTick & \rpxTick \\
grounding questions & \rpxTick & \rpxTick & \dsNo \\
questionnaire & \dsNo & \dsNo & \rpxTick \\
\rpxHMid
\textbf{total frames} & \multicolumn{3}{c}{\textbf{133,121}} \\
\rpxHBottom
\end{rpxfit}
\end{table}

The outdoor share is 33\% of Easy, 30\% of Medium, and 56\% of Hard scenes.
The tiers come from the \coolname{} difficulty score (RPX-DS; Appendix~\ref{sec:appendix-difficulty}), which does not use the environment label.

\subsection{Objects}
\label{sec:appendix-objects}\label{sec:appendix-sos}

\begin{figure*}[t]
\centering
\input{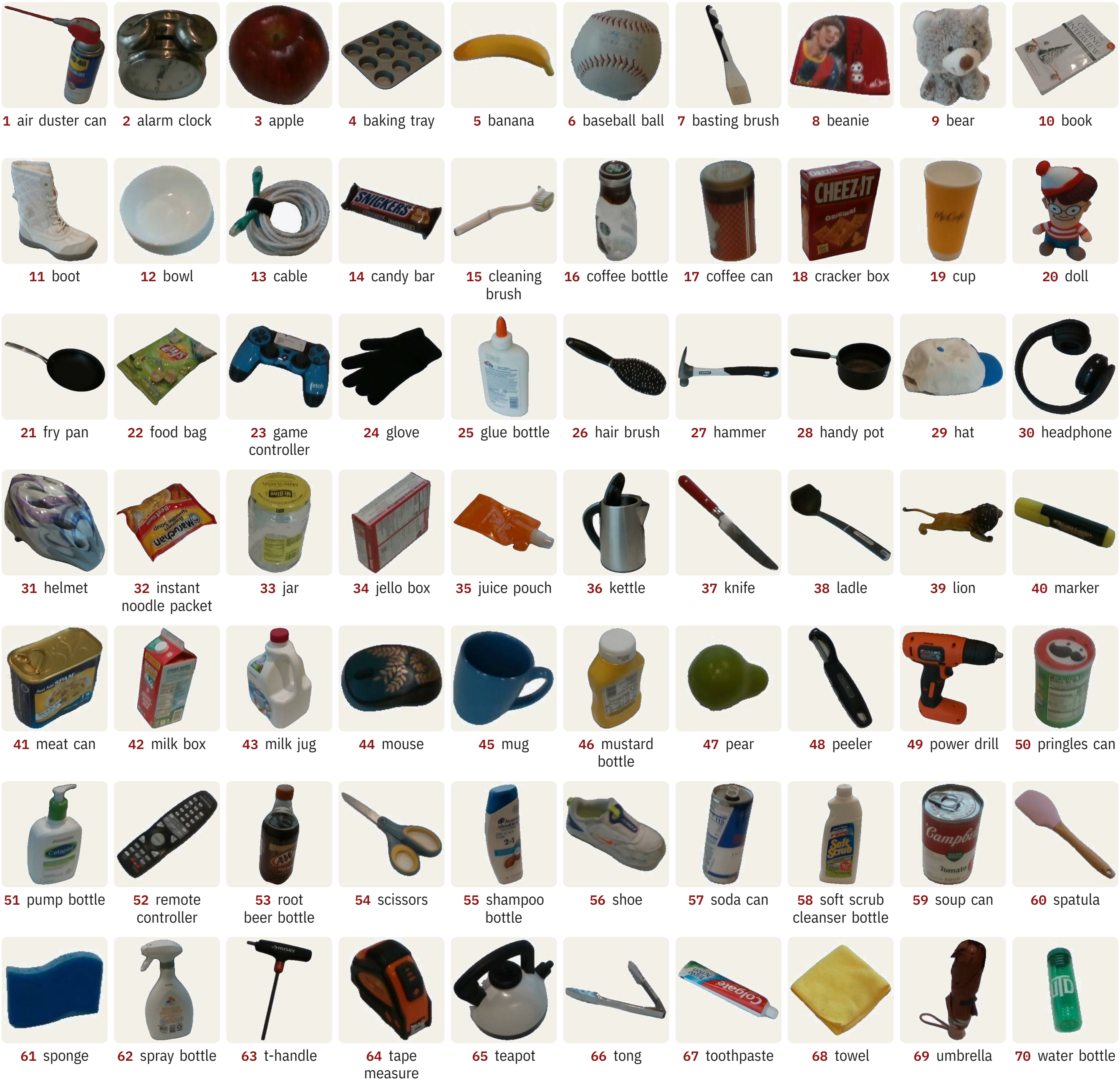}
\caption{All 70 SOS objects, each cut out of one released SOS frame with its instance mask and numbered as in Table~\ref{tab:sos-canonical}.}
\label{fig:sos-gallery}
\end{figure*}

\coolname{} uses 70 everyday objects (Fig.~\ref{fig:sos-gallery}).
Each is first recorded alone, in a single-object capture (SOS) of 500 frames (35,000 in total) while the hand-held rig orbits it; SOS images are the reference images for in-context grounding (T6).
Annotators also describe each object in a short questionnaire (Table~\ref{tab:sos-schema}), and Table~\ref{tab:sos-canonical} lists one canonical response per field for every object.

\begin{table}[!htb]
\centering
\caption{SOS questionnaire schema.}
\label{tab:sos-schema}
\rpxHeatStyle{5.6}{1.8pt}%
\renewcommand{\arraystretch}{1.18}%
\definecolor{sqGold}{HTML}{FBEFD5}\definecolor{sqGoldH}{HTML}{F6E2B4}%
\newcommand{\sqSection}[1]{\noalign{\vskip 1.4pt{\color{rpxHTrack}\hrule height 8.6pt depth 0pt}\vskip-8.6pt}\multicolumn{3}{@{\hspace{3pt}}l@{}}{\rule[-2.4pt]{0pt}{8.6pt}\rpxHSection{#1}}\\[-1.2pt]\noalign{\vskip 1.6pt}}%
\newcommand{\sqOff}[1]{\textcolor{rpxHMute}{#1}}%
\newcommand{\sqFoff}[1]{\textcolor{rpxHMute}{#1}}%
\newcommand{\sqF}[1]{\textcolor{rpxOutdoor}{#1}}%
\newcommand{\sqVR}{{\color{rpxHMute!40}\vrule width 0.4pt}}%
\vspace{2pt}
\begin{tabularx}{\columnwidth}{@{\hspace{3pt}}>{\bfseries}l@{\hspace{4pt}}!{\sqVR}@{\hspace{4pt}}>{\rmfamily\fontsize{7.2}{8.2}\selectfont\raggedright\arraybackslash}X@{\hspace{1pt}}>{\columncolor{sqGold}[1.8pt][2pt]\hspace{1.4pt}}l@{\hspace{2pt}}}
\rpxHTop
\multicolumn{1}{@{\hspace{3pt}}l@{\hspace{8.4pt}}}{\rule{0pt}{9pt}\rpxHHead{field}} & \multicolumn{1}{l@{\hspace{3pt}}}{\rpxHHead{annotator prompt}} & \cellcolor{sqGoldH}\rpxHHead{\color{rpxHInk}use in T5/T6} \\
\rpxHMid
\sqSection{catalogue labels}
\sqFoff{Name} & Names and valid aliases & \sqOff{Catalogue label only} \\
\sqFoff{Category} & Semantic object class & \sqOff{Catalogue label only} \\
\sqSection{question attributes}
\sqF{Colour} & Visible colour descriptions & Colour questions \\
\sqF{Material} & Constituent material & Material, composition, odd-one-out \\
\sqF{Function} & Common purpose or action & Function, composition \\
\rpxHBottom
\end{tabularx}
\end{table}

\subsection{Modalities captured}
\label{sec:appendix-capture}

\begin{figure*}[t]
\centering
\includegraphics[width=\textwidth]{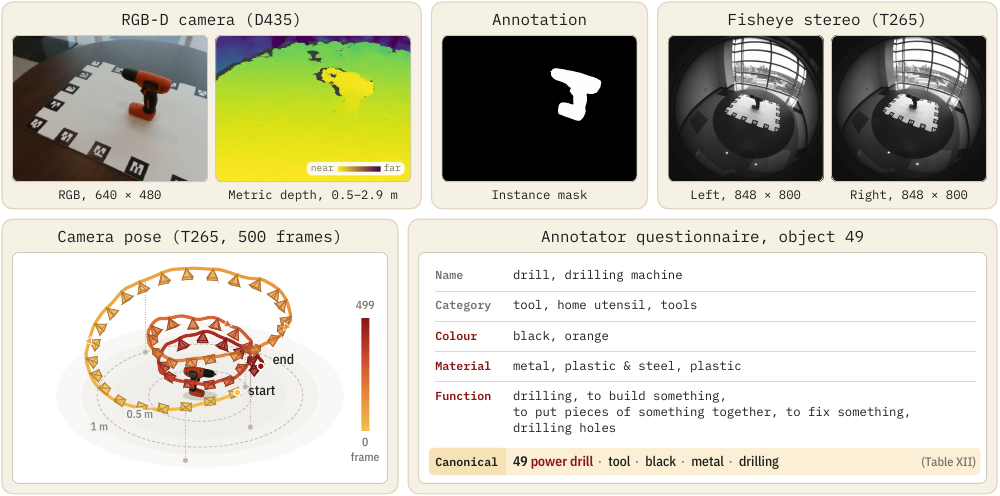}
\caption{SOS modalities for one frame (289 of 500) of object 49, power drill: RGB and metric depth (D435), the released instance mask, the calibrated T265 fisheye stereo pair, the T265 camera trajectory over all 500 frames (camera frustums coloured by frame, the object at the cameras' common look-at point, and floor rings at 0.5 m and 1 m), and the questionnaire responses behind the canonical entry. The exocentric MOS captures record the same RGB-D, mask, stereo, and pose modalities; the MOS figures show only the RGB phase images.}
\label{fig:sos-modalities}
\end{figure*}

One hand-held rig records every capture: a D435 for RGB and metric depth and a T265 for calibrated fisheye stereo and 6-DoF camera pose, with a head-mounted GoPro adding the egocentric view during Interaction (Secs.~\ref{sec:three-phase} and~\ref{sec:dataset}).
Figure~\ref{fig:sos-modalities} shows every modality for one SOS frame, together with the instance mask (Appendix~\ref{sec:appendix-preprocessing}) and the questionnaire.
The multi-object scenes record the same modalities in each exocentric phase (Table~\ref{tab:appendix-dataset-summary}); pose is recorded in all three exocentric phases, and relative-pose evaluation uses Clutter and Clean pairs only.
The exocentric (D435, T265) and egocentric (GoPro) streams run on independent clocks, and phase boundaries are set manually during recording.

The rig uses only commercially available parts (Table~\ref{tab:bom}).
\begin{table}[!htb]
\centering
\caption{Capture rig bill of materials. Prices are approximate retail at the time of capture (USD).}
\label{tab:bom}
\rpxHeatStyle{5.6}{1.8pt}%
\renewcommand{\arraystretch}{1.18}%
\definecolor{bmGold}{HTML}{FBEFD5}\definecolor{bmGoldH}{HTML}{F6E2B4}%
\newcommand{\bmSection}[1]{\noalign{\vskip 1.4pt{\color{rpxHTrack}\hrule height 8.6pt depth 0pt}\vskip-8.6pt}\multicolumn{3}{@{\hspace{3pt}}l@{}}{\rule[-2.4pt]{0pt}{8.6pt}\rpxHSection{#1}}\\[-1.2pt]\noalign{\vskip 1.6pt}}%
\newcommand{\bmVR}{{\color{rpxHMute!40}\vrule width 0.4pt}}%
\vspace{2pt}
\begin{tabularx}{\columnwidth}{@{\hspace{3pt}}>{\bfseries}l@{\hspace{4pt}}!{\bmVR}@{\hspace{4pt}}>{\rmfamily\fontsize{7.2}{8.2}\selectfont\raggedright\arraybackslash}X<{\rule[-4.2pt]{0pt}{0pt}}@{\hspace{7pt}}>{\columncolor{bmGold}[3pt][3pt]}r@{\hspace{3pt}}}
\rpxHTop
\multicolumn{1}{@{\hspace{3pt}}l@{\hspace{8.4pt}}}{\rule{0pt}{9pt}\rpxHHead{component}} & \multicolumn{1}{l@{\hspace{3pt}}}{\rpxHHead{role}} & \cellcolor{bmGoldH}\rpxHHead{\color{rpxHInk}price} \\
\rpxHMid
\bmSection{sensors}
Intel D435 & RGB-D ($640{\times}480$, 30\,fps; metric depth via active IR stereo) & \$180 \\
Intel T265 & 6-DoF VIO at up to 200\,Hz (30\,fps fisheye stereo, IMU-interpolated); two wide-FOV cameras. Discontinued 2021; secondary market & \$250--400 \\
GoPro & Egocentric RGB during Interaction, head-mounted (independent clock) & \$350 \\
\bmSection{mounting}
3D-printed bracket & Couples D435 and T265 using Intel's reference geometry & \$5 \\
Commercial clamp & Attaches assembly to laptop screen for hand-held use & \$15 \\
\rpxHMid
\multicolumn{2}{@{\hspace{3pt}}l}{\textbf{Total}} & \cellcolor{bmGoldH}\textbf{$\sim$\$1K} \\
\rpxHBottom
\end{tabularx}
\end{table}

Replication requires a T265 or another calibrated 6-DoF visual--inertial pose source that is time-aligned with the D435.
The bracket and the D435--T265 extrinsic follow Intel's reference design for this camera pair.
\label{sec:appendix-bom}

\subsection{Capture sites}
\label{sec:appendix-sites}

\begin{figure*}[p]
\centering
\rpxRoundedImage[width=\textwidth]{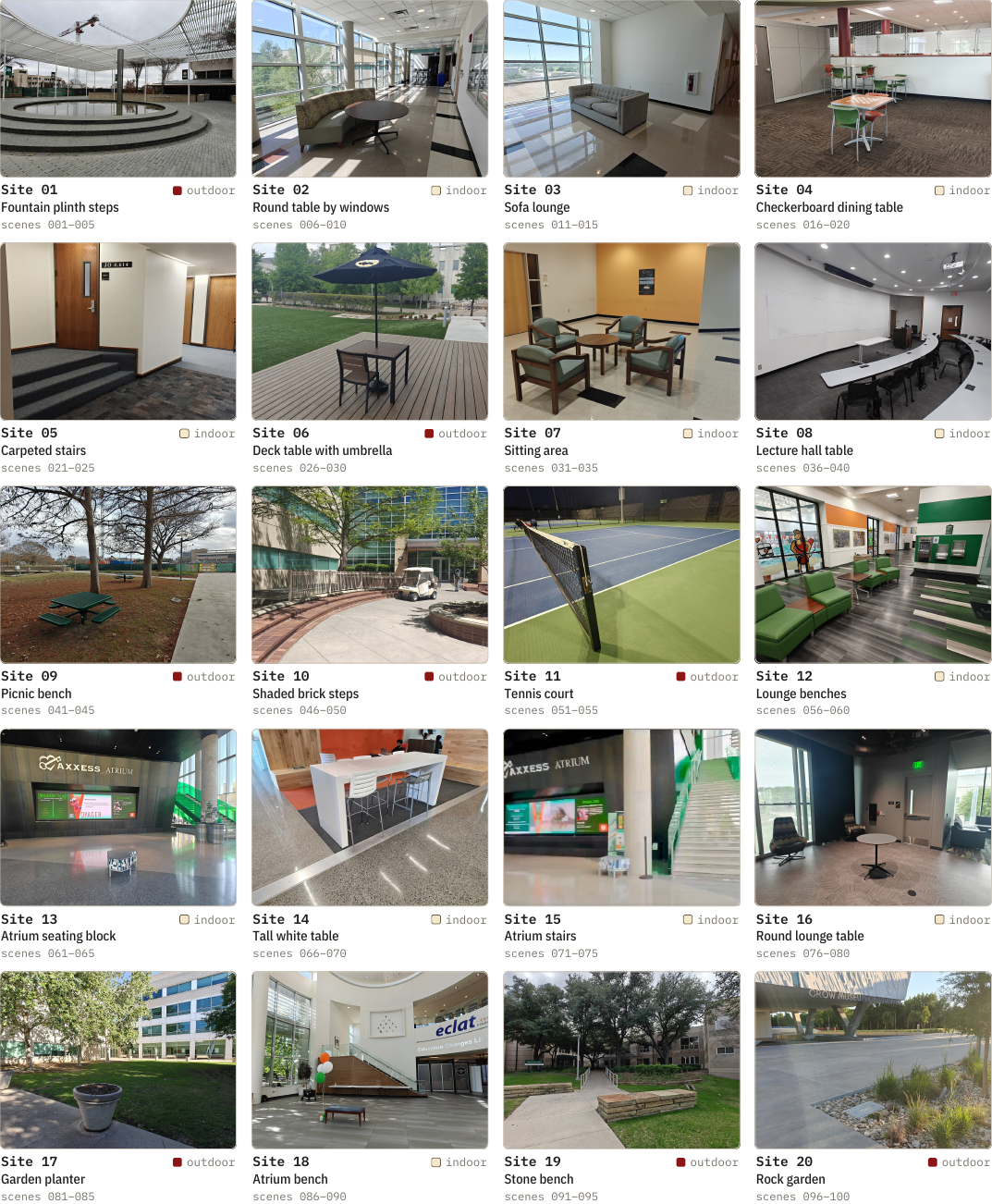}
\caption{The 20 capture sites, photographed without the scene objects. Each site hosts five consecutive MOS scenes.}
\label{fig:appendix-capture-sites}
\end{figure*}

The multi-object scenes were recorded at 20 sites, 12 indoor and 8 outdoor, with five consecutive scenes per site (Fig.~\ref{fig:appendix-capture-sites}).
Each site is photographed without the scene objects; the environment of a scene is that of its site.

\subsection{Indoor and outdoor scenes}
\label{sec:appendix-scenes}

\begin{figure*}[t]
\centering
\rpxRoundedImage[width=\textwidth]{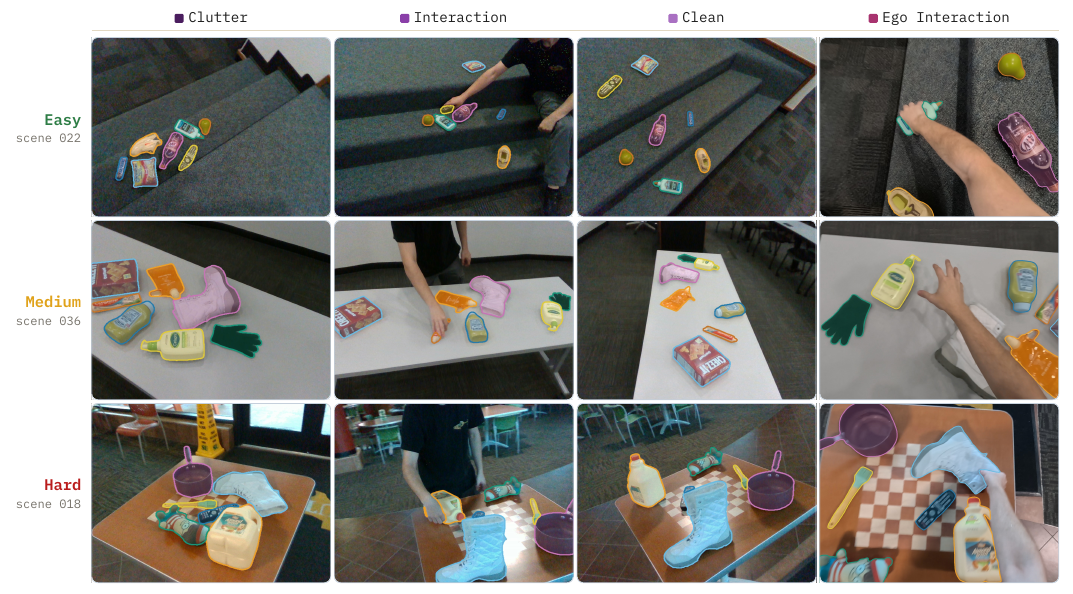}
\caption{Representative \indoorlabel\;indoor MOS scenes, one per difficulty tier, in the four capture views; overlays are the released instance masks, and each object keeps one colour across its views. Percentiles rank a scene's phase-averaged descriptor (Appendix~\ref{sec:appendix-difficulty}) among all 100 scenes. \textbf{Easy}, scene~022 (carpeted stairs; RPX-DS 0.29, 1st percentile): little object overlap (9th) and nearly complete depth (0.3\% of pixels invalid). \textbf{Medium}, scene~036 (lecture-hall table; RPX-DS 0.49, 47th): typical on most descriptors, with more dark regions lacking depth (84th). \textbf{Hard}, scene~018 (checkerboard dining table; RPX-DS 0.63, 92nd): a wide depth range (82nd), objects changing size in view (93rd), glare and dark regions lacking depth (79th, 88th), and more annotation correction rounds (86th).}
\label{fig:appendix-representative-indoor}
\end{figure*}

\begin{figure*}[t]
\centering
\rpxRoundedImage[width=\textwidth]{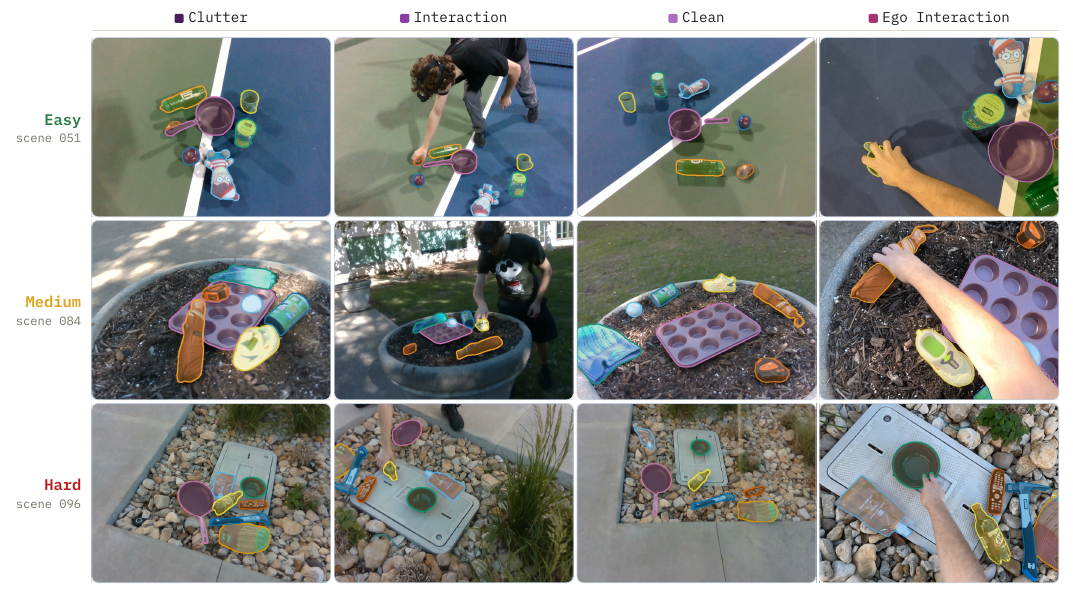}
\caption{Representative \outdoorlabel\;outdoor MOS scenes, one per difficulty tier, shown as in Fig.~\ref{fig:appendix-representative-indoor}. \textbf{Easy}, scene~051 (tennis court; RPX-DS 0.43, 21st percentile): little overlap and few depth gaps, but strong glare (70th) and fast camera motion (83rd). \textbf{Medium}, scene~084 (garden planter; RPX-DS 0.53, 61st): heavy object overlap (88th), a wide depth range (93rd), very strong glare (94th), and objects changing size in view (95th). \textbf{Hard}, scene~096 (rock garden; RPX-DS 0.60, 84th): high object overlap (74th) and among the most annotation correction rounds (99th).}
\label{fig:appendix-representative-outdoor}
\end{figure*}

At each site the 70 objects are combined into multi-object scenes (MOS) and captured with all of the modalities above in the Clutter, Interaction, and Clean phases, plus the egocentric view during Interaction.
Figures~\ref{fig:appendix-representative-indoor} and~\ref{fig:appendix-representative-outdoor} show one indoor and one outdoor scene per difficulty tier in all four views, with the released instance masks overlaid; each object keeps one colour across views (Appendix~\ref{sec:appendix-ids}).

The complete visual catalogue of all 100 scenes, one middle frame (index $\lfloor n/2\rfloor$ of $n$) per view, is available on the \href{\webpage}{project page}.
\label{fig:appendix-scene-catalogue}
% Full 100-scene visual catalogue available in the dataset release at \url{\webpage}
% Omitted from arXiv submission to meet the 50\,MB file-size limit.

Each scene contains six to eight SOS objects, seven in 98 of the 100 scenes.
The object set is the same in all three exocentric phases, and each of the 70 objects appears in 9--11 scenes (Table~\ref{tab:appendix-scene-objects}).

\rpxTableContext{Table~\ref{tab:appendix-scene-objects} lists the objects in every MOS scene by SOS catalogue ID.}{generated/scene_objects_table}

\subsection{Object identifiers}
\label{sec:appendix-ids}

Mask IDs are local to each MOS \texttt{(scene\_id, phase)} pair and to each Ego clip.
Each capture's \texttt{mask\_to\_object.json} maps a local ID to a catalogue ID, and the object catalogue maps that ID to a dataset-level \texttt{global\_object\_id}, shared with the SOS capture of the same object.
All 2,100 MOS masks resolve; in Ego, 639 of 699 masks resolve and 60 are marked unlabelled.
The global identity connects an object across phases, views, and its SOS capture; in-context grounding (T6) uses it to select SOS reference images.

\begingroup
\setlength{\smallskipamount}{1pt}
\rpxTableContext{Table~\ref{tab:sos-canonical} lists all 70 objects with their canonical responses.}{generated/sos_catalogue}
\endgroup

%BOUNDARY
\section{Data Preprocessing}
\label{sec:appendix-preprocessing}

\subsection{Mask and tracklet annotation}

\begin{figure*}[t]
\centering
\rpxRoundedImage[width=\textwidth]{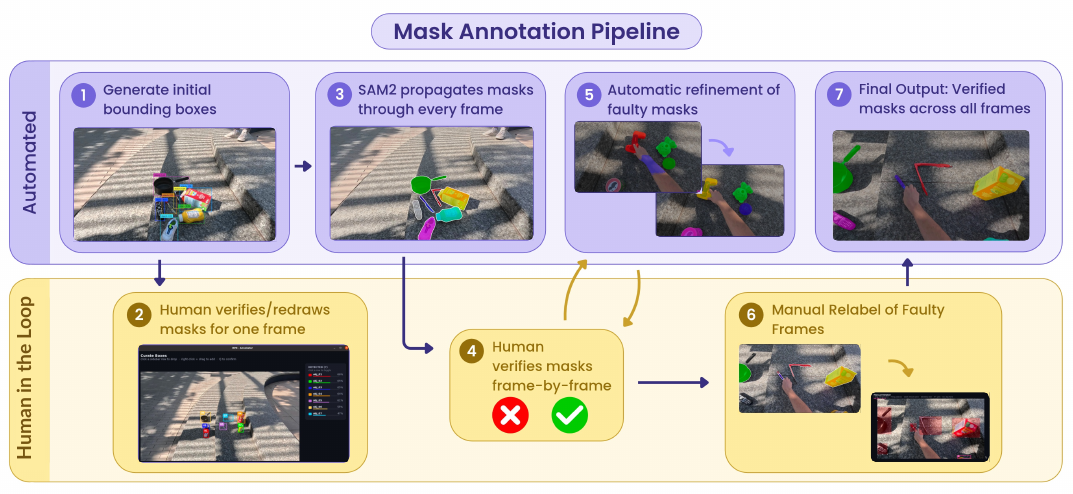}
\caption{Instance-mask annotation workflow using RoboKit~\cite{p2024robokit}. Grounded proposals initialize SAM2 masks; bidirectional propagation alternates with human correction and identity and continuity checks.}
\label{fig:appendix-mask-pipeline}
\end{figure*}

The annotation pipeline uses RoboKit~\cite{p2024robokit} to initialize SAM2 masks from grounded detections and propagate them bidirectionally through each sequence (Fig.~\ref{fig:appendix-mask-pipeline}).
Annotators correct missed, merged, or drifting instances with point or box prompts; propagation restarts at the corrected frame.
Quality control checks nonempty foreground, identity continuity, temporal area continuity, and the scene--phase object mapping.
Tracklets inherit the corrected persistent mask identities.

\subsection{Egocentric annotation}

The GoPro and D435 share no timecode.
Egocentric clips are sampled offline to at most 250 frames, including both endpoints; the release contains 195--250 frames per scene (23,121 in total), and \texttt{ego\_frame\_map.json} maps each sample to its source frame.
Masks follow the MOS pipeline.
Annotators give each egocentric instance the object identity of its exocentric counterpart, so identities match across cameras without pixel or temporal alignment.

\subsection{Difficulty descriptors}
\label{sec:appendix-difficulty}\label{sec:appendix-feature-defs}

Table~\ref{tab:appendix-esd-features} lists 27 difficulty descriptors, computed from the aligned frame sequence of each of the 300 exocentric scene--phase cells.
Two measure annotation effort; the other 25 cover scene complexity, occlusion, depth quality, photometric conflict, temporal stability, camera motion, and fisheye stereo.
The release includes all descriptor values.
Table~\ref{tab:appendix-feature-notation} gives the shared notation; group-specific symbols are introduced where needed.

\begin{table}[!htb]
\centering
\caption{Shared notation for the 27 difficulty descriptors.}
\label{tab:appendix-feature-notation}
\rpxHeatStyle{5.6}{1.8pt}%
\renewcommand{\arraystretch}{1.16}%
\definecolor{ntGold}{HTML}{FBEFD5}\definecolor{ntGoldH}{HTML}{F6E2B4}%
\newcommand{\ntSection}[1]{\noalign{\vskip 1.4pt{\color{rpxHTrack}\hrule height 8.6pt depth 0pt}\vskip-8.6pt}\multicolumn{2}{@{\hspace{3pt}}l@{}}{\rule[-2.4pt]{0pt}{8.6pt}\rpxHSection{#1}}\\[-1.2pt]\noalign{\vskip 1.6pt}}%
\newcommand{\ntVR}{{\color{rpxHMute!40}\vrule width 0.4pt}}%
\vspace{2pt}
\begin{tabularx}{\columnwidth}{@{\hspace{3pt}}>{\columncolor{ntGold}[3pt][4pt]}l@{\hspace{5pt}}!{\ntVR}@{\hspace{5pt}}>{\rmfamily\fontsize{7.4}{8.6}\selectfont\raggedright\arraybackslash}X@{\hspace{2pt}}}
\rpxHTop
\multicolumn{1}{@{\hspace{3pt}}>{\columncolor{ntGoldH}[3pt][4pt]}l@{\hspace{10.4pt}}}{\rule{0pt}{9pt}\rpxHHead{\color{rpxHInk}symbol}} & \multicolumn{1}{l}{\rpxHHead{definition}} \\
\rpxHMid
\ntSection{inputs}
$T$ & Number of aligned frames in the cell \\
$M_t$ & Instance mask at frame $t$ (pixel value $=$ instance ID, $0{=}$background) \\
$D_t$ & Metric depth map (metres; $0{=}$invalid) \\
$L_t$ & ITU-R BT.601 luma of the RGB frame \\
$p_t\in\mathbb R^3$ & T265 camera position \\
$q_t\in\mathbb H$ & T265 unit orientation quaternion \\
$H{\times}W$ & Frame resolution ($480{\times}640$) \\
\ntSection{derived sets and counts}
$\mathcal I_t$ & Visible instance IDs at frame~$t$; $n_t=|\mathcal I_t|$ \\
$\mathcal I$ & Union of all instance IDs across frames \\
$N$ & Total labelled instance count (from mask-to-object metadata) \\
$V_t$ & Valid-depth pixel set $\{(u,v):D_t(u,v)>0\}$ \\
$\mathcal M_t$ & Mask-covered pixel set $\{(u,v):M_t(u,v)>0\}$ \\
\rpxHBottom
\end{tabularx}
\end{table}
\begin{table*}[t]
\centering
\caption{Supplementary difficulty descriptors: schema, weights in RPX-DS (Eq.~\eqref{eq:appendix-esd-score}), and raw distribution over the 300 scene--phase cells. Statistics are raw descriptor values before percentile ranking, shown in the unit listed (e.g.\ $10^{-3}$: value $\times 10^{-3}$).}
\label{tab:appendix-esd-features}
\rpxHeatStyle{6.0}{0pt}%
\renewcommand{\arraystretch}{1.0}%
\definecolor{dfGold}{HTML}{FBEFD5}\definecolor{dfGoldH}{HTML}{F6E2B4}%
\newcommand{\dfSection}[1]{\noalign{\vskip 1.0pt{\color{rpxHTrack}\hrule height 8.2pt depth 0pt}\vskip-8.2pt}\multicolumn{10}{@{\hspace{3pt}}l@{}}{\rule[-2.3pt]{0pt}{8.2pt}\rpxHSection{#1}}\\[-1.2pt]\noalign{\vskip 1.0pt}}%
\newcommand{\dfFoot}[1]{{\color{rpxHMute}\rmfamily\fontsize{7.2}{8.4}\selectfont #1}}%
\begin{rpxfit}{\textwidth}{18}{@{\hspace{3pt}}l@{\hspace{\tabcolsep}}>{\columncolor{dfGold}[\tabcolsep][\tabcolsep]}c@{\hspace{\tabcolsep}}c@{\hspace{\tabcolsep}}r@{\hspace{\tabcolsep}}r@{\hspace{\tabcolsep}}r@{\hspace{\tabcolsep}}r@{\hspace{\tabcolsep}}r@{\hspace{\tabcolsep}}r@{\hspace{\tabcolsep}}r@{\hspace{3pt}}}
\rpxHTop
\multicolumn{1}{@{\hspace{3pt}}l@{\hspace{\tabcolsep}}}{\rule{0pt}{9pt}\rpxHHead{descriptor}} & \cellcolor{dfGoldH}\rpxHHead{\color{rpxHInk}weight} & \rpxHHead{unit} & \rpxHHead{mean} & \rpxHHead{SD} & \rpxHHead{median} & \rpxHHead{P10} & \rpxHHead{P90} & \rpxHHead{min} & \rpxHHead{max} \\
\rpxHMid
\dfSection{annotation effort}
iter\_mean & 0.125 &  & 1.40 & 0.56 & 1.13 & 1.00 & 2.12 & 1.00 & 3.90 \\
iter\_max & 0.125 &  & 2.31 & 1.05 & 3.00 & 1.00 & 4.00 & 1.00 & 5.00 \\
\dfSection{scene complexity}
obj\_mean & 0.030 &  & 6.97 & 0.16 & 7.00 & 6.90 & 7.00 & 5.90 & 8.00 \\
obj\_std & 0.030 &  & 0.101 & 0.158 & 0 & 0 & 0.311 & 0 & 0.803 \\
obj\_consist & 0.030 &  & 0.97 & 0.07 & 1.00 & 0.92 & 1.00 & 0.12 & 1.00 \\
\dfSection{occlusion}
occ\_mean & 0.030 &  & 0.185 & 0.169 & 0.139 & 0.020 & 0.445 & 0 & 0.818 \\
occ\_p90 & 0.030 &  & 0.32 & 0.29 & 0.24 & 0.04 & 0.74 & 0 & 1.54 \\
occ\_heavy & 0.030 &  & 0.20 & 0.26 & 0 & 0 & 0.57 & 0 & 1.00 \\
\dfSection{depth quality}
depth\_invalid & 0.030 & \smash{$10^{-2}$} & 2.37 & 1.74 & 2.08 & 0.46 & 4.77 & 0.03 & 9.23 \\
depth\_invalid\_mask & 0.030 &  & 0.018 & 0.022 & 0.007 & ${<}$0.001 & 0.051 & ${<}$0.001 & 0.108 \\
depth\_std & 0.030 &  & 2.14 & 1.80 & 1.59 & 0.45 & 4.86 & 0.20 & 8.96 \\
depth\_std\_mask & 0.030 &  & 0.18 & 0.10 & 0.17 & 0.09 & 0.24 & 0.06 & 1.04 \\
\dfSection{photometric conflict}
specular & 0.030 & \smash{$10^{-2}$} & 0.34 & 0.45 & 0.22 & ${<}$0.01 & 0.87 & 0 & 4.42 \\
dark & 0.030 & \smash{$10^{-3}$} & 0.70 & 1.14 & 0.18 & 0.01 & 2.41 & 0 & 7.10 \\
\dfSection{temporal stability}
area\_cv & 0.030 &  & 0.330 & 0.082 & 0.329 & 0.227 & 0.446 & 0.163 & 0.542 \\
area\_drop & 0.030 & \smash{$10^{-2}$} & 0.24 & 0.41 & 0.06 & 0 & 0.63 & 0 & 2.80 \\
vis\_instability & 0.030 & \smash{$10^{-2}$} & 0.24 & 0.54 & 0 & 0 & 0.58 & 0 & 4.51 \\
\dfSection{camera motion}
trans\_mean & 0.030 &  & 0.023 & 0.011 & 0.022 & 0.015 & 0.029 & 0.005 & 0.148 \\
trans\_p90 & 0.030 &  & 0.032 & 0.021 & 0.030 & 0.022 & 0.039 & 0.008 & 0.313 \\
rot\_mean & 0.030 & \smash{$10^{-2}$} & 2.36 & 0.57 & 2.54 & 1.57 & 2.96 & 0.67 & 3.95 \\
rot\_p90 & 0.030 & \smash{$10^{-2}$} & 3.49 & 0.75 & 3.61 & 2.41 & 4.35 & 1.29 & 6.06 \\
jerk & 0.030 & \smash{$10^{-2}$} & 0.55 & 0.50 & 0.50 & 0.31 & 0.71 & 0.15 & 6.58 \\
\dfSection{fisheye/stereo}
fisheye\_dark & 0.030 &  & 0.342 & 0.037 & 0.328 & 0.304 & 0.402 & 0.290 & 0.462 \\
fisheye\_bright & 0.030 &  & 0.062 & 0.024 & 0.059 & 0.029 & 0.093 & 0.008 & 0.107 \\
fisheye\_sharpness & 0.030 &  & 399 & 209 & 329 & 205 & 724 & 146 & 1046 \\
fisheye\_corr & 0.030 &  & 0.870 & 0.031 & 0.868 & 0.833 & 0.908 & 0.770 & 0.933 \\
fisheye\_texture & 0.030 &  & 4.75 & 1.24 & 4.52 & 3.44 & 6.33 & 3.04 & 8.22 \\
\rpxHMid
\multicolumn{4}{@{\hspace{3pt}}l}{\dfFoot{27 descriptors $=$ 2 effort $+$ 25 non-effort; $\sum_i w_i=1$}} & \multicolumn{6}{r@{\hspace{3pt}}}{\dfFoot{$n=300$ scene--phase cells}} \\
\rpxHBottom
\end{rpxfit}
\end{table*}

\subsubsection{Annotation effort (2 descriptors)}
\label{sec:feat-effort}

Let $\mathcal F_i$ contain the frame indices still marked faulty after manual correction round~$i$ ($i{=}1,\ldots,4$).
The per-frame effort count
\begin{equation}
k_t = 1 + \bigl|\{\, i : t \in \mathcal{F}_i \,\}\bigr|
\label{eq:effort-k}
\end{equation}
counts review passes for frame~$t$: $k_t{=}1$ if accepted on the first pass, $k_t{=}5$ if still faulty after four rounds.
\begin{equation}
\texttt{iter\_mean} = \frac{1}{T}\sum_{t=1}^{T} k_t, \qquad
\texttt{iter\_max} = \max_{t}\, k_t.
\label{eq:effort-feats}
\end{equation}

\subsubsection{Scene complexity (3 descriptors)}
\label{sec:feat-complexity}

\begin{align}
\texttt{obj\_mean} &= \frac{1}{T}\sum_{t} n_t,
\label{eq:obj-mean}\\
\texttt{obj\_std} &= \operatorname{std}_{t}\, n_t,
\label{eq:obj-std}\\
\texttt{obj\_consist} &= \frac{1}{T}\sum_{t} \mathbf{1}[n_t = N].
\label{eq:obj-consist}
\end{align}
\texttt{obj\_consist} is the fraction of frames in which all $N$ labelled instances are visible.

\subsubsection{Occlusion (3 descriptors)}
\enlargethispage{\baselineskip}
\label{sec:feat-occlusion}

For each visible instance~$i$ at frame~$t$, let $B_i^t$ be its tight bounding box with area $|B_i^t|$.
The normalized pairwise overlap is
\begin{equation}
o_i^t = \frac{1}{|B_i^t|} \sum_{j \in \mathcal{I}_t \setminus \{i\}} \bigl|B_i^t \cap B_j^t\bigr|.
\label{eq:occ-ratio}
\end{equation}
Taking the temporal mean per instance, $\bar o_i{=}\operatorname{mean}_t o_i^t$ (over frames where $i$ is visible), the three features are
\begin{align}
\texttt{occ\_mean} &= \operatorname{mean}_{i}\, \bar o_i,
\label{eq:occ-mean}\\
\texttt{occ\_p90} &= P_{90}\bigl(\{\bar o_i\}\bigr),
\label{eq:occ-p90}\\
\texttt{occ\_heavy} &= \frac{1}{|\mathcal{I}|}\sum_{i} \mathbf{1}[\bar o_i > 0.3].
\label{eq:occ-heavy}
\end{align}

\subsubsection{Depth quality (4 descriptors)}
\label{sec:feat-depth}

\begin{align}
\texttt{depth\_invalid} &= \frac{1}{T}\sum_{t} \frac{|\overline{V_t}|}{HW},
\label{eq:depth-invalid}\\
\texttt{depth\_invalid\_mask} &= \frac{1}{T}\sum_{t} \frac{|\overline{V_t} \cap \mathcal{M}_t|}{|\mathcal{M}_t|}.
\label{eq:depth-invalid-mask}
\end{align}
Depth spread is measured over all valid pixels and over object pixels:
\begin{align}
\texttt{depth\_std} &= \frac{1}{T}\sum_{t} \sigma\bigl(D_t\big|_{V_t}\bigr),
\label{eq:depth-std}\\
\texttt{depth\_std\_mask} &= \frac{1}{T}\sum_{t} \sigma\bigl(D_t\big|_{V_t \cap \mathcal{M}_t}\bigr),
\label{eq:depth-std-mask}
\end{align}
where $\sigma(D_t|_S)$ is the standard deviation of depth values over pixel set~$S$.

\subsubsection{Photometric conflict (2 descriptors)}
\label{sec:feat-photometric}

These descriptors count very bright pixels $\mathcal{B}_t$ ($L_t>230$) and very dark pixels $\mathcal{K}_t$ ($L_t<30$) that have no valid depth:
\begin{align}
\texttt{specular} &= \frac{1}{T\,HW}\sum_{t} \bigl|\mathcal{B}_t \cap \overline{V_t}\bigr|,
\label{eq:specular}\\
\texttt{dark} &= \frac{1}{T\,HW}\sum_{t} \bigl|\mathcal{K}_t \cap \overline{V_t}\bigr|.
\label{eq:dark}
\end{align}

\subsubsection{Temporal stability (3 descriptors)}
\label{sec:feat-temporal}

Let $A_i^t$ be the pixel area of instance~$i$ at frame~$t$ ($0$ if not visible), $v_i^t=\mathbf{1}[A_i^t{>}0]$ its visibility indicator, and $r_i^t=(A_i^{t-1}{-}A_i^t)/A_i^{t-1}$ the relative area drop (defined when $A_i^{t-1}{>}0$).
\begin{align}
\texttt{area\_cv} &= \frac{1}{|\mathcal{I}|}\sum_{i} \frac{\operatorname{std}_{t} A_i^t}{\operatorname{mean}_{t} A_i^t},
\label{eq:area-cv}\\
\texttt{area\_drop} &= \frac{1}{T\,|\mathcal{I}|}\sum_{i}\sum_{t=2}^{T} \mathbf{1}\bigl[r_i^t > \tfrac{1}{2}\bigr],
\label{eq:area-drop}\\
\texttt{vis\_instability} &= \frac{1}{T\,|\mathcal{I}|}\sum_{i}\sum_{t=2}^{T} \mathbf{1}\bigl[v_i^t \neq v_i^{t-1}\bigr].
\label{eq:vis-instab}
\end{align}
\texttt{area\_cv} (Eq.~\ref{eq:area-cv}) is computed only over frames where $A_i^t{>}0$.
\texttt{area\_drop} is the fraction of frames in which an instance loses more than half of its area, and \texttt{vis\_instability} the fraction in which it appears or disappears.

\subsubsection{Camera motion (5 descriptors)}
\label{sec:feat-camera}

From the T265 trajectory $(p_t,q_t)_{t=1}^{T}$, the per-frame increments are
\begin{equation}
\begin{aligned}
\delta_t^{\mathrm{trans}} &= \lVert p_{t+1} - p_t \rVert_2,\\
\delta_t^{\mathrm{rot}} &= 2\arccos\bigl|\langle q_t, q_{t+1} \rangle\bigr|,
\end{aligned}
\label{eq:motion-delta}
\end{equation}
where $\langle q_t,q_{t+1}\rangle$ is the quaternion inner product and the absolute value handles the double-cover ambiguity.
Before the statistics are computed, increments more than $5\times1.4826$ median absolute deviations above the median are replaced by the median.
\begin{align}
\texttt{trans\_mean} &= \operatorname{mean}_{t}\, \delta_t^{\mathrm{trans}},
\label{eq:trans-mean}\\
\texttt{trans\_p90} &= P_{90}\bigl(\delta_t^{\mathrm{trans}}\bigr),
\label{eq:trans-p90}\\
\texttt{rot\_mean} &= \operatorname{mean}_{t}\, \delta_t^{\mathrm{rot}},
\label{eq:rot-mean}\\
\texttt{rot\_p90} &= P_{90}\bigl(\delta_t^{\mathrm{rot}}\bigr),
\label{eq:rot-p90}\\
\texttt{jerk} &= \frac{1}{T-2}\sum_{t=1}^{T-2} \bigl|\delta_{t+1}^{\mathrm{trans}} - \delta_t^{\mathrm{trans}}\bigr|.
\label{eq:jerk}
\end{align}

\subsubsection{Fisheye/stereo quality (5 descriptors)}
\label{sec:feat-fisheye}

Computed from the T265 wide-FOV fisheye image pairs $(F_t^L, F_t^R)$.
Let $L_t^{\rm fe}$ denote the grayscale intensity of the left fisheye frame, with resolution $H_{\rm fe}{\times}W_{\rm fe}$, and $\mathcal{B}_t^{\rm fe}$, $\mathcal{K}_t^{\rm fe}$ its pixels above 230 and below 30.
\begin{align}
\texttt{fisheye\_dark} &= \frac{1}{T H_{\mathrm{fe}} W_{\mathrm{fe}}}\sum_{t} \bigl|\mathcal{K}_t^{\mathrm{fe}}\bigr|,
\label{eq:fe-dark}\\
\texttt{fisheye\_bright} &= \frac{1}{T H_{\mathrm{fe}} W_{\mathrm{fe}}}\sum_{t} \bigl|\mathcal{B}_t^{\mathrm{fe}}\bigr|.
\label{eq:fe-bright}
\end{align}
Sharpness uses the variance of the Laplacian, and texture density uses the mean gradient magnitude:
\begin{align}
\texttt{fisheye\_sharpness} &= \frac{1}{T}\sum_{t} \operatorname{Var}\bigl(\nabla^2 F_t^L\bigr),
\label{eq:fe-sharp}\\
\texttt{fisheye\_texture} &= \frac{1}{T}\sum_{t} \operatorname{mean}_{u,v} \lVert \nabla F_t^L \rVert_2.
\label{eq:fe-texture}
\end{align}
The left--right stereo correlation is
\begin{equation}
\texttt{fisheye\_corr} = \frac{1}{T}\sum_{t} \rho\bigl(\operatorname{vec} F_t^L,\, \operatorname{vec} F_t^R\bigr),
\label{eq:fe-corr}
\end{equation}
where $\rho$ is the Pearson correlation over all pixel values.
\subsection{Feature matrix}

\begin{figure*}[t]
\centering
\rpxRoundedImage[width=\textwidth]{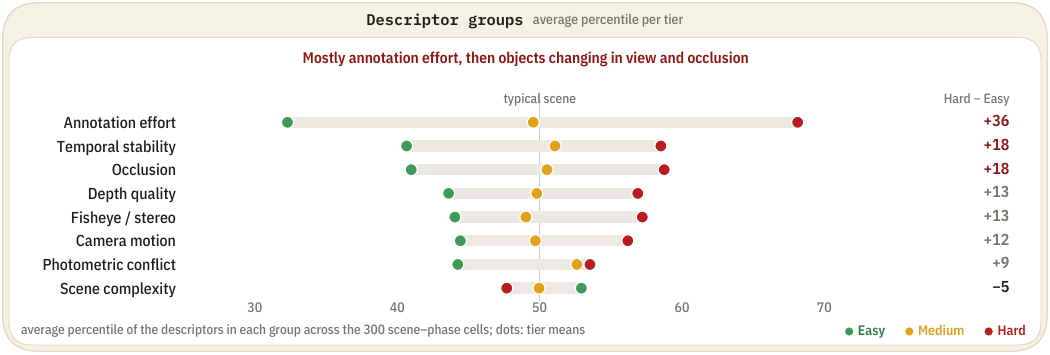}
\caption{What separates the difficulty tiers. Each descriptor is percentile-ranked over the 300 scene--phase cells, as in RPX-DS, averaged over a scene's phases and over the descriptors of each group; dots are the Easy, Medium, and Hard means and the right column is the Hard minus Easy gap in percentile points. Annotation effort separates the tiers most; Hard scenes show fewer visible objects, so scene complexity runs slightly the other way.}
\label{fig:appendix-esd-heatmap}
\end{figure*}

Figure~\ref{fig:appendix-esd-heatmap} shows which descriptor groups separate the tiers.

\begin{figure*}[t]
\centering
\begin{subfigure}[t]{158.4bp}\centering
\includegraphics{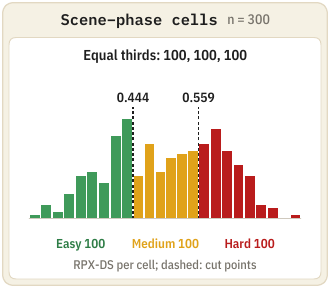}
\caption{Scene--phase cells ($n{=}300$)}\label{fig:rpx-ds-cells}
\end{subfigure}\hfill
\begin{subfigure}[t]{158.4bp}\centering
\includegraphics{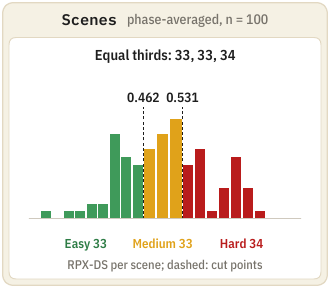}
\caption{Phase-averaged scenes ($n{=}100$)}\label{fig:rpx-ds-scenes}
\end{subfigure}\hfill
\begin{subfigure}[t]{158.4bp}\centering
\includegraphics{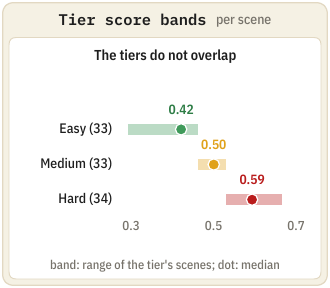}
\caption{Scenes by difficulty tier}\label{fig:rpx-ds-tiers}
\end{subfigure}
\caption{RPX-DS scores. (a, b)~Score histograms with the two cut points that split cells and scenes into equal thirds. (c)~Score range (band) and median (dot) of the scenes in each tier; the tiers do not overlap.}
\label{fig:appendix-esd-distributions}\label{fig:scene-swarm}
\end{figure*}

\subsection{Score construction}

The descriptors are combined into the \coolname{} difficulty score (RPX-DS).
Each raw descriptor is percentile-ranked across nonmissing scene--phase cells; ties receive their average rank, constant columns map to $0.5$, and unavailable optional modalities receive the neutral midpoint.
% keep the lead-in with its display: a float page otherwise splits the sentence
\begin{samepage}
With effort set $E$ and $\alpha=0.25$:
\begin{equation}
S_c = \sum_{i} w_i\, \widetilde F_{c,i}, \qquad
w_i =
\begin{cases}
  \dfrac{\alpha}{2}, & i \in E,\\[6pt]
  \dfrac{1-\alpha}{25}, & i \notin E,
\end{cases}
\label{eq:appendix-esd-score}
\end{equation}
so each effort feature has weight $0.125$ and each non-effort feature $0.03$.
\end{samepage}
The three exocentric phase scores are averaged per scene.
\subsection{Score distributions}

The 33\% and 67\% quantiles are 0.444 and 0.559 for the 300 cells and 0.462 and 0.531 for the 100 scenes.

Figures~\ref{fig:rpx-ds-cells} and~\ref{fig:rpx-ds-scenes} show the cell and scene distributions with their cut points, and Fig.~\ref{fig:rpx-ds-tiers} shows the per-scene scores in each tier.

\subsection{Difficulty-tier membership}

Table~\ref{tab:appendix-scene-objects} gives the tier and environment of every scene, and Fig.~\ref{fig:environment-tier-composition} the counts; the same tier labels are used throughout \coolname{} and index the scene catalogue of Appendix~\ref{sec:appendix-scenes}.

\section{Evaluation Mechanics}
\label{sec:appendix-eval}

\subsection{Task metric definitions}
\label{sec:appendix-metrics}

Table~\ref{tab:appendix-desirability} lists every metric in the MANOVA $K$-vectors with its equation.
Full definitions follow, grouped by task.

\subsubsection{T1--T2: depth ($K{=}4$ image; $K{=}6$ video)}
\label{sec:metrics-depth}

Let $\hat D_t, D_t$ be predicted and ground-truth depth at frame~$t$, evaluated over valid pixels $V_t{=}\{(u,v){:}0.3{<}D_t(u,v){<}5\,\mathrm{m}\}$ with $m_t{=}|V_t|$.
Write $e_i=\log\hat D_i-\log D_i$ and $\Delta D_t=D_{t+1}-D_t$.
\begin{align}
\mathrm{AbsRel} &= \frac{1}{m_t}\sum_{V_t} \frac{|\hat D - D|}{D},
\label{eq:absrel}\\
\mathrm{RMSE} &= \sqrt{\frac{1}{m_t}\sum_{V_t} (\hat D - D)^2},
\label{eq:rmse}\\
\mathrm{SILog} &= 100\,\sqrt{\frac{1}{m_t}\sum_{i} e_i^2 - \Bigl(\frac{1}{m_t}\sum_{i} e_i\Bigr)^{2}},
\label{eq:silog}\\
\delta_1 &= \frac{1}{m_t}\,\Bigl|\Bigl\{ V_t : \max\Bigl(\frac{\hat D}{D}, \frac{D}{\hat D}\Bigr) < 1.25 \Bigr\}\Bigr|.
\label{eq:delta1}
\end{align}
All four are computed per frame and averaged over a cell.
Video depth (T2) adds two temporal metrics on consecutive-frame depth changes; TGM uses only the static pixels $S_t{=}\{V_t{:}|\Delta D_t|{<}0.05\,\mathrm{m}\}$:
\begin{align}
\mathrm{TGM} &= \frac{1}{T-1}\sum_{t=1}^{T-1} \frac{1}{|S_t|}\sum_{S_t} \Bigl|\, |\Delta\hat D_t| - |\Delta D_t| \,\Bigr|,
\label{eq:tgm}\\
\mathrm{TGSE} &= \frac{1}{T-1}\sum_{t=1}^{T-1} \frac{1}{m_t}\sum_{V_t} \bigl(\Delta\hat D_t - \Delta D_t\bigr)^2.
\label{eq:tgse}
\end{align}
\textit{Alignment.}\enspace
Relative-depth models in both T1 and T2 receive one least-squares affine fit per scene--phase cell, pooled over all of its frames; per-frame fitting is not used.
Metric-depth models are evaluated without fitting.
All evaluation is restricted to $(0.3, 5.0)$\,m; invalid D435 readings are excluded.

\subsubsection{T3: tracking ($K{=}3$)}
\label{sec:metrics-tracking}

Let $\mathrm{TP}_t$, $\mathrm{FP}_t$, $\mathrm{FN}_t$ be detection counts at frame~$t$, $\mathrm{GT}_t$ the ground-truth count, and $\mathrm{IDSW}_t$ the identity-switch count.
\begin{align}
\mathrm{HOTA} &= \frac{1}{|\mathcal{A}|}\sum_{\alpha \in \mathcal{A}} \sqrt{\mathrm{DetA}_\alpha \cdot \mathrm{AssA}_\alpha},
\label{eq:hota}\\
\mathrm{MOTA} &= 1 - \frac{\sum_t \bigl(\mathrm{FP}_t + \mathrm{FN}_t + \mathrm{IDSW}_t\bigr)}{\sum_t \mathrm{GT}_t},
\label{eq:mota}\\
\mathrm{IDSW} &= \sum_{t=1}^{T} \mathrm{IDSW}_t,
\label{eq:idsw}
\end{align}
where $\mathrm{DetA}_\alpha{=}|\mathrm{TP}_\alpha|/(|\mathrm{TP}_\alpha|{+}|\mathrm{FP}_\alpha|{+}|\mathrm{FN}_\alpha|)$ at localization threshold~$\alpha$, and $\mathrm{AssA}_\alpha$ measures long-range identity consistency of the matched detections.
HOTA, DetA, and AssA are averaged over $\mathcal{A}{=}\{0.05, 0.10, \dots, 0.95\}$; MOTA and IDSW use the MOTChallenge IoU threshold of $0.5$, and IDSW is the total number of identity switches in the clip.

\subsubsection{T4: relative camera pose ($K{=}3$)}
\label{sec:metrics-pose}

For a predicted relative pose $(\hat R,\hat t)$ and ground truth $(R,t)$,
\begin{equation}
e_R = \arccos\frac{\operatorname{tr}(\hat R^{\top} R) - 1}{2}, \qquad
e_t = \arccos\frac{\hat t^{\top} t}{\lVert \hat t \rVert\, \lVert t \rVert}.
\label{eq:pose-err}
\end{equation}
The combined error $e{=}\max(e_R,e_t)$ yields
\begin{equation}
\begin{aligned}
R(x) &= \frac{1}{N}\sum_{i=1}^{N} \mathbf{1}[e_i \le x],\\
\mathrm{AUC@}\tau &= \frac{1}{\tau}\int_{0}^{\tau} R(x)\,\mathrm{d}x,
\end{aligned}
\label{eq:auc}
\end{equation}
with $\tau\in\{5^\circ,10^\circ,20^\circ\}$.
$R$ is the empirical cumulative distribution of the pose error, and $\mathrm{AUC@}\tau$ is its normalized area up to $\tau$.
T4 evaluates Clutter and Clean only; Interaction is excluded because objects move during manipulation and pairwise relative pose assumes a static scene.

\subsubsection{T5--T6: visual grounding ($K{=}3$)}
\label{sec:metrics-vqa}

Each model returns a box $\hat B$; ground truth $B$ is the tight axis-aligned enclosure of the instance mask.
Let $C$ be the smallest box enclosing $\hat B\cup B$.
\begin{align}
\mathrm{Acc@0.5} &= \mathbf{1}\bigl[\mathrm{IoU}(\hat B, B) \ge 0.5\bigr],
\label{eq:acc05}\\
\mathrm{GIoU} &= \mathrm{IoU}(\hat B, B) - \frac{|C \setminus (\hat B \cup B)|}{|C|},
\label{eq:giou}\\
\mathrm{CiGT} &= \mathbf{1}\bigl[\operatorname{center}(\hat B) \in B\bigr].
\label{eq:cigt}
\end{align}
GIoU lies in $[-1,1]$ and, unlike IoU, keeps decreasing as a non-overlapping box moves away from the target.
CiGT credits a box whose centre lies in the target even when its extent is wrong.
A missing, unparseable, or invalid box scores $0$ on Acc@0.5 and CiGT and $-1$ on GIoU.
\subsection{VQA question design}
\label{sec:appendix-qa}\label{sec:appendix-vqa-gen}

Scored questions are generated from the colour, material, and function responses of the SOS questionnaire (Table~\ref{tab:sos-schema}) and the instance masks; they do not use object names or categories.
The canonical catalogue (Table~\ref{tab:sos-canonical}) shows one normalized response per field; the release keeps all original responses and aliases.
Both tasks are scored with the grounding metrics of Appendix~\ref{sec:metrics-vqa}.

\subsubsection{Regular questions (T5)}

The question plan contains eight T5 types, five attribute and three spatial; the published T5 scores ($\mathcal J$, $\Phi$) use the five attribute types.
Three attribute types refer to the target by a single colour, material, or function (e.g.\ ``the red object'', ``the metal object'', ``the object used for cutting'').
The remaining two combine or negate questionnaire attributes: a composition type names both material and function (``the object made of \emph{metal} that is used for \emph{cutting}''), and an odd-one-out type asks for the object \emph{not} made of a given material.
The spatial types (leftmost or rightmost, closest in depth, farthest in the image plane) use exocentric views only and are unscored, as are five further released types: yes/no presence of a named object or a colour, material counts, and left--right and up--down binaries.

\subsubsection{In-context questions (T6)}

T6 replaces the textual subject reference with an SOS photograph followed by the target scene image.
Its plan contains the five attribute types of T5 and the farthest-object spatial type; the published T6 scores use the five attribute types.
The model must localize the referenced object in that scene.

\subsection{MANOVA estimator choice}

$\Phi$ is computed from Wilks' $\Lambda$.
Let $\lambda_1,\dots,\lambda_r$ be the nonzero eigenvalues of $\mathbf{H}\mathbf{E}^{-1}$, with $r=\min(K,\,P{-}1)$, and $\rho_i^2=\lambda_i/(1{+}\lambda_i)$ the squared canonical correlations.
Wilks and Pillai yield:
\begin{equation}
\eta^2_{\mathrm{W}} = 1 - \operatorname{GM}\bigl(1 - \rho_i^2\bigr),
\qquad
\eta^2_{\mathrm{P}} = 1 - \operatorname{AM}\bigl(1 - \rho_i^2\bigr).
\end{equation}
Because $\mathrm{GM} \leq \mathrm{AM}$, these definitions give $\Phi_{\text{Wilks}} \leq \Phi_{\text{Pillai}}$.
We report the Wilks-based $\Phi$, the smaller of the two.

\subsection{Model evaluation configurations}
\label{sec:appendix-model-config}

Table~\ref{tab:appendix-model-config} lists, for every model configuration, the input handling, initialization or prompt, and depth alignment used in evaluation, beyond what the main-paper result tables report.
\begin{table*}[t]
\centering
\caption{Evaluation configuration of every model--task evaluation, beyond what main-paper Tables~\ref{tab:t1}--\ref{tab:vqa} list.}
\label{tab:appendix-model-config}
\rpxHeatStyle{5.6}{0pt}%
\renewcommand{\arraystretch}{1.16}%
\definecolor{mcGold}{HTML}{FBEFD5}\definecolor{mcGoldH}{HTML}{F6E2B4}%
\newcommand{\mcSection}[1]{\noalign{\vskip 1.4pt{\color{rpxHTrack}\hrule height 8.6pt depth 0pt}\vskip-8.6pt}\multicolumn{4}{@{\hspace{3pt}}l@{}}{\rule[-2.4pt]{0pt}{8.6pt}\rpxHSection{#1}}\\[-1.2pt]\noalign{\vskip 1.6pt}}%
\begin{tabular}{@{\hspace{3pt}}>{\columncolor{mcGold}[3pt][4pt]\rmfamily\fontsize{7.4}{8.6}\selectfont\raggedright\arraybackslash}p{136pt}@{\hspace{8pt}}>{\rmfamily\fontsize{7.4}{8.6}\selectfont\raggedright\arraybackslash}p{92pt}@{\hspace{8pt}}>{\rmfamily\fontsize{7.4}{8.6}\selectfont\raggedright\arraybackslash}p{106pt}@{\hspace{8pt}}>{\rmfamily\fontsize{7.4}{8.6}\selectfont\raggedright\arraybackslash}p{141.8pt}@{\hspace{3pt}}}
\rpxHTop
\multicolumn{1}{@{\hspace{3pt}}>{\columncolor{mcGoldH}[3pt][4pt]}l@{\hspace{8pt}}}{\rule{0pt}{9pt}\fontfamily{IBMPlexMono-TLF}\selectfont\rpxHHead{\color{rpxHInk}configurations}} & \multicolumn{1}{l@{\hspace{8pt}}}{\fontfamily{IBMPlexMono-TLF}\selectfont\rpxHHead{input handling}} & \multicolumn{1}{l@{\hspace{8pt}}}{\fontfamily{IBMPlexMono-TLF}\selectfont\rpxHHead{initialization / prompt}} & \multicolumn{1}{l@{\hspace{3pt}}}{\fontfamily{IBMPlexMono-TLF}\selectfont\rpxHHead{alignment / special setting}} \\
\rpxHMid
\mcSection{T1 image depth}
DA3 Metric-L, DA-V2 Large, Depth Pro, UniDepth V2, Metric3D V2, MoGe-2 ViT-L, HyDen & single RGB frame & N/A & metric output; no scale or shift fitting \\
Lotus-2, FE2E, ZipDepth & single RGB frame & N/A & relative output; least-squares scale and shift to D435 depth \\
\mcSection{T2 video depth}
DA3 (video), Video-DA, ViGeo, VGGT, GemDepth, DVD v1.1 & phase clip in frame order & N/A & relative output; one scale and shift per clip \\
DepthCrafter, ChronoDepth & phase clip; sliding windows & N/A & relative output; one scale and shift per clip \\
RollingDepth & phase clip; streaming & N/A & relative output; one scale and shift per clip \\
MonST3R & phase clip; 4-frame multi-view chunks & N/A & relative output; one scale and shift per clip \\
\mcSection{T3 multi-object tracking}
SAM 2 Long, DAM4SAM, EdgeTAM, SAM 2++, SAM 2, Cutie, MiTS, XMem & RGB sequence & ground-truth masks of the first frame & persistent IDs scored one-to-one \\
SAM 3.1 (bbox) & RGB sequence & ground-truth boxes of the first frame & persistent IDs scored one-to-one \\
SAM 3.1 (text) & RGB sequence & one fixed prompt per object and clip, as for GSAM2 (text) & persistent IDs scored one-to-one \\
GSAM2 (text) & RGB sequence & one fixed prompt per object and clip: ``$\langle$primary colour$\rangle$ $\langle$object name$\rangle$'' & persistent IDs scored one-to-one \\
\mcSection{T4 relative camera pose}
all 10 models of the T4 table in the main paper & RGB frame pair within one capture & N/A & Clutter and Clean pairs scored separately; five-frame stride; GT in OpenCV camera axes \\
\mcSection{T5 regular grounding}
Qwen3-VL 8B / 2B & target image + text question & question text & JSON box on a 0--1000 coordinate grid; one scored generation; greedy decoding; 96-token limit \\
Cosmos Reason2, InternVL 3.5, PaliGemma 2, DeepSeek-VL2, Florence 2 (all sizes) & target image + text question & deterministic referring expression from the question & native or architecture-specific grounding output; one scored generation; greedy decoding; 64--128-token limit \\
\mcSection{T6 in-context grounding}
same 12 configurations as T5 & SOS reference image then target image & question referring to the reference & same grounding mode and decoding as T5 \\
\mcSection{T5--T6 shared}
all VLMs & task-specific image input & task-specific question or referring expression & boxes converted to inclusive pixel XYXY in original image coordinates; missing or invalid box = miss \\
\rpxHBottom
\end{tabular}
\end{table*}

\subsection{Metric desirability windows}
\label{sec:appendix-desirability}

Each native metric $v_k$ enters $\mathcal J$ through the desirability transform (Eq.~\eqref{eq:jedi}):
\[
d_k=\operatorname{clip}\!\Bigl(\frac{v_k-v_k^{\mathrm{worst}}}{v_k^{\mathrm{best}}-v_k^{\mathrm{worst}}},\;0,\;1\Bigr).
\]
Windows are frozen and shared by all models: bounded metrics use their natural range, and unbounded errors use a fixed worse endpoint beyond which $d_k=0$.
A value outside its window is clipped rather than refitted, so adding a model never changes the scores of the others.

Table~\ref{tab:appendix-desirability} lists the desirability window of every metric used in T1--T6.
\begin{table}[!htb]
\centering
\caption{Desirability windows of all native metrics. $d_k=0$ at the worse endpoint and $1$ at the better endpoint; range: natural bounds of the metric; fixed: frozen error cap shared by all models. Values beyond an endpoint are clipped: MOTA, which can be negative, at 0 (all phase means are at least 0.59), and SILog at 25, which the phase means of three video-depth models exceed.}
\label{tab:appendix-desirability}
\rpxHeatStyle{6.6}{0pt}%
\renewcommand{\arraystretch}{1.0}%
\definecolor{dwGold}{HTML}{FBEFD5}\definecolor{dwGoldH}{HTML}{F6E2B4}%
\newcommand{\dwSection}[1]{\noalign{\vskip 1.2pt{\color{rpxHTrack}\hrule height 8.2pt depth 0pt}\vskip-8.2pt}\multicolumn{7}{@{\hspace{3pt}}l@{}}{\rule[-2.3pt]{0pt}{8.2pt}\rpxHSection{#1}}\\[-1.2pt]\noalign{\vskip 1.0pt}}%
\begin{tabular*}{\columnwidth}{@{\extracolsep{\fill}}@{\hspace{3pt}}ll>{\columncolor{dwGold}[4pt][3pt]}c@{\extracolsep{0pt}\hspace{6pt}}>{\columncolor{dwGold}[3pt][4pt]}c@{\extracolsep{\fill}}clr@{\hspace{3pt}}}
\rpxHTop
\multicolumn{1}{@{\hspace{3pt}}l}{\rule{0pt}{9pt}\fontfamily{IBMPlexMono-TLF}\selectfont\rpxHHead{metric}} & \multicolumn{1}{l}{\fontfamily{IBMPlexMono-TLF}\selectfont\rpxHHead{task}} & \multicolumn{1}{>{\columncolor{dwGoldH}[4pt][3pt]}c@{\extracolsep{0pt}\hspace{6pt}}}{\fontfamily{IBMPlexMono-TLF}\selectfont\rpxHHead{\color{rpxHInk}worse}} & \multicolumn{1}{>{\columncolor{dwGoldH}[3pt][4pt]}c@{\extracolsep{\fill}}}{\fontfamily{IBMPlexMono-TLF}\selectfont\rpxHHead{\color{rpxHInk}better}} & \multicolumn{1}{c}{\fontfamily{IBMPlexMono-TLF}\selectfont\rpxHHead{dir.}} & \multicolumn{1}{l}{\fontfamily{IBMPlexMono-TLF}\selectfont\rpxHHead{basis}} & \multicolumn{1}{r@{\hspace{3pt}}}{\fontfamily{IBMPlexMono-TLF}\selectfont\rpxHHead{eq.}} \\
\rpxHMid
\dwSection{depth}
AbsRel & {\color{rpxHMute}T1, T2} & 0.5 & 0 & \rpxDown{} & fixed & \ref{eq:absrel} \\
RMSE{\rmfamily\fontsize{7.2}{8.4}\selectfont\color{rpxHMute}\ (m)} & {\color{rpxHMute}T1, T2} & 2.0 & 0 & \rpxDown{} & fixed & \ref{eq:rmse} \\
SILog & {\color{rpxHMute}T1, T2} & 25 & 0 & \rpxDown{} & fixed & \ref{eq:silog} \\
$\delta_1$ & {\color{rpxHMute}T1, T2} & 0 & 1 & \rpxUp{} & {\color{rpxHMute}range} & \ref{eq:delta1} \\
TGM & {\color{rpxHMute}T2} & 0.1 & 0 & \rpxDown{} & fixed & \ref{eq:tgm} \\
TGSE & {\color{rpxHMute}T2} & 0.1 & 0 & \rpxDown{} & fixed & \ref{eq:tgse} \\
\dwSection{tracking}
HOTA & {\color{rpxHMute}T3} & 0 & 1 & \rpxUp{} & {\color{rpxHMute}range} & \ref{eq:hota} \\
MOTA & {\color{rpxHMute}T3} & 0 & 1 & \rpxUp{} & {\color{rpxHMute}range} & \ref{eq:mota} \\
IDSW{\rmfamily\fontsize{7.2}{8.4}\selectfont\color{rpxHMute}\ (per clip)} & {\color{rpxHMute}T3} & 100 & 0 & \rpxDown{} & fixed & \ref{eq:idsw} \\
\dwSection{relative pose}
AUC@$5^\circ$ & {\color{rpxHMute}T4} & 0 & 1 & \rpxUp{} & {\color{rpxHMute}range} & \ref{eq:auc} \\
AUC@$10^\circ$ & {\color{rpxHMute}T4} & 0 & 1 & \rpxUp{} & {\color{rpxHMute}range} & \ref{eq:auc} \\
AUC@$20^\circ$ & {\color{rpxHMute}T4} & 0 & 1 & \rpxUp{} & {\color{rpxHMute}range} & \ref{eq:auc} \\
\dwSection{grounding}
Acc@0.5 & {\color{rpxHMute}T5, T6} & 0 & 1 & \rpxUp{} & {\color{rpxHMute}range} & \ref{eq:acc05} \\
GIoU & {\color{rpxHMute}T5, T6} & $-$1 & 1 & \rpxUp{} & {\color{rpxHMute}range} & \ref{eq:giou} \\
Center-in-GT & {\color{rpxHMute}T5, T6} & 0 & 1 & \rpxUp{} & {\color{rpxHMute}range} & \ref{eq:cigt} \\
\rpxHBottom
\end{tabular*}
\end{table}

\section{Benchmarking}
\label{sec:appendix-bench}
\subsection{Findings}
\label{sec:appendix-findings}
Eight findings from the 65 evaluations and 76 hardware profiles; each links to its detailed subsection.

\smallskip
\noindent\textbf{F1. No model clears both diagnostic thresholds.}
None of the 65 evaluations reaches the golden region $\mathcal J_{\min}\geq0.75$, $\Phi\geq0.86$.
The closest model per task is $0.10$ (VGGT, video depth) to $0.54$ (Qwen3-VL 2B, in-context VQA) away from it (Sec.~\ref{sec:appendix-diagnostic-frontiers}).

\smallskip
\noindent\textbf{F2. Quality and robustness rank models differently.}
Within tasks, Spearman $\rho(\mathcal J_{\min},\Phi)$ ranges from $-0.86$ (T6) to $+0.37$ (T3), and the highest-$\mathcal J_{\min}$ model is also the most robust only in video depth.
A single score would hide this trade-off.

\smallskip
\noindent\textbf{F3. Phase sensitivity concentrates at Interaction in geometric tasks.}
For 30 of 31 depth and tracking models, the transition with the lowest $\Phi$ involves Interaction, and Interaction is the lowest-quality phase for all 11 trackers (Sec.~\ref{sec:appendix-robustness-sources}).

\smallskip
\noindent\textbf{F4. Harder scenes cost robustness only in geometric tasks.}
Hard-tier $\Phi$ is below Easy-tier $\Phi$ for 29 of 31 depth and tracking models but for only 7 of 24 VQA evaluations.
Scene difficulty therefore stresses geometry and tracking more than grounding.

\smallskip
\noindent\textbf{F5. A high $\Phi$ can mean uniformly weak predictions.}
Replacing text with a reference image lowers $\mathcal J_{\min}$ for all 12 VLMs but raises $\Phi$ for 8, and the most robust VQA configurations are Florence-2 models with $\mathcal J_{\min}\leq0.14$.
$\Phi$ must be read together with $\mathcal J_{\min}$.

\smallskip
\noindent\textbf{F6. Reference images cut grounding quality by $49$--$93\%$.}
In-context $\mathcal J_{\min}$ is lower than regular $\mathcal J_{\min}$ for all 12 VLMs, while model rankings are largely preserved (Spearman $\rho=0.87$; Sec.~\ref{sec:appendix-text-image}).

\smallskip
\noindent\textbf{F7. The egocentric camera changes the metric profile, not tracking quality.}
Ego and exo tracking quality differ by a median of $0.00$, yet $\Phi_{\mathrm{EE}}\leq0.86$ for 7 of 11 trackers; grounding improves on the ego view for all 12 regular-VQA models (Sec.~\ref{sec:appendix-ego-exo}).

\smallskip
\noindent\textbf{F8. Near-best quality is often far cheaper.}
UniDepth V2 is $22.7\times$ cheaper than Metric3D V2 for $0.02$ lower $\mathcal J_{\min}$, SAM~2 is $4.7\times$ cheaper than SAM~2~Long for $0.01$, and Qwen3-VL 2B is $2.2\times$ cheaper than 8B in-context for $0.01$.
In image depth, compute does not predict quality: parameter count and $\mathcal J_{\min}$ are uncorrelated ($\rho=+0.02$; Sec.~\ref{sec:appendix-hardware}).

\subsection{Shared evaluation framework}
All six tasks share one pipeline (Fig.~\ref{fig:overview}b,c).
A versioned split fixes sample identity and available modalities, and a task adapter invokes a reference model or any user-supplied local or API callable.
Predictions pass task-specific checks of shape, units, coordinates, completeness, and identifiers before scoring (Table~\ref{tab:appendix-task-protocols}); native metrics follow the definitions of Appendix~\ref{sec:appendix-eval}.
For VQA, malformed outputs and inference failures stay in the denominator; for the other tasks, the runner rejects predictions of the wrong type, shape, or count. Reports record the task, model, split, and dataset repository with stable sample identifiers.
\begin{table*}[t]
\centering
\caption{Task-specific evaluation protocols. Metrics are listed in Table~\ref{tab:appendix-desirability}.}
\label{tab:appendix-task-protocols}
\rpxHeatStyle{5.6}{0pt}%
\renewcommand{\arraystretch}{1.52}%
\definecolor{tpGold}{HTML}{FBEFD5}\definecolor{tpGoldH}{HTML}{F6E2B4}%
\newcommand{\tpId}[1]{\leavevmode{\fontfamily{IBMPlexMono-TLF}\fontsize{6.6}{8}\selectfont\color{rpxHMute}#1}}%
\begin{tabular}{@{\hspace{3pt}}>{\columncolor{tpGold}[3pt][4pt]\rmfamily\fontsize{7.6}{8.8}\selectfont\raggedright\arraybackslash}p{2.35cm}@{\hspace{10pt}}>{\rmfamily\fontsize{7.6}{8.8}\selectfont\raggedright\arraybackslash}p{4.9cm}@{\hspace{10pt}}>{\rmfamily\fontsize{7.6}{8.8}\selectfont\raggedright\arraybackslash}p{9.62cm}@{\hspace{3pt}}}
\rpxHTop
\multicolumn{1}{@{\hspace{3pt}}>{\columncolor{tpGoldH}[3pt][4pt]}l@{\hspace{10pt}}}{\rule{0pt}{9pt}\fontfamily{IBMPlexMono-TLF}\selectfont\rpxHHead{\color{rpxHInk}task}} & \multicolumn{1}{l@{\hspace{10pt}}}{\fontfamily{IBMPlexMono-TLF}\selectfont\rpxHHead{input / ground truth}} & \multicolumn{1}{l@{\hspace{3pt}}}{\fontfamily{IBMPlexMono-TLF}\selectfont\rpxHHead{key protocol detail}} \\
\rpxHMid
\tpId{T1}\ Image depth & RGB / aligned D435 depth & Valid range $(0.3,5.0)$\,m; one affine alignment per scene--phase cell for relative depth. \\
\arrayrulecolor{rpxHMute!30}\specialrule{0.3pt}{0pt}{0pt}
\tpId{T2}\ Video depth & RGB clip / ordered D435 depth & Preserve frame order and IDs; one relative alignment per clip. \\
\arrayrulecolor{rpxHMute!30}\specialrule{0.3pt}{0pt}{0pt}
\tpId{T3}\ Tracking & RGB sequence + initialization / masks and persistent IDs & Declare initialization; IDs persist across frames. \\
\arrayrulecolor{rpxHMute!30}\specialrule{0.3pt}{0pt}{0pt}
\tpId{T4}\ Relative pose & RGB pair / T265 relative pose & Clutter/Clean only; validate the pose convention. \\
\arrayrulecolor{rpxHMute!30}\specialrule{0.3pt}{0pt}{0pt}
\tpId{T5}\ Regular VQA & RGB + attribute question / mask-derived box & Five attribute question types scored; missing or invalid boxes are misses. \\
\arrayrulecolor{rpxHMute!30}\specialrule{0.3pt}{0pt}{0pt}
\tpId{T6}\ In-context VQA & Ordered SOS reference + query / mask-derived box & Reference image precedes the query; same grounding scoring as T5. \\
\rpxHBottom
\end{tabular}
\end{table*}
\subsection{Diagnostics used in this section}
\label{sec:appendix-derived}
Let $\mathcal J_{s,p}$ be the desirability of scene $s$ in phase $p$ (Eq.~\eqref{eq:jedi}) and $\mathcal J_p$ its scene average, so that $\mathcal J_{\min}=\min_p\mathcal J_p$.
$\Phi=1-\eta_p^2$ is the Wilks-based phase robustness over the scored phases (Appendix~\ref{sec:appendix-eval}). Restricting the same MANOVA gives the variants used below:
\begin{align}
\Phi_{p\leftrightarrow q} &= 1 - \eta_p^2\,\big|_{\{p,\,q\}}, \label{eq:phi-transition}\\
\Phi_{g} &= 1 - \eta_p^2\,\big|_{s \in \mathcal{S}_g}, \quad g \in \{E, M, H\}, \label{eq:phi-tier}\\
\Phi_{\mathrm{EE}} &= 1 - \eta_v^2\,\big|_{v \in \{\mathrm{ego},\, \mathrm{exo\text{-}Int}\}}, \label{eq:phi-ee}
\end{align}
with $\mathcal S_g$ the scenes of tier $g$ and $\eta_v^2$ the camera-view effect size.
An evaluation $a$ dominates $b$ when $\mathcal J_{\min}^{a}\geq\mathcal J_{\min}^{b}$ and $\Phi^{a}\geq\Phi^{b}$ with one inequality strict.
The distance to the golden region and the relative in-context drop are
\begin{align}
G &= \sqrt{(0.75 - \mathcal{J}_{\min})_{+}^{2} + (0.86 - \Phi)_{+}^{2}}, \label{eq:gap}\\
\Delta_{\mathrm{T6}} &= 100\, \frac{\mathcal{J}_{\min}^{\mathrm{T5}} - \mathcal{J}_{\min}^{\mathrm{T6}}}{\mathcal{J}_{\min}^{\mathrm{T5}}}. \label{eq:drop}
\end{align}
Here $(x)_{+}=\max(0,x)$. Associations are Spearman rank correlations $\rho$ across the models of one task.

\subsection{Quality--robustness trade-offs}
\label{sec:appendix-diagnostic-frontiers}
Main-paper Fig.~\ref{fig:phi-jmin} places all 65 evaluations against the two thresholds of the golden region; none meets both.
Ten of the 11 trackers pass the quality threshold (SAM~3.1 with text prompts falls short at $0.72$) but none passes robustness; all 10 relative-pose models and 3 regular and 5 in-context VQA configurations pass robustness but none passes quality; no depth model passes either.
The closest evaluations are VGGT and GSAM2 ($G=0.10$, $0.11$; Eq.~\eqref{eq:gap}), followed by SAM~2 and SAM~2++ ($0.13$).
Within tasks the two diagnostics are weakly or negatively associated ($\rho(\mathcal J_{\min},\Phi)=-0.43$, $-0.25$, $+0.37$, $-0.23$, $-0.59$, $-0.86$).
\subsection{Where robustness is lost}
\label{sec:appendix-robustness-sources}
\begin{figure*}[t]
\centering
\rpxRoundedImage[width=\textwidth]{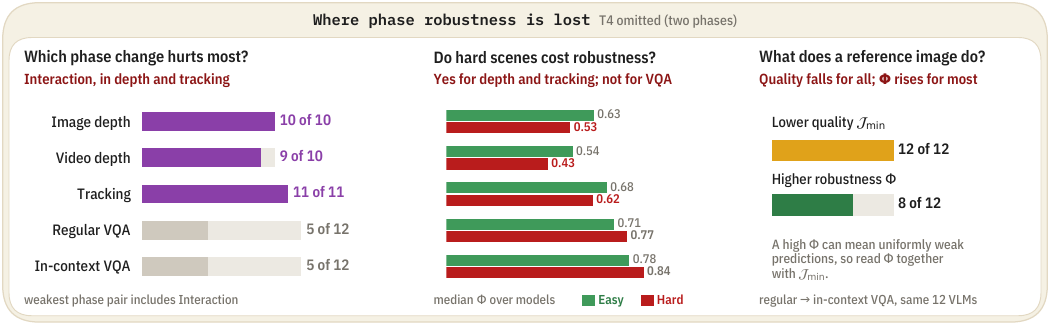}
\caption{Where phase robustness is lost. (a)~Models whose weakest phase transition, the lowest pairwise $\Phi$ (Eq.~\eqref{eq:phi-transition}), involves Interaction. (b)~Median tier robustness (Eq.~\eqref{eq:phi-tier}) on Easy and Hard scenes. (c)~Regular to in-context VQA for the same 12 VLMs: all lose $\mathcal J_{\min}$ and 8 gain $\Phi$. T4 is omitted (two phases).}
\label{fig:robustness-sources}
\end{figure*}

\noindent\textbf{Transitions.}
$\Phi$ follows each model's weakest transition (Eq.~\eqref{eq:phi-transition}; $\rho=0.95$, $0.81$, $0.97$, $0.96$, $0.97$ in T1, T2, T3, T5, T6).
In depth and tracking that transition involves Interaction for 30 of 31 models (Fig.~\ref{fig:robustness-sources}a).
In grounding it is Clutter$\leftrightarrow$Clean for 7 of 12 models in both tasks, although Interaction is the lowest-quality phase for 8 of 12 T5 and 5 of 12 T6 models (9 and 12 counting ties).

\smallskip
\noindent\textbf{Scene difficulty.}
Tier robustness (Eq.~\eqref{eq:phi-tier}) falls from Easy to Hard in the geometric tasks: $\Phi_H<\Phi_E$ for 10/10, 9/10, and 10/11 models in T1--T3, with medians falling from 0.63 to 0.53, 0.54 to 0.43, and 0.68 to 0.62 (Fig.~\ref{fig:robustness-sources}b).
Grounding shows no such decline ($\Phi_H<\Phi_E$ for 3/12 and 4/12 models), and better video-depth models lose less robustness from Easy to Hard ($\rho(\mathcal J_{\min},\Phi_E-\Phi_H)=-0.92$).

\smallskip
\noindent\textbf{Low-quality floor.}
A reference image lowers $\mathcal J_{\min}$ for all 12 VLMs yet raises $\Phi$ for 8 (Fig.~\ref{fig:robustness-sources}c), and the highest-$\Phi$ configuration in each VQA task is a Florence-2 model with $\mathcal J_{\min}\leq0.14$.
The two diagnostics agree in direction: the gap $\mathcal J-\mathcal J_{\min}$ shrinks as $\Phi$ rises in all six tasks, significantly in T2 and T6 ($\rho=-0.82$, $-0.83$; $p<0.01$).

\subsection{Detailed per-model metrics}
\label{sec:appendix-raw-metrics}
These phase aggregates summarize scene-level measurements; applying the geometric mean to a phase-averaged metric vector does not reproduce the scene-averaged quality score.
\rpxTableContext{\textbf{Depth.} Tables~\ref{tab:raw-t1} and~\ref{tab:raw-t2} report image-depth and video-depth metrics, respectively.}{generated/raw_t1_table}
\begin{table*}[tp]
\centering
\caption{T2 video depth: native metrics per phase. \textbf{Bold}: best; \underline{underlined}: each model's worst phase. Shade: \rpxHSwatchR\ higher and \rpxHSwatchL\ lower is better, as in Table~\ref{tab:raw-t1}.}\label{tab:raw-t2}
\rpxHeatStyle{6.3}{2pt}
\begin{rpxfit}{\textwidth}{24}{@{}l@{}@{\hspace{3.30pt}\hspace{5pt}\hspace{3.30pt}}ccc@{\hspace{3.30pt}\hspace{5pt}\hspace{3.30pt}}ccc@{\hspace{3.30pt}\hspace{5pt}\hspace{3.30pt}}ccc@{\hspace{3.30pt}\hspace{5pt}\hspace{3.30pt}}ccc@{\hspace{3.30pt}\hspace{5pt}\hspace{3.30pt}}ccc@{\hspace{3.30pt}\hspace{5pt}\hspace{3.30pt}}ccc@{}}
\rpxHTop
\rule{0pt}{9pt}\rpxHGroup{Model} & \multicolumn{3}{c}{\rpxHGroup{AbsRel\,\rpxDown{}}} & \multicolumn{3}{c}{\rpxHGroup{RMSE\,\rpxDown{}}} & \multicolumn{3}{c}{\rpxHGroup{SILog\,\rpxDown{}}} & \multicolumn{3}{c}{\rpxHGroup{$\delta_1$\,\rpxUp{}}} & \multicolumn{3}{c}{\rpxHGroup{TGM\,\rpxDown{}}} & \multicolumn{3}{c}{\rpxHGroup{TGSE\,\rpxDown{}}} \\
\rpxHGroupRules{\cmidrule(lr){2-4}\cmidrule(lr){5-7}\cmidrule(lr){8-10}\cmidrule(lr){11-13}\cmidrule(lr){14-16}\cmidrule(lr){17-19}}
& \rpxHHead{clu} & \rpxHHead{int} & \rpxHHead{cln} & \rpxHHead{clu} & \rpxHHead{int} & \rpxHHead{cln} & \rpxHHead{clu} & \rpxHHead{int} & \rpxHHead{cln} & \rpxHHead{clu} & \rpxHHead{int} & \rpxHHead{cln} & \rpxHHead{clu} & \rpxHHead{int} & \rpxHHead{cln} & \rpxHHead{clu} & \rpxHHead{int} & \rpxHHead{cln} \\
\rpxHMid
DA3 (video) & \rpxHCell{rpxHL}{0.988}{\textbf{0.060}} & \rpxHCell{rpxHL}{0.911}{\underline{0.079}} & \rpxHCell{rpxHL}{0.996}{0.058} & \rpxHCell{rpxHL}{1.000}{\textbf{0.209}} & \rpxHCell{rpxHL}{0.724}{\underline{0.316}} & \rpxHCell{rpxHL}{0.933}{\textbf{0.235}} & \rpxHCell{rpxHL}{1.000}{\textbf{9.03}} & \rpxHCell{rpxHL}{0.856}{\underline{12.45}} & \rpxHCell{rpxHL}{0.988}{\textbf{9.32}} & \rpxHCell{rpxHR}{1.000}{\textbf{0.969}} & \rpxHCell{rpxHR}{0.963}{\underline{0.951}} & \rpxHCell{rpxHR}{1.000}{\textbf{0.969}} & \rpxHCell{rpxHL}{0.890}{0.0210} & \rpxHCell{rpxHL}{0.645}{\underline{0.0284}} & \rpxHCell{rpxHL}{0.814}{0.0233} & \rpxHCell{rpxHL}{0.995}{0.0312} & \rpxHCell{rpxHL}{0.433}{\underline{0.0524}} & \rpxHCell{rpxHL}{0.767}{\textbf{0.0398}} \\
Video-DA & \rpxHCell{rpxHL}{0.393}{\underline{0.207}} & \rpxHCell{rpxHL}{0.494}{0.182} & \rpxHCell{rpxHL}{0.409}{0.203} & \rpxHCell{rpxHL}{0.473}{0.413} & \rpxHCell{rpxHL}{0.359}{\underline{0.457}} & \rpxHCell{rpxHL}{0.367}{0.454} & \rpxHCell{rpxHL}{0.338}{24.75} & \rpxHCell{rpxHL}{0.381}{23.73} & \rpxHCell{rpxHL}{0.296}{\underline{25.74}} & \rpxHCell{rpxHR}{0.405}{\underline{0.680}} & \rpxHCell{rpxHR}{0.529}{0.740} & \rpxHCell{rpxHR}{0.430}{0.692} & \rpxHCell{rpxHL}{0.774}{0.0245} & \rpxHCell{rpxHL}{0.565}{\underline{0.0308}} & \rpxHCell{rpxHL}{0.641}{0.0285} & \rpxHCell{rpxHL}{0.939}{0.0333} & \rpxHCell{rpxHL}{0.494}{\underline{0.0501}} & \rpxHCell{rpxHL}{0.656}{0.0440} \\
ViGeo & \rpxHCell{rpxHL}{0.866}{\underline{0.090}} & \rpxHCell{rpxHL}{0.935}{0.073} & \rpxHCell{rpxHL}{0.980}{0.062} & \rpxHCell{rpxHL}{0.773}{0.297} & \rpxHCell{rpxHL}{0.744}{\underline{0.308}} & \rpxHCell{rpxHL}{0.837}{0.272} & \rpxHCell{rpxHL}{0.782}{\underline{14.21}} & \rpxHCell{rpxHL}{0.857}{12.42} & \rpxHCell{rpxHL}{0.894}{11.55} & \rpxHCell{rpxHR}{0.874}{\underline{0.908}} & \rpxHCell{rpxHR}{0.979}{0.959} & \rpxHCell{rpxHR}{0.975}{0.957} & \rpxHCell{rpxHL}{0.934}{0.0197} & \rpxHCell{rpxHL}{0.688}{\underline{0.0271}} & \rpxHCell{rpxHL}{0.844}{0.0224} & \rpxHCell{rpxHL}{0.521}{0.0491} & \rpxHCell{rpxHL}{0.333}{\underline{0.0562}} & \rpxHCell{rpxHL}{0.473}{0.0509} \\
VGGT & \rpxHCell{rpxHL}{0.960}{\underline{0.067}} & \rpxHCell{rpxHL}{0.980}{\textbf{0.062}} & \rpxHCell{rpxHL}{1.000}{\textbf{0.057}} & \rpxHCell{rpxHL}{0.871}{0.259} & \rpxHCell{rpxHL}{0.837}{\underline{\textbf{0.272}}} & \rpxHCell{rpxHL}{0.881}{0.255} & \rpxHCell{rpxHL}{0.920}{10.93} & \rpxHCell{rpxHL}{0.915}{\underline{\textbf{11.05}}} & \rpxHCell{rpxHL}{0.958}{10.02} & \rpxHCell{rpxHR}{0.971}{\underline{0.955}} & \rpxHCell{rpxHR}{0.988}{\textbf{0.963}} & \rpxHCell{rpxHR}{0.988}{0.963} & \rpxHCell{rpxHL}{0.821}{\underline{0.0231}} & \rpxHCell{rpxHL}{0.837}{\textbf{0.0226}} & \rpxHCell{rpxHL}{0.837}{0.0226} & \rpxHCell{rpxHL}{0.658}{0.0439} & \rpxHCell{rpxHL}{0.722}{\textbf{0.0415}} & \rpxHCell{rpxHL}{0.624}{\underline{0.0452}} \\
MonST3R & \rpxHCell{rpxHL}{0.458}{\underline{0.191}} & \rpxHCell{rpxHL}{0.547}{0.169} & \rpxHCell{rpxHL}{0.506}{0.179} & \rpxHCell{rpxHL}{0.473}{0.413} & \rpxHCell{rpxHL}{0.354}{\underline{0.459}} & \rpxHCell{rpxHL}{0.388}{0.446} & \rpxHCell{rpxHL}{0.772}{14.44} & \rpxHCell{rpxHL}{0.665}{\underline{16.98}} & \rpxHCell{rpxHL}{0.780}{14.26} & \rpxHCell{rpxHR}{0.432}{\underline{0.693}} & \rpxHCell{rpxHR}{0.588}{0.769} & \rpxHCell{rpxHR}{0.492}{0.722} & \rpxHCell{rpxHL}{0.362}{0.0369} & \rpxHCell{rpxHL}{0.150}{\underline{0.0433}} & \rpxHCell{rpxHL}{0.230}{0.0409} & \rpxHCell{rpxHL}{0.881}{0.0355} & \rpxHCell{rpxHL}{0.312}{\underline{0.0570}} & \rpxHCell{rpxHL}{0.592}{0.0464} \\
GemDepth & \rpxHCell{rpxHL}{0.834}{\underline{0.098}} & \rpxHCell{rpxHL}{0.842}{0.096} & \rpxHCell{rpxHL}{0.875}{0.088} & \rpxHCell{rpxHL}{0.793}{0.289} & \rpxHCell{rpxHL}{0.597}{\underline{0.365}} & \rpxHCell{rpxHL}{0.736}{0.311} & \rpxHCell{rpxHL}{0.948}{10.26} & \rpxHCell{rpxHL}{0.816}{\underline{13.41}} & \rpxHCell{rpxHL}{0.930}{10.70} & \rpxHCell{rpxHR}{0.893}{\underline{0.917}} & \rpxHCell{rpxHR}{0.914}{0.927} & \rpxHCell{rpxHR}{0.930}{0.935} & \rpxHCell{rpxHL}{1.000}{\textbf{0.0177}} & \rpxHCell{rpxHL}{0.681}{\underline{0.0273}} & \rpxHCell{rpxHL}{0.867}{\textbf{0.0217}} & \rpxHCell{rpxHL}{0.976}{0.0319} & \rpxHCell{rpxHL}{0.343}{\underline{0.0558}} & \rpxHCell{rpxHL}{0.690}{0.0427} \\
DepthCrafter & \rpxHCell{rpxHL}{0.316}{0.226} & \rpxHCell{rpxHL}{0.352}{0.217} & \rpxHCell{rpxHL}{0.312}{\underline{0.227}} & \rpxHCell{rpxHL}{0.313}{0.475} & \rpxHCell{rpxHL}{0.150}{\underline{0.538}} & \rpxHCell{rpxHL}{0.163}{0.533} & \rpxHCell{rpxHL}{0.326}{25.03} & \rpxHCell{rpxHL}{0.328}{24.98} & \rpxHCell{rpxHL}{0.252}{\underline{26.80}} & \rpxHCell{rpxHR}{0.292}{0.625} & \rpxHCell{rpxHR}{0.327}{0.642} & \rpxHCell{rpxHR}{0.273}{\underline{0.616}} & \rpxHCell{rpxHL}{0.648}{0.0283} & \rpxHCell{rpxHL}{0.469}{0.0337} & \rpxHCell{rpxHL}{0.439}{\underline{0.0346}} & \rpxHCell{rpxHL}{0.860}{0.0363} & \rpxHCell{rpxHL}{0.431}{\underline{0.0525}} & \rpxHCell{rpxHL}{0.542}{0.0483} \\
DVD v1.1 & \rpxHCell{rpxHL}{0.822}{\underline{0.101}} & \rpxHCell{rpxHL}{0.834}{0.098} & \rpxHCell{rpxHL}{0.850}{0.094} & \rpxHCell{rpxHL}{0.669}{0.337} & \rpxHCell{rpxHL}{0.532}{\underline{0.390}} & \rpxHCell{rpxHL}{0.633}{0.351} & \rpxHCell{rpxHL}{0.850}{12.58} & \rpxHCell{rpxHL}{0.770}{\underline{14.48}} & \rpxHCell{rpxHL}{0.862}{12.31} & \rpxHCell{rpxHR}{0.842}{\underline{0.892}} & \rpxHCell{rpxHR}{0.887}{0.914} & \rpxHCell{rpxHR}{0.883}{0.912} & \rpxHCell{rpxHL}{0.977}{0.0184} & \rpxHCell{rpxHL}{0.565}{\underline{0.0308}} & \rpxHCell{rpxHL}{0.787}{0.0241} & \rpxHCell{rpxHL}{0.854}{0.0365} & \rpxHCell{rpxHL}{0.150}{\underline{0.0631}} & \rpxHCell{rpxHL}{0.478}{0.0507} \\
ChronoDepth & \rpxHCell{rpxHL}{0.474}{\underline{0.187}} & \rpxHCell{rpxHL}{0.514}{0.177} & \rpxHCell{rpxHL}{0.482}{0.185} & \rpxHCell{rpxHL}{0.401}{0.441} & \rpxHCell{rpxHL}{0.251}{\underline{0.499}} & \rpxHCell{rpxHL}{0.261}{0.495} & \rpxHCell{rpxHL}{0.686}{16.48} & \rpxHCell{rpxHL}{0.615}{\underline{18.18}} & \rpxHCell{rpxHL}{0.670}{16.87} & \rpxHCell{rpxHR}{0.465}{0.709} & \rpxHCell{rpxHR}{0.553}{0.752} & \rpxHCell{rpxHR}{0.459}{\underline{0.706}} & \rpxHCell{rpxHL}{0.907}{0.0205} & \rpxHCell{rpxHL}{0.641}{\underline{0.0285}} & \rpxHCell{rpxHL}{0.787}{0.0241} & \rpxHCell{rpxHL}{1.000}{\textbf{0.0310}} & \rpxHCell{rpxHL}{0.510}{\underline{0.0495}} & \rpxHCell{rpxHL}{0.767}{\textbf{0.0398}} \\
RollingDepth & \rpxHCell{rpxHL}{0.150}{\underline{0.267}} & \rpxHCell{rpxHL}{0.356}{0.216} & \rpxHCell{rpxHL}{0.259}{0.240} & \rpxHCell{rpxHL}{0.222}{0.510} & \rpxHCell{rpxHL}{0.199}{0.519} & \rpxHCell{rpxHL}{0.173}{\underline{0.529}} & \rpxHCell{rpxHL}{0.150}{\underline{29.21}} & \rpxHCell{rpxHL}{0.235}{27.20} & \rpxHCell{rpxHL}{0.159}{29.00} & \rpxHCell{rpxHR}{0.150}{\underline{0.556}} & \rpxHCell{rpxHR}{0.331}{0.644} & \rpxHCell{rpxHR}{0.236}{0.598} & \rpxHCell{rpxHL}{0.748}{0.0253} & \rpxHCell{rpxHL}{0.512}{\underline{0.0324}} & \rpxHCell{rpxHL}{0.548}{0.0313} & \rpxHCell{rpxHL}{0.984}{0.0316} & \rpxHCell{rpxHL}{0.640}{\underline{0.0446}} & \rpxHCell{rpxHL}{0.706}{0.0421} \\
\rpxHMid
\textcolor{rpxHMute}{\textit{Times worst}} & \textcolor{rpxHMute}{8} & \textcolor{rpxHMute}{1} & \textcolor{rpxHMute}{1} & \textcolor{rpxHMute}{0} & \textcolor{rpxHMute}{9} & \textcolor{rpxHMute}{1} & \textcolor{rpxHMute}{2} & \textcolor{rpxHMute}{6} & \textcolor{rpxHMute}{2} & \textcolor{rpxHMute}{7} & \textcolor{rpxHMute}{1} & \textcolor{rpxHMute}{2} & \textcolor{rpxHMute}{1} & \textcolor{rpxHMute}{8} & \textcolor{rpxHMute}{1} & \textcolor{rpxHMute}{0} & \textcolor{rpxHMute}{9} & \textcolor{rpxHMute}{1} \\
\rpxHBottom
\end{rpxfit}
\end{table*}

\rpxTableContext{\textbf{Tracking and pose.} Tables~\ref{tab:raw-t3} and~\ref{tab:raw-t4} report tracking across four views and relative pose on the Clutter and Clean pair sets.}{generated/raw_t3_table}
% raw_t3_table already contains both T3 and T4; raw_t4_table is its duplicate.
\rpxTableContext{\textbf{Grounding.} Tables~\ref{tab:raw-t5} and~\ref{tab:raw-t6} report regular and in-context grounding for the same 12 VLM configurations.}{generated/raw_t5_table}
\begin{table*}[tp]
\centering
\caption{T6 in-context VQA: native grounding metrics per view and phase (Center: Center-in-GT), pooled over all planned question types, including the unscored spatial type, as in Table~\ref{tab:raw-t5}. \textbf{Bold}: best; \underline{underlined}: each model's worst phase or view. Shade: \rpxHSwatchR\ higher and \rpxHSwatchL\ lower is better, as in Table~\ref{tab:raw-t1}.}\label{tab:raw-t6}
\rpxHeatStyle{6.3}{2pt}
\begin{rpxfit}{\textwidth}{18}{@{}l@{}@{\hspace{3.30pt}\hspace{5pt}\hspace{3.30pt}}cccc@{\hspace{3.30pt}\hspace{5pt}\hspace{3.30pt}}cccc@{\hspace{3.30pt}\hspace{5pt}\hspace{3.30pt}}cccc@{}}
\rpxHTop
\rule{0pt}{9pt}\rpxHGroup{Model} & \multicolumn{4}{c}{\rpxHGroup{Acc@0.5\,\rpxUp{}}} & \multicolumn{4}{c}{\rpxHGroup{GIoU\,\rpxUp{}}} & \multicolumn{4}{c}{\rpxHGroup{Center\,\rpxUp{}}} \\
\rpxHGroupRules{\cmidrule(lr){2-5}\cmidrule(lr){6-9}\cmidrule(lr){10-13}}
& \rpxHHead{clu} & \rpxHHead{int} & \rpxHHead{cln} & \rpxHHead{ego} & \rpxHHead{clu} & \rpxHHead{int} & \rpxHHead{cln} & \rpxHHead{ego} & \rpxHHead{clu} & \rpxHHead{int} & \rpxHHead{cln} & \rpxHHead{ego} \\
\rpxHMid
Cosmos Reason2 8B & \rpxHCell{rpxHR}{0.398}{0.082} & \rpxHCell{rpxHR}{0.330}{\underline{0.060}} & \rpxHCell{rpxHR}{0.355}{0.068} & \rpxHCell{rpxHR}{0.997}{0.275} & \rpxHCell{rpxHR}{0.765}{$-$0.265} & \rpxHCell{rpxHR}{0.615}{\underline{$-$0.418}} & \rpxHCell{rpxHR}{0.621}{$-$0.412} & \rpxHCell{rpxHR}{0.901}{$-$0.127} & \rpxHCell{rpxHR}{0.795}{0.268} & \rpxHCell{rpxHR}{0.635}{\underline{\textbf{0.206}}} & \rpxHCell{rpxHR}{0.640}{0.208} & \rpxHCell{rpxHR}{0.944}{0.326} \\
Cosmos Reason2 2B & \rpxHCell{rpxHR}{0.274}{0.042} & \rpxHCell{rpxHR}{0.243}{\underline{0.032}} & \rpxHCell{rpxHR}{0.252}{0.035} & \rpxHCell{rpxHR}{0.631}{0.157} & \rpxHCell{rpxHR}{0.561}{$-$0.473} & \rpxHCell{rpxHR}{0.470}{$-$0.565} & \rpxHCell{rpxHR}{0.435}{\underline{$-$0.601}} & \rpxHCell{rpxHR}{0.726}{$-$0.305} & \rpxHCell{rpxHR}{0.407}{0.117} & \rpxHCell{rpxHR}{0.327}{0.086} & \rpxHCell{rpxHR}{0.317}{\underline{0.082}} & \rpxHCell{rpxHR}{0.653}{0.213} \\
Qwen3-VL 8B & \rpxHCell{rpxHR}{0.764}{\textbf{0.200}} & \rpxHCell{rpxHR}{0.463}{\underline{0.103}} & \rpxHCell{rpxHR}{0.674}{\textbf{0.171}} & \rpxHCell{rpxHR}{1.000}{\textbf{0.276}} & \rpxHCell{rpxHR}{0.915}{$-$0.112} & \rpxHCell{rpxHR}{0.754}{$-$0.276} & \rpxHCell{rpxHR}{0.729}{\underline{$-$0.302}} & \rpxHCell{rpxHR}{0.937}{$-$0.090} & \rpxHCell{rpxHR}{0.813}{\textbf{0.275}} & \rpxHCell{rpxHR}{0.569}{\underline{0.180}} & \rpxHCell{rpxHR}{0.697}{\textbf{0.230}} & \rpxHCell{rpxHR}{1.000}{\textbf{0.348}} \\
Qwen3-VL 2B & \rpxHCell{rpxHR}{0.510}{0.118} & \rpxHCell{rpxHR}{0.479}{\underline{\textbf{0.108}}} & \rpxHCell{rpxHR}{0.510}{0.118} & \rpxHCell{rpxHR}{0.994}{0.274} & \rpxHCell{rpxHR}{0.819}{$-$0.210} & \rpxHCell{rpxHR}{0.689}{$-$0.342} & \rpxHCell{rpxHR}{0.636}{\underline{$-$0.396}} & \rpxHCell{rpxHR}{1.000}{\textbf{{\boldmath$-$}0.026}} & \rpxHCell{rpxHR}{0.556}{0.175} & \rpxHCell{rpxHR}{0.481}{\underline{0.146}} & \rpxHCell{rpxHR}{0.489}{0.149} & \rpxHCell{rpxHR}{0.969}{0.336} \\
InternVL 3.5 14B & \rpxHCell{rpxHR}{0.646}{0.162} & \rpxHCell{rpxHR}{0.451}{0.099} & \rpxHCell{rpxHR}{0.528}{0.124} & \rpxHCell{rpxHR}{0.330}{\underline{0.060}} & \rpxHCell{rpxHR}{0.954}{\textbf{{\boldmath$-$}0.073}} & \rpxHCell{rpxHR}{0.810}{\textbf{{\boldmath$-$}0.219}} & \rpxHCell{rpxHR}{0.790}{\underline{\textbf{{\boldmath$-$}0.240}}} & \rpxHCell{rpxHR}{0.837}{$-$0.192} & \rpxHCell{rpxHR}{0.807}{0.273} & \rpxHCell{rpxHR}{0.587}{\underline{0.187}} & \rpxHCell{rpxHR}{0.651}{0.212} & \rpxHCell{rpxHR}{0.705}{0.233} \\
InternVL 3.5 1B & \rpxHCell{rpxHR}{0.274}{0.042} & \rpxHCell{rpxHR}{0.212}{0.022} & \rpxHCell{rpxHR}{0.234}{0.029} & \rpxHCell{rpxHR}{0.175}{\underline{0.010}} & \rpxHCell{rpxHR}{0.790}{$-$0.240} & \rpxHCell{rpxHR}{0.662}{$-$0.370} & \rpxHCell{rpxHR}{0.631}{\underline{$-$0.402}} & \rpxHCell{rpxHR}{0.792}{$-$0.238} & \rpxHCell{rpxHR}{0.474}{0.143} & \rpxHCell{rpxHR}{0.309}{\underline{0.079}} & \rpxHCell{rpxHR}{0.345}{0.093} & \rpxHCell{rpxHR}{0.361}{0.099} \\
PaliGemma 2 10B & \rpxHCell{rpxHR}{0.352}{0.067} & \rpxHCell{rpxHR}{0.299}{\underline{0.050}} & \rpxHCell{rpxHR}{0.305}{0.052} & \rpxHCell{rpxHR}{0.355}{0.068} & \rpxHCell{rpxHR}{0.549}{$-$0.485} & \rpxHCell{rpxHR}{0.483}{$-$0.552} & \rpxHCell{rpxHR}{0.454}{\underline{$-$0.582}} & \rpxHCell{rpxHR}{0.470}{$-$0.565} & \rpxHCell{rpxHR}{0.330}{0.087} & \rpxHCell{rpxHR}{0.276}{0.066} & \rpxHCell{rpxHR}{0.273}{\underline{0.065}} & \rpxHCell{rpxHR}{0.325}{0.085} \\
PaliGemma 2 3B & \rpxHCell{rpxHR}{0.423}{0.090} & \rpxHCell{rpxHR}{0.358}{0.069} & \rpxHCell{rpxHR}{0.398}{0.082} & \rpxHCell{rpxHR}{0.327}{\underline{0.059}} & \rpxHCell{rpxHR}{0.562}{$-$0.472} & \rpxHCell{rpxHR}{0.385}{$-$0.652} & \rpxHCell{rpxHR}{0.447}{$-$0.589} & \rpxHCell{rpxHR}{0.374}{\underline{$-$0.663}} & \rpxHCell{rpxHR}{0.386}{0.109} & \rpxHCell{rpxHR}{0.302}{0.076} & \rpxHCell{rpxHR}{0.348}{0.094} & \rpxHCell{rpxHR}{0.289}{\underline{0.071}} \\
DeepSeek-VL2 4.5B & \rpxHCell{rpxHR}{0.255}{0.036} & \rpxHCell{rpxHR}{0.209}{0.021} & \rpxHCell{rpxHR}{0.200}{\underline{0.018}} & \rpxHCell{rpxHR}{0.345}{0.065} & \rpxHCell{rpxHR}{0.503}{$-$0.532} & \rpxHCell{rpxHR}{0.412}{$-$0.624} & \rpxHCell{rpxHR}{0.397}{\underline{$-$0.640}} & \rpxHCell{rpxHR}{0.563}{$-$0.471} & \rpxHCell{rpxHR}{0.291}{0.072} & \rpxHCell{rpxHR}{0.230}{0.048} & \rpxHCell{rpxHR}{0.222}{\underline{0.045}} & \rpxHCell{rpxHR}{0.476}{0.144} \\
DeepSeek-VL2 1.0B & \rpxHCell{rpxHR}{0.228}{0.027} & \rpxHCell{rpxHR}{0.203}{0.019} & \rpxHCell{rpxHR}{0.206}{0.020} & \rpxHCell{rpxHR}{0.150}{\underline{0.002}} & \rpxHCell{rpxHR}{0.857}{$-$0.172} & \rpxHCell{rpxHR}{0.800}{$-$0.230} & \rpxHCell{rpxHR}{0.782}{\underline{$-$0.248}} & \rpxHCell{rpxHR}{0.817}{$-$0.212} & \rpxHCell{rpxHR}{0.466}{0.140} & \rpxHCell{rpxHR}{0.345}{0.093} & \rpxHCell{rpxHR}{0.353}{0.096} & \rpxHCell{rpxHR}{0.340}{\underline{0.091}} \\
Florence 2 0.77B & \rpxHCell{rpxHR}{0.262}{0.038} & \rpxHCell{rpxHR}{0.224}{0.026} & \rpxHCell{rpxHR}{0.237}{0.030} & \rpxHCell{rpxHR}{0.206}{\underline{0.020}} & \rpxHCell{rpxHR}{0.374}{$-$0.663} & \rpxHCell{rpxHR}{0.292}{$-$0.746} & \rpxHCell{rpxHR}{0.326}{$-$0.712} & \rpxHCell{rpxHR}{0.244}{\underline{$-$0.795}} & \rpxHCell{rpxHR}{0.266}{0.062} & \rpxHCell{rpxHR}{0.194}{0.034} & \rpxHCell{rpxHR}{0.209}{0.040} & \rpxHCell{rpxHR}{0.176}{\underline{0.027}} \\
Florence 2 0.23B & \rpxHCell{rpxHR}{0.277}{0.043} & \rpxHCell{rpxHR}{0.280}{0.044} & \rpxHCell{rpxHR}{0.280}{0.044} & \rpxHCell{rpxHR}{0.190}{\underline{0.015}} & \rpxHCell{rpxHR}{0.353}{$-$0.684} & \rpxHCell{rpxHR}{0.298}{$-$0.740} & \rpxHCell{rpxHR}{0.301}{$-$0.737} & \rpxHCell{rpxHR}{0.150}{\underline{$-$0.891}} & \rpxHCell{rpxHR}{0.271}{0.064} & \rpxHCell{rpxHR}{0.237}{0.051} & \rpxHCell{rpxHR}{0.237}{0.051} & \rpxHCell{rpxHR}{0.150}{\underline{0.017}} \\
\rpxHMid
\textcolor{rpxHMute}{\textit{Times worst}} & \textcolor{rpxHMute}{0} & \textcolor{rpxHMute}{5} & \textcolor{rpxHMute}{1} & \textcolor{rpxHMute}{6} & \textcolor{rpxHMute}{0} & \textcolor{rpxHMute}{1} & \textcolor{rpxHMute}{8} & \textcolor{rpxHMute}{3} & \textcolor{rpxHMute}{0} & \textcolor{rpxHMute}{5} & \textcolor{rpxHMute}{3} & \textcolor{rpxHMute}{4} \\
\rpxHBottom
\end{rpxfit}
\end{table*}

\subsection{Per-phase quality}
\begin{figure*}[t]
\centering
\rpxRoundedImage[width=\textwidth]{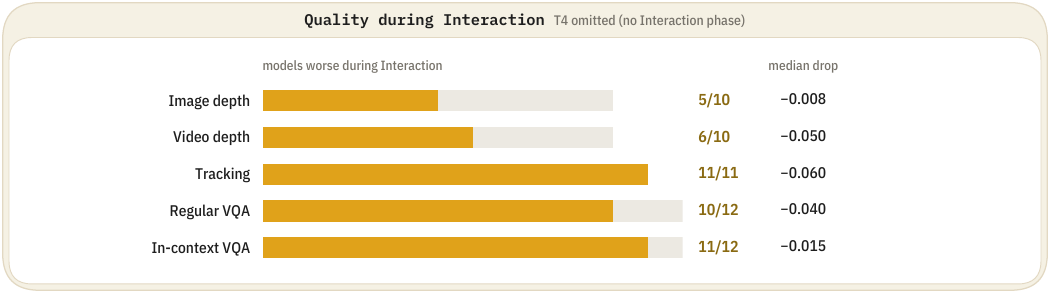}
\caption{Quality during Interaction. Bar and count: models whose Interaction $\mathcal J$ is below the mean of their Clutter and Clean $\mathcal J$. Right: median drop $\mathcal{J}_{\mathrm{Int}}-\tfrac12(\mathcal{J}_{\mathrm{Clu}}+\mathcal{J}_{\mathrm{Cln}})$ per task. T4 excludes Interaction and is omitted.}
\label{fig:slope-all}
\end{figure*}

Quality during Interaction is below the mean of Clutter and Clean for 5/10, 6/10, 11/11, 10/12, and 11/12 models in T1, T2, T3, T5, and T6 (Fig.~\ref{fig:slope-all}). The dip is most consistent in tracking (median $-0.06$) and largest for individual video-depth models (up to $-0.125$, GemDepth).
\subsection{Ego vs.\ exo comparison}
\label{sec:appendix-ego-exo}
\begin{figure*}[t]
\centering
\rpxRoundedImage[width=\textwidth]{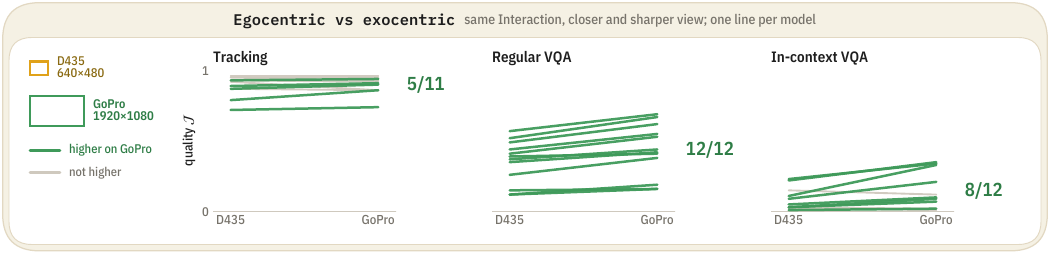}
\caption{Egocentric versus exocentric evaluation during Interaction for T3, T5, and T6. The head-mounted GoPro ($1920\times1080$) sees the scene closer and with $6.75\times$ the pixels of the D435 ($640\times480$). Each line is one model's $\mathcal J$ on the D435 and on the GoPro; green: higher on the GoPro (count at right).}
\label{fig:ego-vs-exo}
\end{figure*}

Tracking quality is unchanged by the head-mounted camera (ego exceeds exo for 5 of 11 trackers; median difference 0.00), whereas grounding improves on the GoPro view for all 12 regular-VQA models (median $+0.09$) and 8 of 12 in-context models (Fig.~\ref{fig:ego-vs-exo}).
Even so, $\Phi_{\mathrm{EE}}\leq0.86$ (Eq.~\eqref{eq:phi-ee}) for 7 of 11 trackers and 22 of 24 VLM evaluations: the camera reshapes the metric profile even when quality rises.
$\Phi_{\mathrm{EE}}$ is only weakly related to phase robustness ($\rho(\Phi,\Phi_{\mathrm{EE}})=+0.19$, $-0.06$, $+0.32$ for T3, T5, T6), so phase robustness does not predict camera robustness.
Viewpoint, resolution ($6.75\times$ pixel count), and framing differ between streams, so no single factor is isolated.
\subsection{Text vs.\ image prompt}
\label{sec:appendix-text-image}
\begin{figure*}[t]
\centering
\rpxRoundedImage[width=\textwidth]{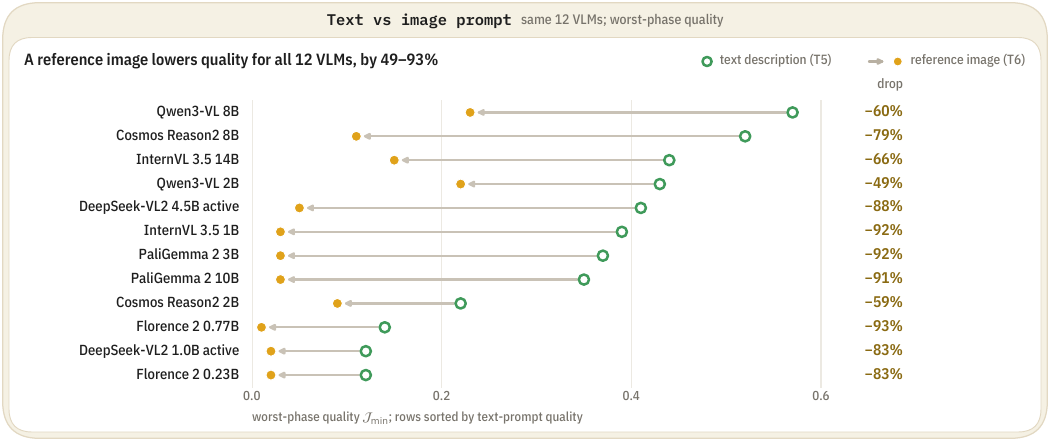}
\caption{What a reference image costs each VLM: worst-phase quality $\mathcal{J}_{\min}$ with a text description (T5, hollow) and with a reference image (T6, filled), and the relative drop $\Delta_{\mathrm{T6}}$ (Eq.~\eqref{eq:drop}, right). Rows are sorted by T5 quality; DeepSeek sizes denote active parameters.}
\label{fig:t5-vs-t6}
\end{figure*}

All 12 VLM configurations score lower on in-context than on regular VQA, with relative drops $\Delta_{\mathrm{T6}}$ (Eq.~\eqref{eq:drop}) of $49$--$93\%$ (Fig.~\ref{fig:t5-vs-t6}).
Rankings are largely kept (Spearman $\rho=0.87$; score Pearson $r=0.69$), and the drop shows no significant relation to regular-VQA quality ($\rho=-0.40$, $p=0.19$).
Qwen3-VL 8B and 2B remain strongest ($\mathcal{J}_{\min}=0.23$, $0.22$); InternVL~3.5 1B and both PaliGemma~2 models fall to $0.03$.
\subsection{Hardware efficiency and deployment trade-offs}
\label{sec:appendix-hardware}
\begin{figure*}[t]
\centering
\rpxRoundedImage[width=\textwidth]{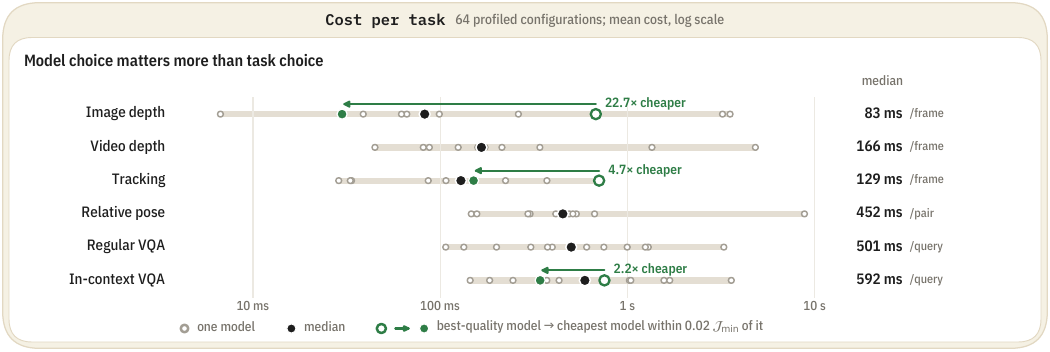}
\caption{Mean processing cost (Eq.~\eqref{eq:hw-cost}) per task on a log scale, for 64 configurations (all but SAM~3.1 with text prompts); the dark dot is the median, also given on the right. Green arrows run from the highest-$\mathcal J_{\min}$ model (ring) to the cheapest model within $0.02$ of it (filled), labelled with the cost ratio $r$ (Eq.~\eqref{eq:hw-ratio}); tasks without an arrow have no cheaper near-tie.}
\label{fig:hardware-summary}
\end{figure*}

\noindent\textbf{Profiling protocol.}
The release contains 76 hardware profiles: all 65 evaluated configurations and 11 egocentric tracking profiles.
All profiles use NVIDIA RTX 6000 Ada Generation GPUs (48\,GB each) and exclude warmup calls.
T1, T2, and T4 profiles use the Easy split; published quality summaries cover the full benchmark.
T1 and T4 time 1,000 adapter calls, T5 and T6 1,000 queries per model, T2 1,000 frames in 20 calls of 50 frames, and T3 1,000 MOS frames in four calls and 1,213 egocentric frames in five calls.
Timing covers synchronized adapter processing and excludes dataset I/O, weight loading, scoring, and persistence.

\smallskip
\noindent\textbf{Cost metrics.}
For calls $k=1,\dots,N_c$ with wall time $t_k$ over $n_k$ units (frame, pair, or query), the mean processing cost, tail cost, throughput, and the ratio between matched configurations $A$ and $B$ are
\begin{align}
\bar c &= \frac{\sum_k t_k}{\sum_k n_k}, & c_{95} &= P_{95}\Bigl(\frac{t_k}{n_k}\Bigr), \label{eq:hw-cost}\\
\mathrm{Rate} &= \frac{1}{\bar c}, & r_{A/B} &= \frac{\bar c_A}{\bar c_B}. \label{eq:hw-ratio}
\end{align}
For T2 and T3, $n_k>1$, so $\bar c$ is amortized per frame and quantiles describe clip calls.
Memory is the peak CUDA reservation for T1--T4 and the process footprint summed over GPUs for T5--T6.

\smallskip
\noindent\textbf{Requirements by task.}
Table~\ref{tab:hardware} lists cost, memory, and model size for all 76 profiles; Fig.~\ref{fig:hardware-summary} and the medians below cover 64 of the 65 configurations (all but SAM~3.1 with text prompts).
Median cost is 83, 166, and 129\,ms/frame for image depth, video depth, and tracking, 452\,ms/pair for relative pose, and 501 and 592\,ms/query for regular and in-context VQA.
Median memory is 4.2, 15.8, 2.9, and 3.5\,GiB for T1--T4 and about 38\,GiB for VQA (largely vLLM preallocation), whose largest model, DeepSeek-VL2, runs on two GPUs (85.8\,GiB in total).
Within a task, cost spans $25$--$527\times$ between the fastest and slowest configuration, so model choice matters more than task choice. Compute predicts speed in image depth and tracking (Spearman $\rho$ of mean cost with GFLOPs: $+0.94$ and $+0.92$) but not in relative pose ($+0.08$), where latency must be measured. Only four configurations run at 30\,fps (${\leq}33.3$\,ms: ZipDepth, UniDepth V2, XMem, MiTS), none for video depth or pose. VLM memory reflects the serving backend more than model size: in T5, the five single-GPU vLLM-served models (1--8\,B) each hold 41.7--43.1\,GiB, consistent with vLLM's up-front GPU preallocation, while Transformers-served models grow with size within each family (PaliGemma~2: 6.8 and 20.1\,GiB).

\smallskip
\noindent\textbf{Near-best quality is cheaper.}
In three tasks a near-best model costs a fraction of the best one: UniDepth V2 is $22.7\times$ cheaper than Metric3D V2 for $0.02$ lower $\mathcal J_{\min}$, SAM~2 is $4.7\times$ cheaper than SAM~2~Long for $0.01$, and Qwen3-VL 2B is $2.2\times$ cheaper than Qwen3-VL 8B in-context for $0.01$.
In video depth, relative pose, and regular VQA the best model (VGGT, DA3, Qwen3-VL 8B) has no cheaper near-tie.
The most robust configuration is cheaper than the highest-$\mathcal J_{\min}$ one in T1, T3, T5, and T6, at lower quality.

\smallskip
\noindent\textbf{No consistent cost--quality pattern.}
Spearman $\rho$ between mean cost and $\mathcal J_{\min}$ is $-0.10$, $-0.57$, $+0.48$, $-0.63$, $+0.62$, and $+0.41$ for T1--T6, and between mean cost and $\Phi$ it is $+0.26$, $+0.56$, $+0.24$, $+0.20$, $-0.72$, and $-0.40$.
The most consistent pattern is in grounding, where larger VLMs score higher ($\rho$ of size with $\mathcal J_{\min}$: $+0.71$ and $+0.66$ for T5 and T6, $p<0.05$); their lower $\Phi$ ($\rho=-0.56$, $-0.46$) is not significant at $n=12$.
In image depth, parameter count and GFLOPs are uncorrelated with $\mathcal J_{\min}$ ($\rho=+0.02$, $-0.18$): Lotus-2 spends $283{,}499$ GFLOPs for $0.35$, DA3 Metric-L $902$ GFLOPs for $0.70$.

\smallskip
\noindent\textbf{Protocol overhead and deployment.}
Egocentric tracking costs a median $1.52\times$ the MOS mean ($1.09$--$2.81\times$ across 11 trackers; Eq.~\eqref{eq:hw-ratio}), and in-context VQA $1.17\times$ the regular mean across 12 VLMs; the median P95/mean ratio is 1.46 (T5) and 1.60 (T6).
DA3 Metric-L (3.1\,GiB, 39\,ms), Cutie (0.8\,GiB, 34\,ms), and XMem (1.4\,GiB, 29\,ms) fit an 8\,GiB envelope at ${>}10$\,fps with $\mathcal J_{\min}=0.70$, $0.93$, and $0.89$.
Together, the highest-$\mathcal J_{\min}$ image-depth, tracking, and pose models (Metric3D V2, SAM~2~Long, DA3) reserve 19.6\,GiB, within a desktop GPU but beyond most edge accelerators; this sum is not a measured concurrent run, and other hardware will differ.

\begin{table*}[tp]
\centering
\caption{Hardware profiles of all 76 configurations: 65 evaluated ones and, under T3, each tracker's egocentric run (ego). Mean and P95 cost per unit (Eq.~\eqref{eq:hw-cost}); shade: per column within each task (T3 MOS and ego separately), on a log scale, lightest for the worst and darkest for the best value; \rpxHSwatchL\ lower is better. Memory is peak CUDA reservation (T1--T4) or process memory summed over GPUs (T5--T6; $^{\dagger}$two GPUs). Params are exact loaded-module counts with GFLOPs per unit (T1--T4) or the declared model size (T5--T6); ---: not reported. Ranked by mean cost within each task (T3 by the MOS run). Bold: lowest per task and protocol.}
\label{tab:hardware}
\rpxHeatStyle{5.6}{1.2pt}%
\definecolor{hwRed}{HTML}{B91C1C}\definecolor{hwAmb}{HTML}{E0A21A}\definecolor{hwGrn}{HTML}{3F9A5A}%
\newcommand{\hwBar}[2]{\begin{tikzpicture}[baseline=-0.6ex]\fill[rpxHTrack,rounded corners=1.4pt](0,-1.4pt)rectangle(\rpxHBarW,1.4pt);\begin{scope}[opacity=#2]\clip[rounded corners=1.4pt](0,-1.4pt)rectangle(#1\rpxHBarW,1.4pt);\shade[left color=hwRed,middle color=hwAmb,right color=hwGrn](0,-1.4pt)rectangle(\rpxHBarW,1.4pt);\end{scope}\end{tikzpicture}}%
\newcommand{\hwSection}[1]{\noalign{\vskip 1.2pt{\color{rpxHTrack}\hrule height 8.2pt depth 0pt}\vskip-8.2pt}\multicolumn{6}{@{\hspace{2pt}}l@{}}{\rule[-2.3pt]{0pt}{8.2pt}\rpxHSection{#1}}\\[-1.2pt]\noalign{\vskip 1.0pt}}%
\renewcommand{\arraystretch}{0.93}\setlength{\rpxHBarW}{14pt}%
\begin{minipage}[t]{0.492\textwidth}
\begin{tabular*}{\linewidth}[t]{@{}r@{\hspace{4pt}}l@{\extracolsep{\fill}}rrrr@{}}
\rpxHTop
\multicolumn{2}{@{}l}{\rule{0pt}{8.5pt}\rpxHGroup{Configuration}} & \multicolumn{2}{c}{\rpxHGroup{Cost per unit (ms)\,\rpxDown{}}} & \rpxHGroup{Memory\,\rpxDown{}} & \rpxHGroup{Model size} \\
\rpxHGroupRules{\cmidrule(r){1-2}\cmidrule(lr){3-4}\cmidrule(lr){5-5}\cmidrule(l){6-6}}
\rpxHHead{\#} & \rpxHHead{name} & \rpxHHead{mean} & \rpxHHead{P95} & \rpxHHead{GiB} & \rpxHHead{params / GFLOPs} \\
\rpxHMid
\hwSection{T1 image depth \hk{(ms/frame)}}
\hk{1} & ZipDepth & \rpxHCell{rpxHL}{1.000}{\textbf{6.7}} & \rpxHCell{rpxHL}{1.000}{\textbf{6.7}} & \rpxHCell{rpxHL}{1.000}{\textbf{0.13}} & \hk{6.1M / 7.9} \\\hg
\hk{2} & UniDepth V2 & \rpxHCell{rpxHL}{0.797}{29.9} & \rpxHCell{rpxHL}{0.780}{33.9} & \rpxHCell{rpxHL}{0.469}{4.26} & \hk{353.8M / 1393.5} \\\hg
\hk{3} & DA3 Metric-L & \rpxHCell{rpxHL}{0.762}{38.8} & \rpxHCell{rpxHL}{0.739}{46.1} & \rpxHCell{rpxHL}{0.515}{3.14} & \hk{334.2M / 902.2} \\\hg
\hk{4} & MoGe-2 ViT-L & \rpxHCell{rpxHL}{0.698}{62.1} & \rpxHCell{rpxHL}{0.692}{65.1} & \rpxHCell{rpxHL}{0.474}{4.12} & \hk{330.9M / 3162.1} \\\hg
\hk{5} & HyDen & \rpxHCell{rpxHL}{0.689}{66.3} & \rpxHCell{rpxHL}{0.680}{71.3} & \rpxHCell{rpxHL}{0.556}{2.41} & \hk{374.4M / 1436.8} \\\hg
\hk{6} & DA-V2 Large & \rpxHCell{rpxHL}{0.635}{98.9} & \rpxHCell{rpxHL}{0.619}{111.3} & \rpxHCell{rpxHL}{0.572}{2.16} & \hk{335.3M / 1806.1} \\\hg
\hk{7} & Depth Pro & \rpxHCell{rpxHL}{0.503}{261.5} & \rpxHCell{rpxHL}{0.496}{277.4} & \rpxHCell{rpxHL}{0.419}{5.92} & \hk{952.0M / 19247.4} \\\hg
\hk{8} & Metric3D V2 & \rpxHCell{rpxHL}{0.374}{677.7} & \rpxHCell{rpxHL}{0.372}{690.9} & \rpxHCell{rpxHL}{0.395}{6.92} & \hk{1377.7M / 13469.8} \\\hg
\hk{9} & FE2E & \rpxHCell{rpxHL}{0.163}{3218.9} & \rpxHCell{rpxHL}{0.162}{3247.5} & \rpxHCell{rpxHL}{0.163}{31.85} & \hk{12511.6M / 195375.4} \\\hg
\hk{10} & Lotus-2 & \rpxHCell{rpxHL}{0.150}{3532.6} & \rpxHCell{rpxHL}{0.150}{3560.6} & \rpxHCell{rpxHL}{0.150}{34.62} & \hk{17587.8M / 283498.9} \\\hg
\hwSection{T2 video depth \hk{(ms/frame)}}
\hk{1} & DA3 (video) & \rpxHCell{rpxHL}{1.000}{\textbf{44.8}} & \rpxHCell{rpxHL}{1.000}{\textbf{46.9}} & \rpxHCell{rpxHL}{0.353}{16.41} & \hk{410.9M / 2769.9} \\\hg
\hk{2} & Video-DA & \rpxHCell{rpxHL}{0.892}{81.2} & \rpxHCell{rpxHL}{0.876}{92.4} & \rpxHCell{rpxHL}{0.559}{9.33} & \hk{384.4M / 3099.4} \\\hg
\hk{3} & DVD v1.1 & \rpxHCell{rpxHL}{0.878}{87.5} & \rpxHCell{rpxHL}{0.884}{88.5} & \rpxHCell{rpxHL}{0.547}{9.66} & \hk{2245.8M / 6338.8} \\\hg
\hk{4} & ViGeo & \rpxHCell{rpxHL}{0.814}{124.7} & \rpxHCell{rpxHL}{0.819}{126.1} & \rpxHCell{rpxHL}{0.391}{14.81} & \hk{1255.4M / 12448.7} \\\hg
\hk{5} & GemDepth & \rpxHCell{rpxHL}{0.770}{158.6} & \rpxHCell{rpxHL}{0.775}{160.2} & \rpxHCell{rpxHL}{0.150}{28.63} & \hk{710.8M / ---} \\\hg
\hk{6} & VGGT & \rpxHCell{rpxHL}{0.754}{173.4} & \rpxHCell{rpxHL}{0.744}{190.0} & \rpxHCell{rpxHL}{0.277}{20.22} & \hk{1256.5M / 7880.3} \\\hg
\hk{7} & MonST3R & \rpxHCell{rpxHL}{0.716}{213.6} & \rpxHCell{rpxHL}{0.709}{229.4} & \rpxHCell{rpxHL}{1.000}{\textbf{2.79}} & \hk{571.2M / 1856.8} \\\hg
\hk{8} & ChronoDepth & \rpxHCell{rpxHL}{0.631}{340.6} & \rpxHCell{rpxHL}{0.636}{342.1} & \rpxHCell{rpxHL}{0.380}{15.26} & \hk{2254.4M / 18356.5} \\\hg
\hk{9} & DepthCrafter & \rpxHCell{rpxHL}{0.381}{1351.5} & \rpxHCell{rpxHL}{0.383}{1359.7} & \rpxHCell{rpxHL}{0.325}{17.74} & \hk{2254.4M / 76924.3} \\\hg
\hk{10} & RollingDepth & \rpxHCell{rpxHL}{0.150}{4822.7} & \rpxHCell{rpxHL}{0.150}{4853.5} & \rpxHCell{rpxHL}{0.351}{16.52} & \hk{1290.0M / 495071.2} \\\hg
\hwSection{T3 tracking \hk{(ms/frame)}}
\hk{1} & XMem & \rpxHCell{rpxHL}{1.000}{\textbf{28.7}} & \rpxHCell{rpxHL}{1.000}{\textbf{33.2}} & \rpxHCell{rpxHL}{0.870}{1.39} & \hk{62.2M / 1145.0} \\\hg
 & \hspace{5pt}\hk{$\hookrightarrow$\,ego} & \rpxHCell{rpxHL}{0.972}{80.2} & \rpxHCell{rpxHL}{0.980}{83.9} & \rpxHCell{rpxHL}{0.962}{1.88} &  \\\hg
\hk{2} & MiTS & \rpxHCell{rpxHL}{0.965}{33.1} & \rpxHCell{rpxHL}{0.995}{33.9} & \rpxHCell{rpxHL}{0.923}{1.12} & \hk{32.7M / 168.7} \\\hg
 & \hspace{5pt}\hk{$\hookrightarrow$\,ego} & \rpxHCell{rpxHL}{0.925}{93.0} & \rpxHCell{rpxHL}{0.933}{97.0} & \rpxHCell{rpxHL}{0.880}{2.46} &  \\\hg
\hk{3} & Cutie & \rpxHCell{rpxHL}{0.961}{33.6} & \rpxHCell{rpxHL}{0.985}{35.2} & \rpxHCell{rpxHL}{1.000}{\textbf{0.82}} & \hk{35.0M / 420.2} \\\hg
 & \hspace{5pt}\hk{$\hookrightarrow$\,ego} & \rpxHCell{rpxHL}{1.000}{\textbf{73.4}} & \rpxHCell{rpxHL}{1.000}{\textbf{78.9}} & \rpxHCell{rpxHL}{0.938}{2.03} &  \\\hg
\hk{4} & EdgeTAM & \rpxHCell{rpxHL}{0.727}{86.3} & \rpxHCell{rpxHL}{0.691}{112.2} & \rpxHCell{rpxHL}{0.858}{1.46} & \hk{13.9M / 770.9} \\\hg
 & \hspace{5pt}\hk{$\hookrightarrow$\,ego} & \rpxHCell{rpxHL}{0.831}{124.7} & \rpxHCell{rpxHL}{0.786}{152.2} & \rpxHCell{rpxHL}{1.000}{\textbf{1.66}} &  \\\hg
\hk{5} & GSAM2 & \rpxHCell{rpxHL}{0.673}{107.3} & \rpxHCell{rpxHL}{0.639}{138.0} & \rpxHCell{rpxHL}{0.543}{5.21} & \hk{396.7M / 4903.9} \\\hg
 & \hspace{5pt}\hk{$\hookrightarrow$\,ego} & \rpxHCell{rpxHL}{0.677}{202.2} & \rpxHCell{rpxHL}{0.637}{240.3} & \rpxHCell{rpxHL}{0.660}{5.04} &  \\\hg
\hk{6} & SAM 2 & \rpxHCell{rpxHL}{0.589}{150.7} & \rpxHCell{rpxHL}{0.561}{187.5} & \rpxHCell{rpxHL}{0.647}{3.43} & \hk{224.4M / 6110.6} \\\hg
 & \hspace{5pt}\hk{$\hookrightarrow$\,ego} & \rpxHCell{rpxHL}{0.655}{216.8} & \rpxHCell{rpxHL}{0.538}{324.9} & \rpxHCell{rpxHL}{0.767}{3.56} &  \\\hg
\hk{7} & SAM 2++ & \rpxHCell{rpxHL}{0.585}{153.1} & \rpxHCell{rpxHL}{0.611}{154.1} & \rpxHCell{rpxHL}{0.710}{2.65} & \hk{101.2M / 5308.9} \\\hg
 & \hspace{5pt}\hk{$\hookrightarrow$\,ego} & \rpxHCell{rpxHL}{0.619}{242.0} & \rpxHCell{rpxHL}{0.611}{259.6} & \rpxHCell{rpxHL}{0.840}{2.80} &  \\\hg
\hk{8} & SAM 3.1 (box) & \rpxHCell{rpxHL}{0.492}{223.3} & \rpxHCell{rpxHL}{0.514}{225.5} & \rpxHCell{rpxHL}{0.305}{13.66} & \hk{873.2M / 12198.2} \\\hg
 & \hspace{5pt}\hk{$\hookrightarrow$\,ego} & \rpxHCell{rpxHL}{0.613}{247.3} & \rpxHCell{rpxHL}{0.620}{252.9} & \rpxHCell{rpxHL}{0.288}{17.05} &  \\\hg
\hk{9} & DAM4SAM & \rpxHCell{rpxHL}{0.366}{371.3} & \rpxHCell{rpxHL}{0.383}{378.2} & \rpxHCell{rpxHL}{0.617}{3.87} & \hk{--- / 16932.4} \\\hg
 & \hspace{5pt}\hk{$\hookrightarrow$\,ego} & \rpxHCell{rpxHL}{0.350}{563.0} & \rpxHCell{rpxHL}{0.353}{573.7} & \rpxHCell{rpxHL}{0.741}{3.87} &  \\\hg
\hk{10} & SAM 2 Long & \rpxHCell{rpxHL}{0.207}{705.7} & \rpxHCell{rpxHL}{0.216}{729.7} & \rpxHCell{rpxHL}{0.676}{3.05} & \hk{224.4M / 25609.5} \\\hg
 & \hspace{5pt}\hk{$\hookrightarrow$\,ego} & \rpxHCell{rpxHL}{0.150}{1054.5} & \rpxHCell{rpxHL}{0.150}{1067.3} & \rpxHCell{rpxHL}{0.575}{6.66} &  \\\hg
\hk{11} & SAM 3.1 (text) & \rpxHCell{rpxHL}{0.150}{887.5} & \rpxHCell{rpxHL}{0.150}{948.0} & \rpxHCell{rpxHL}{0.150}{25.63} & \hk{873.2M / 36426.4} \\\hg
 & \hspace{5pt}\hk{$\hookrightarrow$\,ego} & \rpxHCell{rpxHL}{0.177}{969.1} & \rpxHCell{rpxHL}{0.173}{994.3} & \rpxHCell{rpxHL}{0.150}{26.73} &  \\\hg
\rpxHBottom
\end{tabular*}
\end{minipage}\hfill
\begin{minipage}[t]{0.492\textwidth}
\begin{tabular*}{\linewidth}[t]{@{}r@{\hspace{4pt}}l@{\extracolsep{\fill}}rrrr@{}}
\rpxHTop
\multicolumn{2}{@{}l}{\rule{0pt}{8.5pt}\rpxHGroup{Configuration}} & \multicolumn{2}{c}{\rpxHGroup{Cost per unit (ms)\,\rpxDown{}}} & \rpxHGroup{Memory\,\rpxDown{}} & \rpxHGroup{Model size} \\
\rpxHGroupRules{\cmidrule(r){1-2}\cmidrule(lr){3-4}\cmidrule(lr){5-5}\cmidrule(l){6-6}}
\rpxHHead{\#} & \rpxHHead{name} & \rpxHHead{mean} & \rpxHHead{P95} & \rpxHHead{GiB} & \rpxHHead{params / GFLOPs} \\
\rpxHMid
\hwSection{T4 relative pose \hk{(ms/pair)}}
\hk{1} & DA3 & \rpxHCell{rpxHL}{1.000}{\textbf{146.5}} & \rpxHCell{rpxHL}{1.000}{\textbf{169.0}} & \rpxHCell{rpxHL}{0.151}{9.60} & \hk{1355.7M / 6186.2} \\\hg
\hk{2} & Pi3X & \rpxHCell{rpxHL}{0.986}{156.9} & \rpxHCell{rpxHL}{0.991}{176.0} & \rpxHCell{rpxHL}{0.340}{6.79} & \hk{992.5M / ---} \\\hg
\hk{3} & Reloc3R & \rpxHCell{rpxHL}{0.855}{294.2} & \rpxHCell{rpxHL}{0.857}{331.5} & \rpxHCell{rpxHL}{0.884}{2.50} & \hk{426.5M / 1505.1} \\\hg
\hk{4} & MUSt3R & \rpxHCell{rpxHL}{0.850}{302.1} & \rpxHCell{rpxHL}{0.851}{341.4} & \rpxHCell{rpxHL}{1.000}{\textbf{2.02}} & \hk{423.4M / 2030.6} \\\hg
\hk{5} & VGGT & \rpxHCell{rpxHL}{0.784}{414.2} & \rpxHCell{rpxHL}{0.772}{495.1} & \rpxHCell{rpxHL}{0.150}{9.62} & \hk{1256.5M / 5533.6} \\\hg
\hk{6} & DUSt3R & \rpxHCell{rpxHL}{0.750}{489.5} & \rpxHCell{rpxHL}{0.756}{533.5} & \rpxHCell{rpxHL}{0.822}{2.80} & \hk{571.2M / 3713.6} \\\hg
\hk{7} & Fast3R & \rpxHCell{rpxHL}{0.741}{511.1} & \rpxHCell{rpxHL}{0.694}{712.2} & \rpxHCell{rpxHL}{0.554}{4.58} & \hk{647.6M / 2957.9} \\\hg
\hk{8} & CUT3R & \rpxHCell{rpxHL}{0.732}{533.3} & \rpxHCell{rpxHL}{0.714}{648.4} & \rpxHCell{rpxHL}{0.667}{3.72} & \hk{793.3M / 3211.2} \\\hg
\hk{9} & MonST3R & \rpxHCell{rpxHL}{0.685}{667.4} & \rpxHCell{rpxHL}{0.588}{1176.9} & \rpxHCell{rpxHL}{0.822}{2.80} & \hk{571.2M / 3713.6} \\\hg
\hk{10} & MASt3R & \rpxHCell{rpxHL}{0.150}{8819.2} & \rpxHCell{rpxHL}{0.150}{9219.8} & \rpxHCell{rpxHL}{0.726}{3.34} & \hk{688.6M / 4094.2} \\\hg
\hwSection{T5 regular VQA \hk{(ms/query)}}
\hk{1} & Florence 2 0.23B & \rpxHCell{rpxHL}{1.000}{\textbf{107.0}} & \rpxHCell{rpxHL}{1.000}{\textbf{151.4}} & \rpxHCell{rpxHL}{1.000}{\textbf{1.13}} & \hk{0.23B} \\\hg
\hk{2} & Florence 2 0.77B & \rpxHCell{rpxHL}{0.945}{133.7} & \rpxHCell{rpxHL}{0.945}{185.8} & \rpxHCell{rpxHL}{0.859}{2.32} & \hk{0.77B} \\\hg
\hk{3} & DeepSeek-VL2 1.0B & \rpxHCell{rpxHL}{0.845}{199.8} & \rpxHCell{rpxHL}{0.801}{319.6} & \rpxHCell{rpxHL}{0.285}{43.10} & \hk{1.00B} \\\hg
\hk{4} & Qwen3-VL 2B & \rpxHCell{rpxHL}{0.740}{304.7} & \rpxHCell{rpxHL}{0.637}{588.8} & \rpxHCell{rpxHL}{0.291}{41.75} & \hk{2.00B} \\\hg
\hk{5} & DeepSeek-VL2 4.5B\smash{$^{\dagger}$} & \rpxHCell{rpxHL}{0.689}{374.1} & \rpxHCell{rpxHL}{0.661}{539.4} & \rpxHCell{rpxHL}{0.150}{85.79} & \hk{4.50B} \\\hg
\hk{6} & Cosmos Reason2 2B & \rpxHCell{rpxHL}{0.674}{396.6} & \rpxHCell{rpxHL}{0.577}{739.1} & \rpxHCell{rpxHL}{0.291}{41.75} & \hk{2.44B} \\\hg
\hk{7} & PaliGemma 2 3B & \rpxHCell{rpxHL}{0.569}{605.4} & \rpxHCell{rpxHL}{0.525}{896.7} & \rpxHCell{rpxHL}{0.647}{6.83} & \hk{3.00B} \\\hg
\hk{8} & Qwen3-VL 8B & \rpxHCell{rpxHL}{0.517}{748.3} & \rpxHCell{rpxHL}{0.444}{1215.4} & \rpxHCell{rpxHL}{0.292}{41.70} & \hk{8.00B} \\\hg
\hk{9} & Cosmos Reason2 8B & \rpxHCell{rpxHL}{0.445}{996.9} & \rpxHCell{rpxHL}{0.381}{1540.7} & \rpxHCell{rpxHL}{0.292}{41.66} & \hk{8.00B} \\\hg
\hk{10} & PaliGemma 2 10B & \rpxHCell{rpxHL}{0.390}{1246.4} & \rpxHCell{rpxHL}{0.346}{1757.6} & \rpxHCell{rpxHL}{0.435}{20.08} & \hk{10.00B} \\\hg
\hk{11} & InternVL 3.5 1B & \rpxHCell{rpxHL}{0.381}{1292.1} & \rpxHCell{rpxHL}{0.396}{1455.7} & \rpxHCell{rpxHL}{0.719}{4.72} & \hk{1.10B} \\\hg
\hk{12} & InternVL 3.5 14B & \rpxHCell{rpxHL}{0.150}{3272.0} & \rpxHCell{rpxHL}{0.150}{3656.1} & \rpxHCell{rpxHL}{0.333}{33.85} & \hk{15.10B} \\\hg
\hwSection{T6 in-context VQA \hk{(ms/query)}}
\hk{1} & Florence 2 0.23B & \rpxHCell{rpxHL}{1.000}{\textbf{144.5}} & \rpxHCell{rpxHL}{1.000}{\textbf{250.0}} & \rpxHCell{rpxHL}{1.000}{\textbf{1.13}} & \hk{0.23B} \\\hg
\hk{2} & Florence 2 0.77B & \rpxHCell{rpxHL}{0.937}{183.6} & \rpxHCell{rpxHL}{0.956}{289.3} & \rpxHCell{rpxHL}{0.858}{2.33} & \hk{0.77B} \\\hg
\hk{3} & DeepSeek-VL2 1.0B & \rpxHCell{rpxHL}{0.861}{244.7} & \rpxHCell{rpxHL}{0.876}{378.7} & \rpxHCell{rpxHL}{0.285}{43.14} & \hk{1.00B} \\\hg
\hk{4} & Qwen3-VL 2B & \rpxHCell{rpxHL}{0.772}{341.9} & \rpxHCell{rpxHL}{0.738}{602.4} & \rpxHCell{rpxHL}{0.291}{41.83} & \hk{2.00B} \\\hg
\hk{5} & DeepSeek-VL2 4.5B\smash{$^{\dagger}$} & \rpxHCell{rpxHL}{0.750}{371.1} & \rpxHCell{rpxHL}{0.773}{534.6} & \rpxHCell{rpxHL}{0.150}{85.79} & \hk{4.50B} \\\hg
\hk{6} & Cosmos Reason2 2B & \rpxHCell{rpxHL}{0.711}{431.4} & \rpxHCell{rpxHL}{0.693}{699.4} & \rpxHCell{rpxHL}{0.291}{41.79} & \hk{2.44B} \\\hg
\hk{7} & Qwen3-VL 8B & \rpxHCell{rpxHL}{0.563}{753.5} & \rpxHCell{rpxHL}{0.519}{1254.3} & \rpxHCell{rpxHL}{0.291}{41.87} & \hk{8.00B} \\\hg
\hk{8} & Cosmos Reason2 8B & \rpxHCell{rpxHL}{0.482}{1023.9} & \rpxHCell{rpxHL}{0.462}{1518.6} & \rpxHCell{rpxHL}{0.291}{41.76} & \hk{8.00B} \\\hg
\hk{9} & PaliGemma 2 3B & \rpxHCell{rpxHL}{0.477}{1041.9} & \rpxHCell{rpxHL}{0.366}{2097.5} & \rpxHCell{rpxHL}{0.647}{6.83} & \hk{3.00B} \\\hg
\hk{10} & InternVL 3.5 1B & \rpxHCell{rpxHL}{0.369}{1566.8} & \rpxHCell{rpxHL}{0.415}{1778.4} & \rpxHCell{rpxHL}{0.706}{5.06} & \hk{1.10B} \\\hg
\hk{11} & PaliGemma 2 10B & \rpxHCell{rpxHL}{0.350}{1681.8} & \rpxHCell{rpxHL}{0.190}{3784.2} & \rpxHCell{rpxHL}{0.435}{20.08} & \hk{10.00B} \\\hg
\hk{12} & InternVL 3.5 14B & \rpxHCell{rpxHL}{0.150}{3586.4} & \rpxHCell{rpxHL}{0.150}{4329.2} & \rpxHCell{rpxHL}{0.328}{34.63} & \hk{15.10B} \\\hg
\rpxHBottom
\end{tabular*}
\end{minipage}
\end{table*}

\section{Toolkit and Extensibility}
\label{sec:appendix-toolkit}
The \coolname{} toolkit is open-source and distributed on PyPI as \texttt{rpx-benchmark}\footnote{\url{https://pypi.org/project/rpx-benchmark/}}; installation and usage are documented online\footnote{\url{https://irvlutd.github.io/RPX/toolkit-docs/}}.
Reference adapters for the evaluated models are packaged as task-specific container images on Docker Hub (\texttt{irvlutd/rpx}), so each result can be rerun in its recorded environment.
\subsection{Adding models}
A model is normally composed rather than subclassed: \texttt{BenchmarkableModel} pairs an input adapter, a model callable, and an output adapter, and implements \texttt{setup()} and \texttt{predict(batch)} on the author\textquotesingle{}s behalf. Subclassing \texttt{BenchmarkModel} directly is reserved for cases needing full control over how \texttt{Sample} objects are routed to the model. In either case the model declares its task, loads weights once in \texttt{setup()}, and returns typed predictions (e.g.\ \texttt{DepthPrediction}) from \texttt{predict(batch)}.
Relative-depth models set \texttt{depth\_output\_kind="relative"} so that the runner applies scale-and-shift alignment before scoring.
The runner validates predictions and reports native metrics; $\mathcal{J}_p$, $\mathcal{J}_{\min}$, and $\Phi$ are computed from them with the toolkit's diagnostic functions.
\subsection{Adding tasks}
A new task registers a \texttt{TaskSpec} (primary metric, required modalities, run callable) with \texttt{register\_task}, and each of its metrics registers a \texttt{MetricSpec} with direction and desirability window.
Because $\mathcal{J}$ and $\Phi$ consume only these specifications and per-cell scores, the MANOVA and desirability pipeline applies to the new task unchanged.

\end{document}